\documentclass[10pt,twocolumn,letterpaper]{article}

\usepackage[pagenumbers]{wacv} 
\usepackage{graphicx}
\usepackage{amsmath}
\usepackage{amssymb}
\usepackage{bbm}
\usepackage{booktabs}
\if@mathematic
   
\else
   
\fi

\DeclareMathOperator{\ReLU}{ReLU}

\newcommand\ver[1]{\rotatebox[origin=c]{90}{#1}}

\newcommand{\PAR}[1]{\noindent{\bf #1~}}

\usepackage{algorithm}
\usepackage{algpseudocode}

\usepackage{makecell} 
\usepackage{dsfont} 
\usepackage{pifont}
\usepackage{mathtools} 
\usepackage{multirow}
\usepackage{tikz}
\usetikzlibrary{calc, math}

\makeatletter
\robustify\@latex@warning@no@line
\makeatother
\usepackage{authblk}
\makeatletter
\renewcommand\AB@affilsepx{, \protect\Affilfont}
\makeatother

\definecolor{wacvblue}{rgb}{0.21,0.49,0.74}
\usepackage[pagebackref,breaklinks,colorlinks,allcolors=wacvblue]{hyperref}

\usepackage[accsupp]{axessibility}  

\title{Optimizing against Infeasible Inclusions from Data for Semantic Segmentation through Morphology}

\author[1]{Shamik Basu}
\author[2,3]{Luc Van Gool}
\author[3]{Christos Sakaridis}
\affil[1]{University Of Bologna}
\affil[2]{INSAIT, Sofia University St.~Kliment Ohridski}
\affil[3]{ETH Z\"urich}

\begin{document}
\maketitle
\begin{abstract}
State-of-the-art semantic segmentation models are typically optimized in a data-driven fashion, minimizing solely per-pixel or per-segment classification objectives on their training data. This purely data-driven paradigm often leads to absurd segmentations, especially when the domain of input images is shifted from the one encountered during training. For instance, state-of-the-art models may assign the label ``road'' to a segment that is included by another segment that is respectively labeled as ``sky''. However, the ground truth of the existing dataset at hand dictates that such inclusion is not feasible. Our method, Infeasible Semantic Inclusions (InSeIn), first extracts explicit inclusion constraints that govern spatial class relations from the semantic segmentation training set at hand in an offline, data-driven fashion, and then enforces a morphological yet differentiable loss that penalizes violations of these constraints during training to promote prediction feasibility. InSeIn is a light-weight plug-and-play method, constitutes a novel step towards minimizing infeasible semantic inclusions in the predictions of learned segmentation models, and yields consistent and significant performance improvements over diverse state-of-the-art networks across the ADE20K, Cityscapes, and ACDC datasets. Codebase will be made available.
Code is available at \url{https://github.com/SHAMIK-97/InSeIn}
\end{abstract}

\section{Introduction}

\label{sec:intro}
\begin{figure}[tb]
  \centering
  \subfloat{\includegraphics[width=0.5\linewidth]{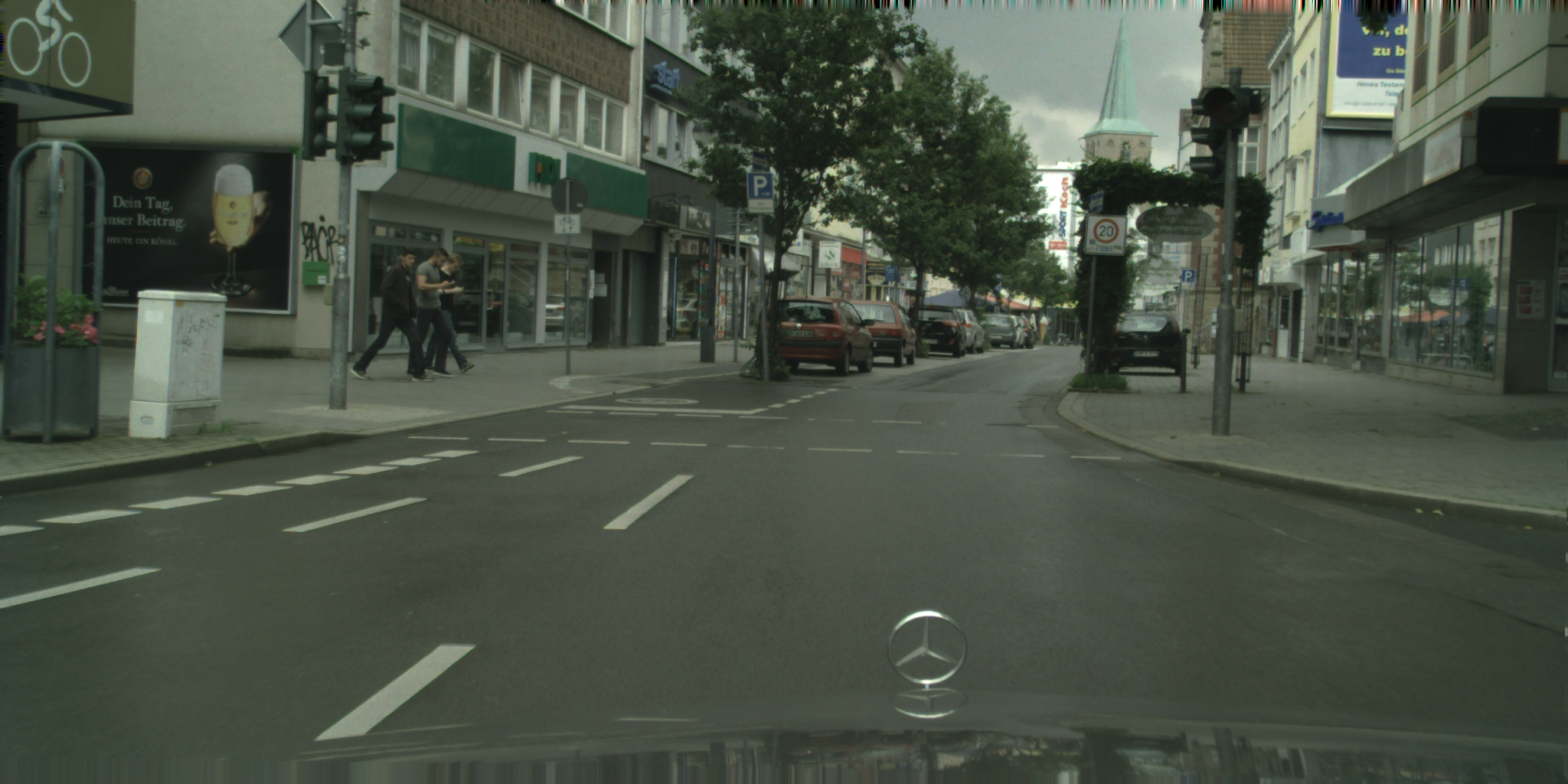}}
  \hfill
   \subfloat{\includegraphics[width=0.5\linewidth]{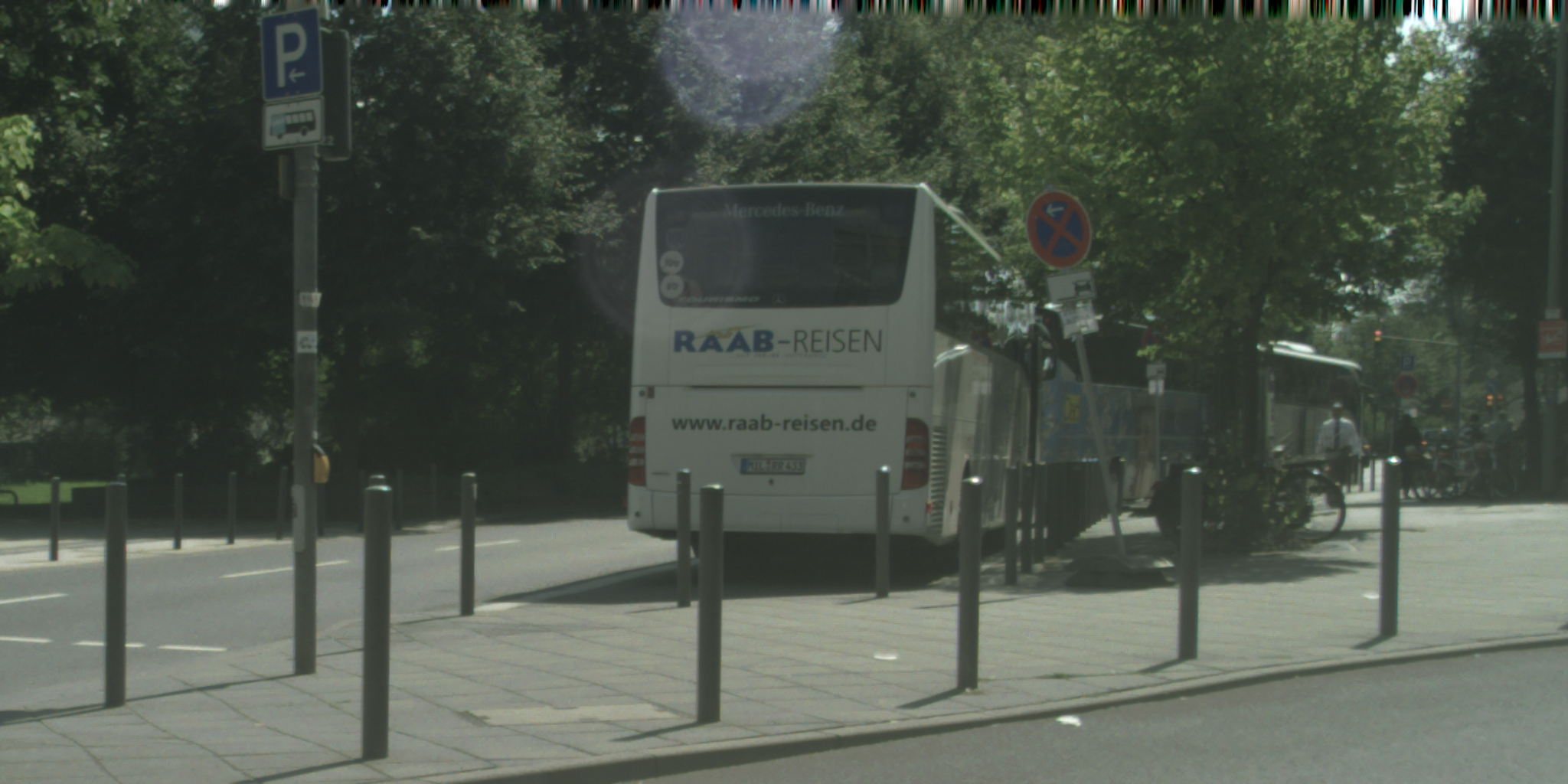}}
  \hfill
  \subfloat{\includegraphics[width=0.5\linewidth]{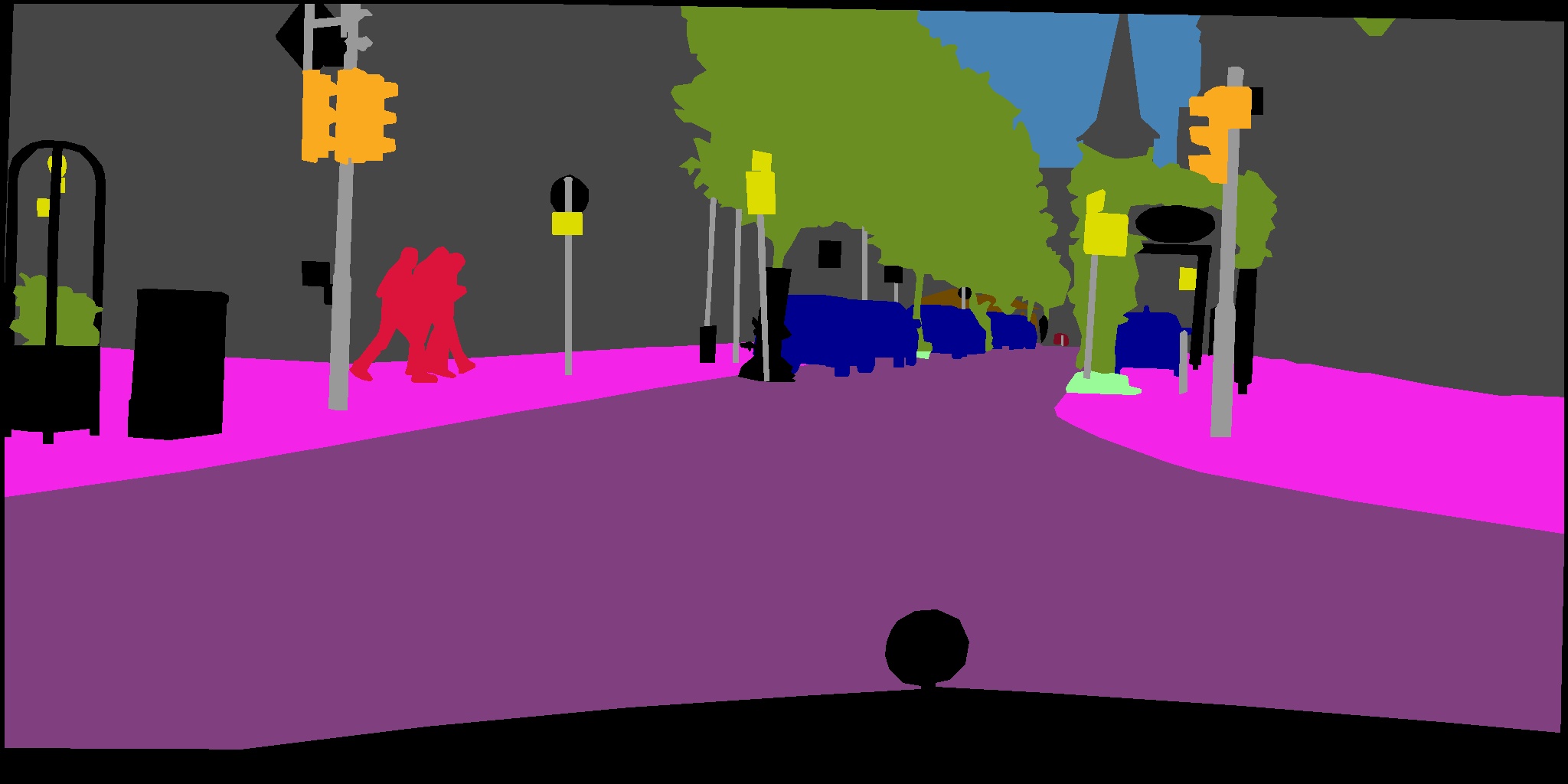}}
  \hfil
  \subfloat{\includegraphics[width=0.5\linewidth]{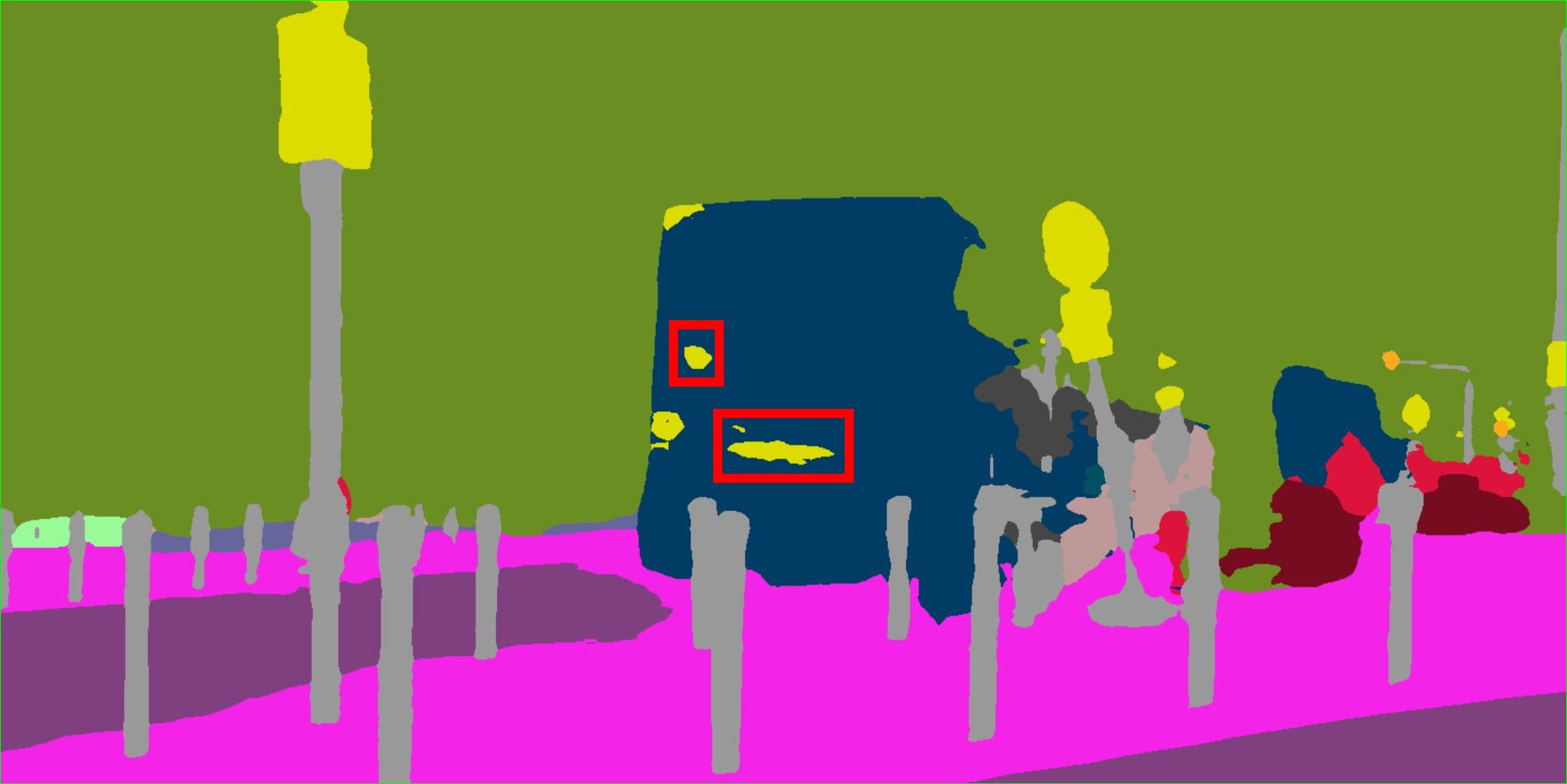}}
  \vspace{-3mm}
  \caption{Left: A \emph{traffic light} segment includes a \emph{building} segment in a semantic annotation from the ground truth of the dataset, which is a feasible inclusion. Right: A \emph{bus} segment includes \emph{traffic sign} segments (in the red boxes) in a semantic prediction, which is an \textbf{infeasible inclusion}. Our method addresses such physical infeasibilities in semantic segmentation. Best viewed on a screen and zoomed in.}
  \label{fig:exp:semseg}
  \vspace{-5mm}
\end{figure}

Semantic segmentation is a fundamental task in computer vision and enables many downstream applications.
The last decade has witnessed dramatic advances in this dense prediction task, driven by the introduction of end-to-end learned network architectures for solving it~\cite{LongFCN2015,PC2015Weak,CY2016Attention,ChenDeepLab2018,yu2015multi,chen2018encoder,guo2022segnext,WangSCJDZLMTWLX19,Zhao2017Pyramid,Cheng2022Mask2Former,YuanCW19OCRNet} and the development of ever stronger backbones coming from image classification~\cite{Simonyan15,he2016deep,Wang2021Pyramid,dosovitskiy2020image,Liu2021Swin,zhang2022resnest}.
Indeed, designing a suitable backbone for dense semantic prediction that balances global context aggregation with preservation of fine local details has been a primary research direction in semantic segmentation.
However, merely optimizing the backbone architecture neglects the \emph{structured prediction} character of the problem. Since structured prediction means that pixel-labels are not independent but are spatially and semantically correlated. In particular, the simple optimization of cross-entropy loss terms even by state-of-the-art network architectures~\cite{WangSCJDZLMTWLX19,Cheng2022Mask2Former} essentially poses semantic segmentation as a set of \emph{independent} pixel-level or- at best---segment-level prediction problems and only loosely promotes \emph{regularity} of the complete image-level outputs via the shared computation in the network across neighboring pixels. Very few deep-learning-based works~\cite{KeAAF2018} have focused on designing objectives that originate from the interaction between different pixels in the output predictions, and those that do focus only on local neighborhoods rather than large image regions. To the best of our knowledge, no previous method in semantic segmentation with deep learning has considered high-level pairwise objectives at the coarse level of semantic segments.

Aiming at promoting conformity of semantic segmentation models to high-level constraints, yet adhering to the taxonomy of the dataset at hand, we present a new method named Infeasible Semantic Inclusions (InSeIn). Our method has been empirically motivated by the observation that various state-of-the-art semantic segmentation networks~\cite{xie2021segformer,Cheng2022Mask2Former,YuanCW19OCRNet,alexandropoulos2024ovenet} make errors in their predictions, which are misclassifications or false positives, as illustrated in Fig.~\ref{fig:exp:semseg}. In particular, based on the Cityscapes~\cite{Cordts_2016_CVPR} taxonomy, a \emph{road} segment \emph{cannot} be entirely included in/surrounded by a \emph{sky} segment, i.e. it cannot happen that the entire boundary of the road segment coincides with a given sky segment, and the same holds for \emph{traffic sign} and \emph{bus} respectively.
We term such segment inclusions \emph{infeasible inclusions} and use them as an exemplary type of constraint that can provide a segmentation model with higher-level conformity based on the feasible spatial configurations of semantic segments in the ground truth.
 
While it is possible to exhaustively consider all pairs of semantic classes in the dataset at hand and manually determine which of them correspond to such infeasible inclusions, we argue and experimentally evidence that infeasible inclusions can be directly identified from \emph{large-scale} data. More specifically, InSeIn first processes the training set offline before training in a single pass and automatically identifies all class pairs for which a segment of the one class includes a segment of the other at least once in the data. InSeIn prunes this inclusion pair set by thresholding the relative frequency of each inclusion with respect to the co-occurrence of the respective pair to limit the impact of potential noise in the labels. The pairs obtained after pruning are \emph{valid inclusion} pairs (cf.\ Fig.~\ref{fig:exp:semseg}-left), and InSeIn should and does not penalize such inclusions in the outputs of the segmentation network. By contrast, pairs of co-occurring classes for which a complete inclusion of one class by the other is never found in the large-scale data are deemed as infeasible inclusion pairs, and they contribute to a feasibility objective that we define for optimization during training.

More specifically, the major contribution of InSeIn is a novel inclusion loss, which is used as an additional objective besides standard cross-entropy for optimizing a generic segmentation network. Our inclusion loss is computed based on the softmax outputs of the employed network, similarly to cross-entropy, which allows its general application to the vast majority of deep-learning-based semantic segmentation models. The key idea of computing this loss is to identify ``soft'' infeasible inclusions, which occur as regions in the softmax score maps of infeasible inclusion class pairs in which the ranking of the two softmax values is inverted compared to their surroundings. We recognize that this simply amounts to a grayscale morphological area opening~\cite{Gonz2009Digital,VincentMorphology1994} and derive a differentiable, parameter-free, and efficient implementation of it, which is normally included in the forward and backward pass of standard backpropagation.

We have achieved a consistent performance improvement with InSeIn across three central semantic segmentation benchmarks, i.e.\ ADE20K~\cite{zhou2019semantic}, Cityscapes~\cite{Cordts_2016_CVPR}, and ACDC~\cite{Sakaridis2021ACDC}, and across a variety of state-of-the-art semantic segmentation networks on which we have implemented our method, i.e.\ SegFormer~\cite{xie2021segformer}, OCRNet~\cite{YuanCW19OCRNet}, Mask2Former~\cite{Cheng2022Mask2Former}, OneFormer~\cite{jain2023oneformer} and SegMAN~\cite{SegMAN}. Importantly, we have observed a substantial quantitative reduction of infeasible inclusions in our semantic predictions.


\section{Related Work}
\label{sec:related}


\begin{figure*}[bt]
\centering
\includegraphics[width=\textwidth,trim={0 150 0 80}]{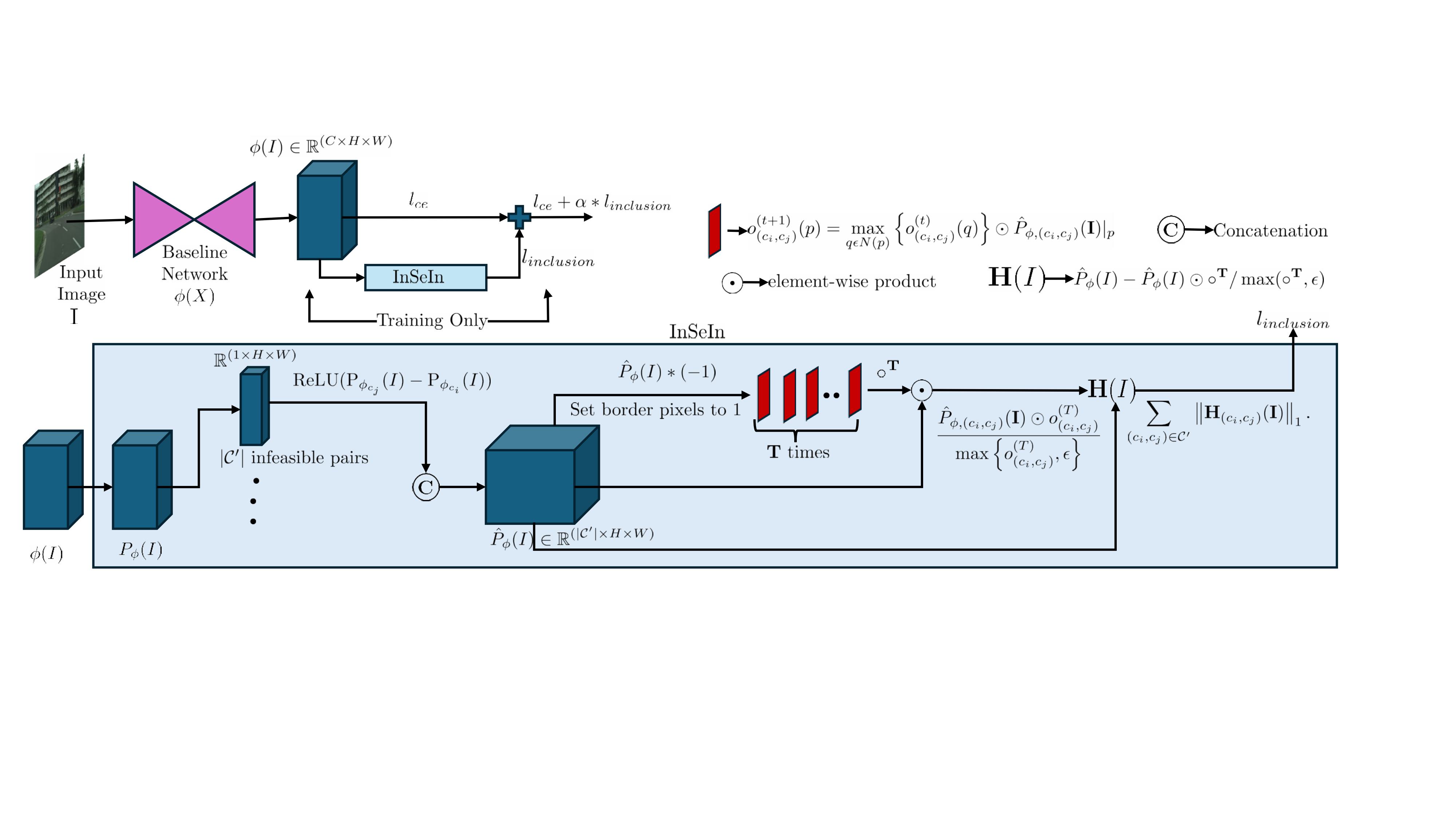}
\caption{\textbf{Overview of InSeIn.} Top left: the complete network architecture, where the standard cross-entropy loss $l_{\text{ce}}$ from the baseline network $\phi$ is added to the inclusion loss $l_{\text{inclusion}}$ computed by InSeIn. cf. Sec.~\ref{sec:method overview}. Bottom: The pipeline of InSeIn, cf. Sec.~\ref{sec: InSeIn}. For the softmax outputs $P_{\phi}(I)$ of the network and for each pair of classes $(c_i,c_j) \in \mathcal{C'}$ that signifies an infeasible inclusion where $c_i$ cannot include $c_j$, we take the difference of the respective softmax scores, $\text{P}_{\phi_{c_j}}(I)- \text{P}_{\phi_{c_i}}(I)$, and rectify it with a $\ReLU$. After concatenating all such $\mathcal{C'}$ rectified difference maps channel-wise into $\hat{P}_{\phi}(I)$, we negate the latter and set its border pixels to 1. After that, an iterative operation for area opening is performed $T$ times on $\hat{P}_{\phi}(I)$. In each iteration, we perform max-pooling with a 3x3 kernel and a stride of 1, and the result is multiplied with $\hat{P}_{\phi}(I)$ element-wise. The final area-opened tensor channels differ from their counterparts in $\hat{P}_{\phi}(I)$ only across regions of incorrect inclusions. This element-wise difference is stored in $\mathbf{H}(I)$ and the $L_1$ norm of the latter, capturing both the spatial extent and the intensity of infeasible inclusions, constitutes the inclusion loss $l_{\text{inclusion}}$ employed in InSeIn.}
\vspace{-4mm}
\label{fig:feature-map} 
\end{figure*}

\PAR{Semantic Segmentation.} Semantic segmentation deals with pixel classification. Pre-deep learning era methods mainly used lines and contour detection to classify similar pixel groups. Pixel misclassifications are mainly associated with boundary regions of any class segment and are studied in~\cite{LOPEZMOLINA20131125}. Here they classified pixel errors into 3 categories: false positives where the predicted edges erroneously appear in non-boundary areas, merging error occurs when the predicted results fail to capture the corresponding edges, resulting in false negatives and displacement error arises when the predicted edges deviate from their true positions. Such displacements may occur when the image features are misaligned. Pre-deep learning era attempted to segment misclassified pixels using graph searching, Laplace operator and other heuristic methods~\cite{VANVLIET1989167,MARTELLI1972169,YIN20041407,1672317,Griffith1973,PAPARI201179}.

\PAR{Modern semantic segmentation methods.} The challenge to classify pixels in semantic segmentation is studied in~\cite{Li2017Semantic} where the authors designed an architecture comprised of several sub-models arranged in a layer cascade fashion. Another paper~\cite{Deng2022NightLab,Ying2025SCDF,Wei2025LNFormer} where the authors devised a framework for nighttime segmentation, where the input data is highly noisy. In this work, they have designed a seperate module to segment hard regions in an image and another module to segment the easy regions. Finally they merged the output of 2 modules to get the final prediction. Another paper ~\cite{Ding2019BFP} where authors divided the input image into several patches and visualized the whole image as a unidirectional acyclic graph where the vertices are patches. Then they constituted a sperate ground truth which constitutes a seperate class for the boundaries and optimzed the model with the newly formed ground truth.

\PAR{Contribution of loss function in Semantic segmentation.} Pixel misclassification problem in segmentation task is handled using depth cues in semantic segmentation in the paper~\cite{Gu2021Depthsegmentation,Nie2024ImageDehazing}. Here the authors constituted 2 loss-weight modules which output a loss weight map by employing two depth-related measurements of hard pixels: Depth Prediction Error and Depth-aware Segmentation Error. The loss weight map is optimized with segmentation loss to train the network. Another paper~\cite{CHEN2025110599} devised a custom loss function to tackle misclassification by leveraging the sigmoid function to increase the loss and loss derivative of misclassified and uncertain pixels. Another popular method is Binary Focal Loss~\cite{Lin2020BFL,Antonutti2025BiParamLoss,Hunafa2025WeightedEnsemble,Huang2025Morphology,Wang2025Bootstrap} where the author used the normal cross entropy loss and down-weighted the easily classified pixels. Another paper~\cite{VALVERDE2023109208} developed a region wise(RW) loss through pixel-wise multiplication of RW map and softmax output for biomedical image segmentation.

All the above methods attempted to solve false positive regions at the pixel level. None of them takes into account the pairwise semantic relationship between the classes, as we do in our approach. We have also observed that false positives occur not only at the hard regions but also in easy-to-classify segments. We focus on the case of false positive segments in the form of infeasible inclusions and aim at optimizing against this type of error.

\begin{figure*}[tb]
  \centering
  \includegraphics[width=\textwidth,trim={0 75 0 280},clip]{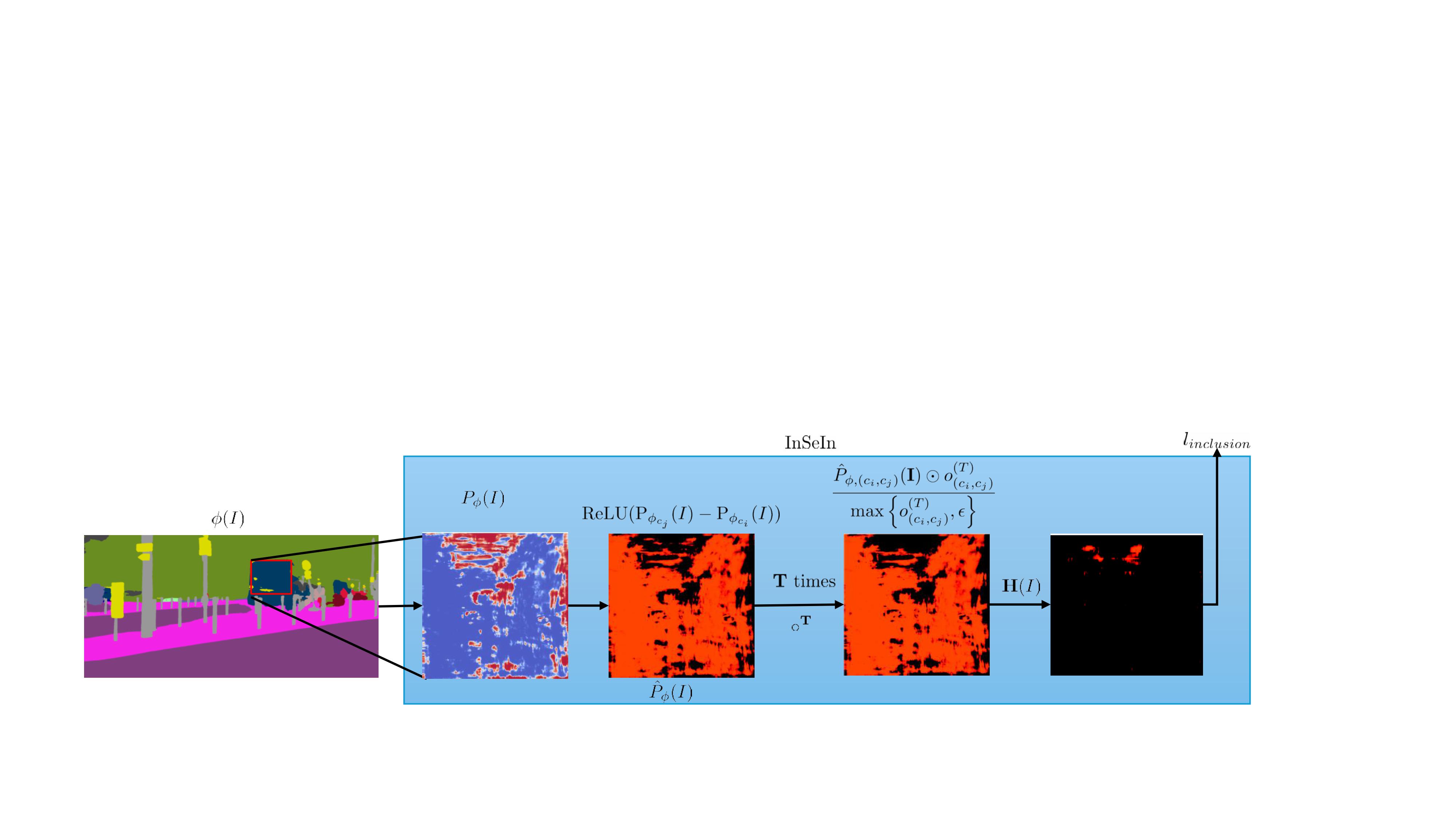}
  \caption{\textbf{Visualization of feature maps within InSeIn for a class pair corresponding to an infeasible inclusion.} In this example semantic prediction of SegFormer~\cite{xie2021segformer} on Cityscapes~\cite{Cordts_2016_CVPR}, \emph{bus} infeasibly includes \emph{traffic sign}. In the blue frame, we first show the difference map between the softmax scores of the two classes using the \emph{coolwarm} colormap, where blue tones indicate larger scores for \emph{traffic sign} and red for \emph{bus}, respectively. We rectify the difference between the two softmax maps and show only the respective channel of the 3D tensor $\hat{P}_{\phi}(I)$, where black indicates zeros and red tones indicate positive values. The area opening operation is performed on $\hat{P}_{\phi}(I)$ and the regions in which $\hat{P}_{\phi}(I)$ is positive (red) but which are not connected to the border, i.e.\ \emph{traffic sign} segments infeasibly included in \emph{bus} segments, are opened. Exactly these segments are isolated in the final tensor $\mathbf{H}(\mathbf{I})$ and are used to compute our inclusion loss $l_{\text{inclusion}}$.}
  \label{fig:softmax}
 \vspace{-5mm} 
\end{figure*}

\section{InSeIn}
\subsection{Method Overview}
\label{sec:method overview}


Fig.~\ref{fig:feature-map} shows the architectural overview of InSeIn. A 2D semantic segmentation network $\phi(X)$ is fed with an image $\mathbf{I} \in \mathbb{R}^{3\times H\times W}$ and produces the initial output $\phi(\mathbf{I}) \in \mathbb{R}^{C\times H\times W}$, where $C$ is the number of classes. InSeIn takes $\phi(\mathbf{I})$ as input and computes a novel inclusion loss, $l_{\text{inclusion}}$. We identify infeasible inclusions for class pairs of interest via a morphological area opening operation, which is fully differentiable and is normally included in the forward and backward pass during training. Our loss $l_{\text{inclusion}}$ is then added to the standard cross-entropy loss $l_{\text{ce}}$ for $\phi(\mathbf{I})$ to obtain the total loss as
\vspace{-3mm}
\begin{equation}
  l_{\text{total}} = l_{\text{ce}} + \alpha{}l_{\text{inclusion}}, 0<\alpha<1.
  \label{eq:total-loss}  
  \vspace{-2mm}
\end{equation}

$l_{\text{total}}$ is normally backpropagated to optimize the weights of $\phi$. Thus, InSeIn is a differentiable morphological module that enhances the optimization and conformity of any segmentation network by integrating high-level data-driven constraints in the form of infeasible inclusions. Notably, the InSeIn module is both learned parameter-free \emph{and} differentiable, allowing gradients to propagate through the network during training. The only extra parameter of InSeIn is $\alpha$ in \eqref{eq:total-loss}, a hyperparameter which is used to balance the inclusion loss and the standard cross-entropy loss.

\begin{algorithm}[tb]
\caption{Computing the set $\mathcal{U}$ of feasible inclusion class pairs from the training set ground-truth annotations $\mathcal{S}$. The set $\mathcal{P}$ contains all pairs of non-identical class labels for the dataset taxonomy. We aim to check whether the first element of the pair, $p$, can include the second one, $q$. We are taking 5 pixels as the minimum threshold for a \texttt{contour} of $q$ completely overlapping with any segment of $p$. Function \textbf{C} returns (i) the number of connected segments of class $q$ in the input annotation, and (ii) a discrete image, with each pixel labeled with its connected component ID or 0 for the background. Function \textbf{F} fills the holes of the mask of the connected component at hand. Then, function \textbf{Cont} returns a binary image in which only pixels belonging to the outer contour of the respective connected component are set.}
\label{alg:check_feasible_inclusion_constraint}
\begin{algorithmic}[1]
\Require $\mathcal{P} = \{(p,q):p \in \{1,\dots,C\},q \in \{1,\dots,C\} \setminus \{p\}\}$
\Require $\mathcal{S} = \{I:\,I \in \mathbb{N}^{1,H,W} \}$
\Ensure Set $\mathcal{U}$ of feasible inclusion class pairs
\State $\texttt{count} = 0$
\State $\texttt{count\_freq} = 0$
\State $\mathcal{U} \gets \emptyset$
\For{ $(p,q) \gets  \mathcal{P}$}
\For{ $S \gets \mathcal{S}$}
\State\texttt{mask\_1} $ \gets \mathbbm{1}[S=p] \in \mathbb{R}^{1\times H\times W}$
\State\texttt{mask\_2} $\gets \mathbbm{1}[S=q] \in \mathbb{R}^{1\times H\times W}$
\If{\texttt{mask\_1} \textbf{not empty} $\land$ \texttt{mask\_2} \textbf{not empty}}
\State $\texttt{count\_freq} \gets \texttt{count\_freq}+1$
\EndIf
\State\texttt{component} , \texttt{noOFLabels} $\gets \textbf{C}(\texttt{mask\_2})$
\For{ $l \gets 1$ to \texttt{noOFLabels}}
\State\texttt{comp\_mask} $\gets (\mathbbm{1}[\texttt{component}=l])$
\State\texttt{comp\_filled} $\gets \textbf{F}(\texttt{comp\_mask})$
\State \texttt{contour} $\gets \textbf{Cont}(\texttt{comp\_filled})$
\State \texttt{result} $\gets \texttt{contour}\land \texttt{mask\_1}$
\If{$\textbf{NNZ}(\texttt{result})=\textbf{NNZ}(\texttt{contour}) \land \textbf{NNZ}(\texttt{contour}) \geq 5$}
\State $\texttt{count} \gets \texttt{count}+1$
\State \textbf{break}
\EndIf
\EndFor
\EndFor
\If{$\texttt{count}/\texttt{count\_freq}*100 > 2$}
\State $\mathcal{U} \gets \mathcal{U} \cup \{(p,q)\}$
\State $\texttt{count} = 0$
\State $\texttt{count\_freq} = 0$
\EndIf
\EndFor
\end{algorithmic}
\end{algorithm}

\subsection{Computing Infeasible Inclusion Pairs}
\label{sec:infeasiblity}

We apply Algorithm~\ref{alg:check_feasible_inclusion_constraint} to the training split of the dataset at hand for optimizing and evaluating the segmentation model, to compute the set $\mathcal{U}$ of class pairs corresponding to feasible, or valid, inclusions. Intuitively, $\mathcal{U}$ contains class pairs for which a segment of the first class includes a segment of the second one at least in one case in the training set. The set of class pairs corresponding to \emph{infeasible inclusions}, $\mathcal{C}^{\prime}$, is then simply the set-theoretic difference $\mathcal{C}^{\prime} = \mathcal{P} \setminus \mathcal{U}$, where 
$\mathcal{P}$ is the set of all possible non-identical class pairs. In particular, any pair in $\mathcal{U}$ whose relative frequency of inclusion with respect to co-occurance of the class pairs is less than 2\% is considered noise or an outlier, and is removed from $\mathcal{U}$.

In Algorithm~\ref{alg:check_feasible_inclusion_constraint}, we first compute for each annotation $S$ in the training set two binary masks of the same dimension as $S$, in which pixels labeled as class $p$ and class $q$ are set as foreground, respectively. We then compute the connected components of class $q$, i.e.\ the class whose possibility to be included is checked, in \texttt{mask\_2}. For each such connected component, we fill any holes inside it to remove any interior contour in the next step, as it is irrelevant. After that, we compute the outer contour of the connected component at hand based on an 8-neighborhood and check whether it fully falls into pixels belonging to class $p$. If the latter is true and the length of the contour is non-negligible, we declare the case at hand as a valid inclusion, and then if the relative frequency of the inclusion to the co-occurrence of the pair is non-negligible, we add the pair $(p,q)$ to the set $\mathcal{U}$.

\subsection{Inclusion Loss}
\label{sec: InSeIn}

To compute our novel inclusion loss, we operate on the softmax maps $P_{\phi}(\mathbf{I})$ output by the segmentation network and index them channel-wise for each of the $|\mathcal{C}^{\prime}|$ infeasible inclusion pairs. Our core intuition is to identify in the pairs of softmax score maps that correspond to each infeasible inclusion case in $\mathcal{C}^{\prime}$ those regions across which class $c_j$---the includee---has a larger score than class $c_i$---the includer---and around which the opposite is true. These regions indicate ``soft'' infeasible inclusions in the softmax predictions, which should be penalized proportionally to the area they occupy and to the variable ``intensity'' of the inclusion, i.e.\ the magnitude of the difference between the softmax scores of classes $c_i$ and $c_j$ at each pixel of these regions. The computation of such a loss, which we term the inclusion loss, can be directly achieved in a differentiable formulation, amenable to standard backpropagation, via a grayscale morphological~\cite{Gonz2009Digital} area opening of the aforementioned softmax difference maps. We implement this opening through an iterative algorithm, detailed in the following.

For each ``infeasible pair'' $(c_i,c_j) \in \mathcal{C}^{\prime}$, we denote with $\text{P}_{\phi_{c_i,c_j}}(\mathbf{I}) \in \mathbb{R}^{2\times H\times W}$ the concatenation of $\text{P}_{\phi_{c_i}}(\mathbf{I})$ and $\text{P}_{\phi_{c_j}}(\mathbf{I})$. The difference between the two channels of the latter tensor is rectified with a ReLU as $\ReLU(\text{P}_{\phi_{c_j}}(\mathbf{I})- \text{P}_{\phi_{c_i}}(\mathbf{I})) \in \mathbb{R}^{1\times H\times W}$ (cf.\ Fig.~\ref{fig:feature-map}). After that, all $|\mathcal{C}^{\prime}|$ rectified pairwise differences are concatenated to obtain the tensor $\hat{P}_{\phi}(\mathbf{I}) \in \mathbb{R}^{|\mathcal{C}^{\prime}|\times H \times W}$.


For each $(c_i,c_j) \in \mathcal{C}^{\prime}$, we initialize a running map $o^{(0)}_{(c_i,c_j)} \in \mathbb{R}^{1\times{}H\times{}W}$ so that its border pixels are set to 1 and the rest are equal to 0. The intuition here is that infeasibly included regions are always disconnected from the border of the image. Thus, positive values of the map $o^{(t)}_{(c_i,c_j)}$ that are gradually propagated from the border to the interior of the image via iterative 8-neighborhood-based dilation (a.k.a.\ max-pooling) will \textit{never} reach the aforementioned regions if at each iteration they are multiplied with $\ReLU(\text{P}_{\phi_{c_j}}(\mathbf{I})- \text{P}_{\phi_{c_i}}(\mathbf{I}))$, as the latter is 0 everywhere around infeasibly included regions.
The above, intuitively described iterative operation can be formally written as
\vspace{-2mm}
\begin{equation}
    o^{(t+1)}_{(c_i,c_j)}(p) = \displaystyle \max_{q \epsilon N(p)} \left\{ 
    o^{(t)}_{(c_i,c_j)}(q) \right\} \odot \hat{P}_{\phi,(c_i,c_j)}(\mathbf{I})|_p,
    \label{eq:1}
\end{equation}
where $N(p)$ denotes the structuring element of the dilation operator or, equivalently, the max-pooling kernel, which we set to an 8-connected neighborhood, i.e.\ a 3x3 kernel. We iterate \eqref{eq:1} $T$ times, where $T = \max\{2, \min\{H, W\}/2\}$ and $H$ and $W$ are the height and width of the input image $\mathbf{I}$. Namely, we iterate for the minimal number of times that is required for initial positive values at image borders in $o^{(0)}_{(c_i,c_j)}$ to be able to reach any pixel in the interior. The final map $o^{(T)}_{(c_i,c_j)}$ is by construction $0$ at all pixels that belong to infeasibly included regions. This property implies that 
\vspace{-1mm}
\begin{equation}
    \label{eq:area:opening}
    \frac{\hat{P}_{\phi,(c_i,c_j)}(\mathbf{I}) \odot o^{(T)}_{(c_i,c_j)}}{\max\left\{o^{(T)}_{(c_i,c_j)}, \epsilon\right\}}
\vspace{-2mm}    
\end{equation}
constitutes the grayscale area opening we are after for the infeasible inclusion class pair $(c_i,c_j)$ (cf.\ Fig.~\ref{fig:softmax}), where $\epsilon$ is a small positive constant.

The relevant quantity for computing our inclusion loss is the difference between the original rectification output and its area opening from~\eqref{eq:area:opening}. This difference is given by
\begin{equation}
   \mathbf{H}_{(c_i,c_j)}(\mathbf{I}) = \displaystyle  \hat{P}_{\phi,(c_i,c_j)}(\mathbf{I}) - \frac{\hat{P}_{\phi,(c_i,c_j)}(\mathbf{I}) \odot o^{(T)}_{(c_i,c_j)}}{\max\left\{o^{(T)}_{(c_i,c_j)}, \epsilon\right\}}.
   \label{eq:final-map}   
\end{equation}
We perform the above computations for all $(c_i,c_j) \in \mathcal{C}^{\prime}$. Finally, our inclusion loss is computed as the sum of the $L_1$ norms of tensors $\mathbf{H}$ across all $(c_i,c_j) \in \mathcal{C}^{\prime}$ via 
\begin{equation}
l_{\text{inclusion}} = \sum_{(c_i,c_j) \in \mathcal{C}^{\prime}} \left\|\mathbf{H}_{(c_i,c_j)}(\mathbf{I})\right\|_1.
\label{eq:l-inclusion} 
\end{equation}

\section{Experiments}
\label{sec:exp}
\subsection{Datasets}

\PAR{Cityscapes.} Cityscapes~\cite{Cordts_2016_CVPR} is a driving dataset. The dataset has 19 classes and includes 2975 training, 500 validation, and 1525 test images with fine annotations and 20000 additional coarsely annotated training images. 

\PAR{ADE20K.} We have used the dataset ADE20K~\cite{zhou2019semantic}. There are 150 classes and diverse scenes with 1038 image-level labels. The dataset is divided into 20K, 2K, and 3K images for training, validation, and testing, respectively.

\PAR{ACDC.} The ACDC dataset~\cite{Sakaridis2021ACDC} consists of driving scenes under adverse conditions such as fog, night, rain, and snow. It has the same 19 classes as Cityscapes. The dataset is divided into 1600 training, 406 validation, and 2000 test images.

\subsection{Implementation Details}

The softmax maps $P_\phi(\mathbf{I})$ input to InSeIn are first resized to $C\times 256\times 256$. The kernel size of the max-pooling used is $3\times 3$ with a stride of 1 and a padding of 1 to preserve spatial dimensions and to check for any connected components not sharing the border pixels. Random sampling on $|\mathcal{C'}|$ is applied offline in each iteration to reduce the memory footprint. We report semantic segmentation performance using mean Intersection over Union (mIoU). The value of $\alpha$ is fixed to $10^{-9}$ for all examined datasets after a search in log-space in the interval $[10^{-10},10^{-5}]$ based on validation performance.

\PAR{Training setup.} For all networks on which we implement InSeIn, we have used a batch size of 4 for Cityscapes and ACDC and 8 for ADE20K. The training protocol and hyperparameters remain the same as those for the base networks on each respective dataset. For ACDC, we applied the same setup as the corresponding baselines applied on the Cityscapes dataset. We used mmsegmentation~\cite{mmseg2020} to implement InSeIn on the baseline network SegFormer-B4~\cite{xie2021segformer}. 

\begin{table}[tb]
    \centering
    \caption{\textbf{Comparison with state-of-the-art semantic segmentation methods on ADE20K, Cityscapes and ACDC.} *: multi-scale testing. Results are reported on the Cityscapes test set, the ADE20K validation set, and the ACDC test set.}
    \label{tab:results}   
    \resizebox{\linewidth}{!}{%
    \begin{tabular}{lcccc}
        \toprule
        \textbf{Method} & \textbf{Backbone} & \textbf{ADE20K} & \textbf{Cityscapes} & \textbf{ACDC} \\
        \midrule
        PSPNet~\cite{Zhao2017Pyramid} & ResNet50 & 44.4 & 78.5 & - \\ 
        DeepLabv2~\cite{ChenDeepLab2018} &ResNet-101 & - & 71.4 & 55.3 \\ 
        DeepLabv3+~\cite{deeplabv3plus2018} &ResNet-101 & 44.1 & 80.9 & 70.0 \\ 
        HRNet~\cite{WangSCJDZLMTWLX19} &ResNet & 43.1 & 80.9 & 75.0 \\ 
        \midrule
        OCRNet~\cite{YuanCW19OCRNet} & HRNetV2-W48 & 45.6* & 82.4* & 66.5 \\ 
        InSeIn w/ OCRNet (ours) & HRNetV2-W48& \textbf{46.8}* & \textbf{82.9}* & \textbf{67.6} \\ 
        \midrule
        SegFormer~\cite{xie2021segformer} & MiT-B4 & 50.3 & 82.2 & 67.1 \\ 
        InSeIn w/ SegFormer (ours) & MiT-B4 & \textbf{51.0} & \textbf{82.7} & \textbf{69.2} \\
        \midrule
        Mask2Former~\cite{Cheng2022Mask2Former} & Swin-L  & 56.1 & 83.5* & 77.2 \\ 
        Mask2Former w/ clDice loss~\cite{clDiceShit2021}  &Swin-L &- & 83.6 &-\\
        InSeIn w/ Mask2Former (ours) & Swin-L & \textbf{56.7} & \textbf{84.0*} & \textbf{77.8} \\
        \midrule
        OneFormer~\cite{jain2023oneformer} & ConvNeXt-XL  & 57.4 & 83.8 & 78.0 \\ 
        InSeIn w/ OneFormer (ours) & ConvNeXt-XL & \textbf{58.0} & \textbf{84.3} & \textbf{78.9} \\
        \midrule
        SegMAN~\cite{SegMAN} & SegMAN-B  & 52.6 & 83.6 & 77.5 \\ 
        InSeIn w/ SegMAN (ours) & SegMAN-B & \textbf{53.7} & \textbf{84.4} & \textbf{78.3} \\
        \bottomrule
    \end{tabular}}       
\end{table}

\begin{table}[tb]
\centering
\small
\caption{\textbf{Comparison to Beyond Pixels~\cite{Howlader2024beyondPixels} on the Cityscapes test set using two baselines, SegFormer and Mask2Former.} Beyond Pixels is also fully supervised, as InSeIn.}
\vspace{-3mm}
\label{tab:sota_plug_in_check}
    \begin{tabular}{lcccc}
    \toprule
    \textbf{Method} & mIoU (\%) \\
    \midrule
    Beyond Pixels w/ SegFormer & 82.5 \\ 
    InSeIn w/ SegFormer (ours) & \textbf{82.7} \\
    \midrule
    Beyond Pixels w/ Mask2Former & 83.7 \\ 
    InSeIn w/ Mask2Former (ours) & \textbf{84.0} \\
    \bottomrule
    \end{tabular}   
\end{table}

\begin{table*}[tb]
  \caption{\textbf{Comparison of class-level IoU of OCRNet, SegFormer, Mask2Former, OneFormer, and SegMAN to their InSeIn-upgraded versions on the Cityscapes validation set.} Training and evaluation are performed using the complete training and validation sets, respectively.}
  \vspace{-3mm}
  \label{table:supervised:all}
  \centering
  \setlength\tabcolsep{3pt}
  \resizebox{\linewidth}{!}{%
  \footnotesize
  \begin{tabular}{lcccccccccccccccccccc}
  \toprule
  \textbf{Method} & \ver{road} & \ver{sidew.} & \ver{build.} & \ver{wall} & \ver{fence} & \ver{pole} & \ver{light} & \ver{sign} & \ver{veget.} & \ver{terrain} & \ver{sky} & \ver{person} & \ver{rider} & \ver{car} & \ver{truck} & \ver{bus} & \ver{train} & \ver{motorc.} & \ver{bicycle} & mIoU\\
  \midrule
  OCRNet~\cite{YuanCW19OCRNet} & 98.2 & 86.8 & 93.1 & \textbf{65.1} & 63.6 & 68.4 & 74.3 & \textbf{81.2} & 92.8 & 63.9 & \textbf{95.5} & 84.3 & 66.4 & 95.2 & 77.6 & 88.7 & 83.1 & 69.2 & 79.0 &         80.3\\
  Ours & \textbf{98.3} & \textbf{86.9} & \textbf{93.4} & 64.4 & \textbf{64.2} & \textbf{68.5} & \textbf{74.8} & 81.0 & \textbf{92.9} & \textbf{64.9} & 95.4 & \textbf{84.6} & \textbf{67.9} & \textbf{95.4} & \textbf{86.1} & \textbf{90.8} & \textbf{83.2} & \textbf{72.3} & \textbf{80.0} &    \textbf{81.3}\\
  \midrule
  SegFormer~\cite{xie2021segformer} & 98.4 & 87.8 & 93.7 & 68.4 & 65.4 & \textbf{69.6} & \textbf{75.6} & \textbf{81.6} & 93.1 & 70.6 & \textbf{95.4} & 84.8 & 68.3 & 95.6 & 81.8 & 90.7 & 83.9 & 73.6 & \textbf{80.0} &    82.0\\
  Ours & \textbf{99.2} & \textbf{91.3} & \textbf{93.8} & \textbf{72.1} & \textbf{72.3}& 56.2 & 73.5 &79.3 & \textbf{93.5} & \textbf{74.0} & 93.8 & \textbf{85.0} & \textbf{74.8} & \textbf{95.7} & \textbf{90.8} & \textbf{95.2} & \textbf{84.7} & \textbf{83.3} & 79.1 & \textbf{83.6}\\
  \midrule
  Mask2Former~\cite{Cheng2022Mask2Former} & 98.6 & 88.7 & 94.0 & 66.7 & 69.6 & 71.6 & 76.2 & 84.3 & 93.4 & 68.4 & \textbf{95.7} & 86.1 & 70.7 & 96.2 & 89.7 & 92.7 & \textbf{84.5} & 74.6 & 81.5 &    83.3\\
  Mask2Former w/ clDice~\cite{clDiceShit2021} &98.8 &87.8 &94.1 &65.1 &63.5 &73.4 &80.1 &83.0 &94.1 &73.4 &95.8 &88.9 &76.8 &96.6 &82.0 &93.7 &90.4 &74.0 &79.9 &83.8\\
  Ours & \textbf{99.0} & \textbf{91.0} & \textbf{94.8} & \textbf{81.4} & \textbf{78.5}& \textbf{73.7} & \textbf{77.3} &\textbf{85.9} & \textbf{93.8} & \textbf{74.4} & 95.6 & \textbf{87.7} & \textbf{77.6} & \textbf{96.5} & \textbf{92.4} & \textbf{94.8} & 63.8 & \textbf{78.6} & \textbf{83.3} & \textbf{85.3}\\
  \midrule 
  OneFormer~\cite{jain2023oneformer} & 99.0 & 89.2 & 94.8 & 69.8 & 67.2 & 73.7 & 76.4 & 82.2 & 94.4 & 74.9 & 96.2 & 89.7 & 77.4 & 96.9 &
  81.4 & 95.1 & \textbf{91.9} & 76.1 & 80.8 & 	84.6\\
  Ours & 99.0 & \textbf{89.4} & \textbf{94.9} & \textbf{71.7} & \textbf{69.1} & \textbf{75.6} & \textbf{82.0} &
  \textbf{84.9} & \textbf{94.5} & \textbf{75.3} & \textbf{96.3} & \textbf{89.9} & \textbf{79.1} & 96.9 & \textbf{81.6} &
  \textbf{95.3} & 87.0 & \textbf{77.1} & \textbf{81.0} & \textbf{85.3}\\
  \midrule 
  SegMAN~\cite{SegMAN} & \textbf{98.9} & \textbf{88.4} & 94.3 & 65.2 & 65.9 & 72.8 & 79.5 & 82.9 & 94.3 & 74.3 & \textbf{96.1} & 88.6 & \textbf{75.9} & 96.6 & 79.3 & 93.8 & \textbf{91.5} & \textbf{74.8} & 79.7 & 83.8\\
  Ours & 98.8 & 88.3 & 94.3 & \textbf{67.9} & \textbf{67.8} & \textbf{73.5} & \textbf{80.6} & \textbf{83.9} & 94.3 & \textbf{74.4} & 96.0 & \textbf{89.2} & 75.8 & \textbf{96.8} & \textbf{83.6} & \textbf{94.1} & 91.2 & 74.0 & \textbf{80.0} & \textbf{84.5}\\
  \bottomrule
  \end{tabular}}
\end{table*}

\begin{table*}[tb]
  \caption{\textbf{Comparison of class-level IoU of OCRNet, SegFormer, Mask2Former, OneFormer, and SegMAN to their InSeIn-upgraded versions on the ACDC test set.} Training and evaluation are performed using the complete training and test sets, respectively.}
  \vspace{-3mm}
  \label{table:ACDC:comparison}
  \centering
  \setlength\tabcolsep{3pt}
  \footnotesize
  \resizebox{\linewidth}{!}{%
  \begin{tabular}{lcccccccccccccccccccc}
  \toprule
  \textbf{Method} & \ver{road} & \ver{sidew.} & \ver{build.} & \ver{wall} & \ver{fence} & \ver{pole} & \ver{light} & \ver{sign} & \ver{veget.} & \ver{terrain} & \ver{sky} & \ver{person} & \ver{rider} & \ver{car} & \ver{truck} & \ver{bus} & \ver{train} & \ver{motorc.} & \ver{bicycle} & mIoU\\
  \midrule
  OCRNet~\cite{YuanCW19OCRNet} & \textbf{93.4}& \textbf{77.2}& 87.1& \textbf{50.0}& \textbf{45.2}& 48.6& 60.4& 61.3& 85.4& \textbf{59.4}& \textbf{95.5}& 58.5& 40.5& \textbf{87.4}& \textbf{63.6}& \textbf{76.3}& 80.6& 41.6& 53.2& 66.5 \\
  Ours & 92.5& 71.6& \textbf{87.2}& 45.3& 39.9& \textbf{54.2}& \textbf{70.2}& \textbf{68.2}& \textbf{86.4}& 50.8& 94.8& \textbf{65.2}& \textbf{45.5}& 87.2& 60.9& 69.9& \textbf{84.5}& \textbf{50.6}& \textbf{59.9}& \textbf{67.6}\\
  \midrule
  SegFormer~\cite{xie2021segformer} & 94.6 & 78.3 & 88.4 & 51.6 & 47.4 & 48.0 & 62.1 & 61.2 & 87.0 & \textbf{66.8} & 95.6 & 59.8 & 39.0 & 87.0 & 63.1 & 73.8 & 79.1 & 38.4 & 52.7 &    67.1\\
  Ours & \textbf{94.8} & \textbf{79.5} & \textbf{88.5} & \textbf{54.2} & \textbf{50.4}& \textbf{48.7} & \textbf{63.7} &\textbf{64.2} & \textbf{87.9} & 66.2 & \textbf{95.9} & \textbf{61.2} & \textbf{42.0} & \textbf{88.2} & \textbf{70.8} & \textbf{77.9} & \textbf{84.1} & \textbf{42.3} & \textbf{54.6} & \textbf{69.2}\\
  \midrule
  Mask2Former~\cite{Cheng2022Mask2Former} & \textbf{96.5}& \textbf{84.7}& \textbf{93.2}& 64.7& 59.5& \textbf{72.0}& \textbf{80.6}& 78.9& \textbf{91.1}& \textbf{72.1}& 96.7& \textbf{77.7}& 44.4& \textbf{91.6}& 75.9& 71.0& \textbf{92.4}& 58.5& 65.9&  77.2\\
  Ours &96.4& 84.3& 92.9& \textbf{65.7}& \textbf{61.2}& 66.7& 78.7& \textbf{79.5}& 91.0& 71.9& \textbf{97.0}& 77.5& \textbf{56.6}& 91.3& \textbf{76.9}& \textbf{73.3}& 91.6& \textbf{58.9}& \textbf{67.0}& \textbf{77.8} \\
  \midrule
  OneFormer~\cite{jain2023oneformer} & 94.5 & 80.9 & 91.9 & \textbf{66.8} & 57.7 & 64.3 & \textbf{74.9} & 75.0 & 79.2 & 68.7 & 87.7 & 78.0 & 62.1 & 92.9 & \textbf{84.9} & \textbf{91.5} & \textbf{94.0} & \textbf{68.7} & 68.4 & 78.0\\
  Ours &\textbf{98.7} &\textbf{86.9} &\textbf{93.3} &58.8 &\textbf{60.3} &\textbf{65.7} &73.0 &\textbf{78.3} &\textbf{93.5} &\textbf{72.8} &\textbf{95.6} &\textbf{85.9} &\textbf{71.2} &\textbf{95.9} &73.3 &82.3 &69.5 &67.2 &\textbf{75.9} &\textbf{78.8}\\
  \midrule
  SegMAN~\cite{SegMAN} &94.5 &78.2 &92.0 & 61.9 & 55.0 & \textbf{64.8} & 73.7 &72.6 &\textbf{88.3} &67.3 &\textbf{95.3} &77.1 &60.2 &92.6 &84.0 &86.8 &92.4 &67.5 &\textbf{68.6} &77.5\\
  Ours &\textbf{94.6} & \textbf{80.7} &\textbf{92.1} &\textbf{69.5} &\textbf{57.9} &64.5 &\textbf{75.2} &\textbf{75.4} &79.2 &\textbf{68.5} &87.9 &\textbf{78.6} &\textbf{62.3} &\textbf{92.7} &\textbf{84.8} &\textbf{91.6} &\textbf{94.0} &\textbf{68.8} &68.5 &\textbf{78.3}\\
  \bottomrule
  \end{tabular}} 
\end{table*}

\begin{figure}[tb]
  \centering
  \subfloat{\includegraphics[width=0.25\linewidth]{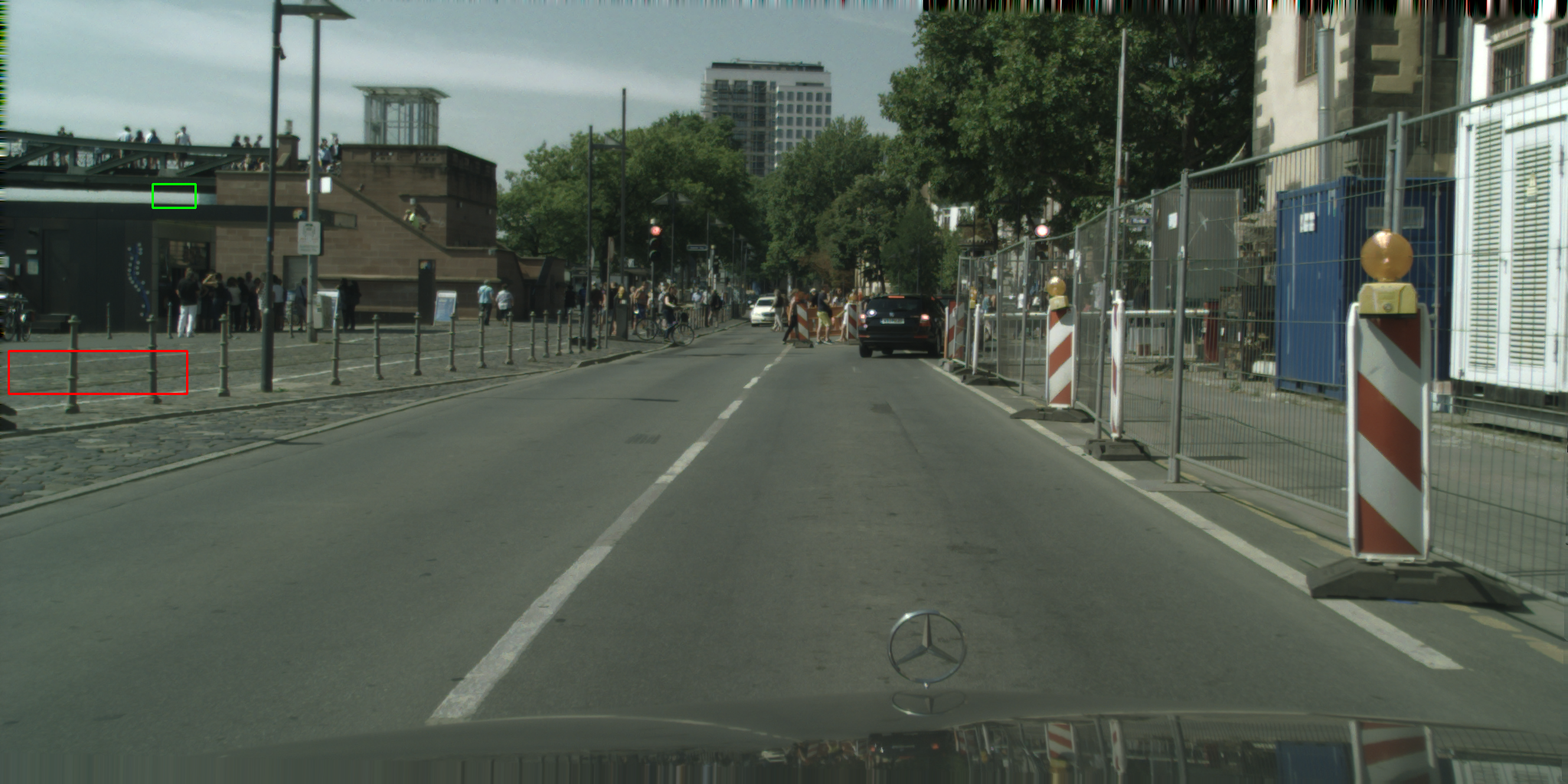}}
  \hfil
  \subfloat{\includegraphics[width=0.25\linewidth]{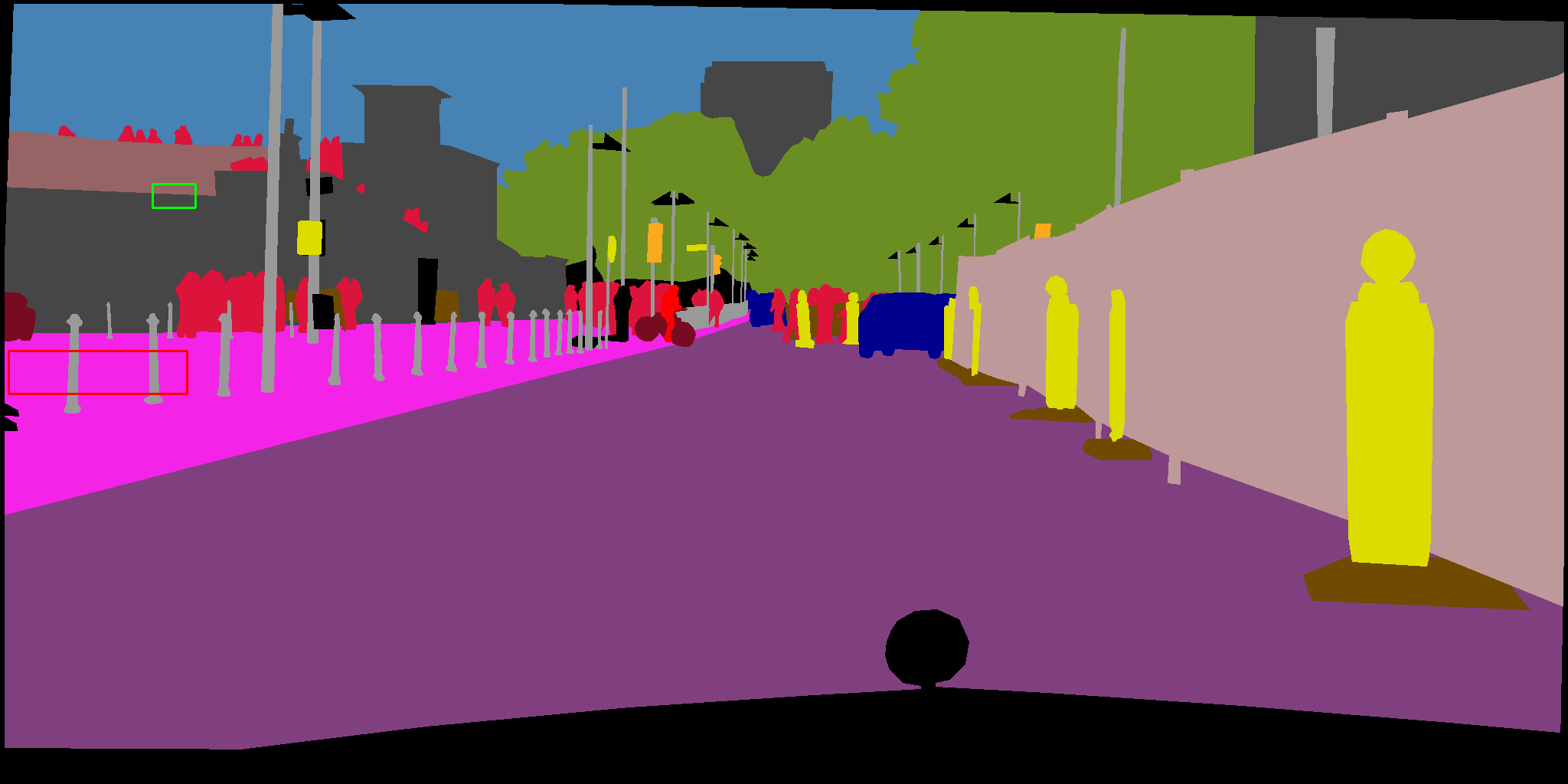}}
  \hfil
  \subfloat{\includegraphics[width=0.25\linewidth]{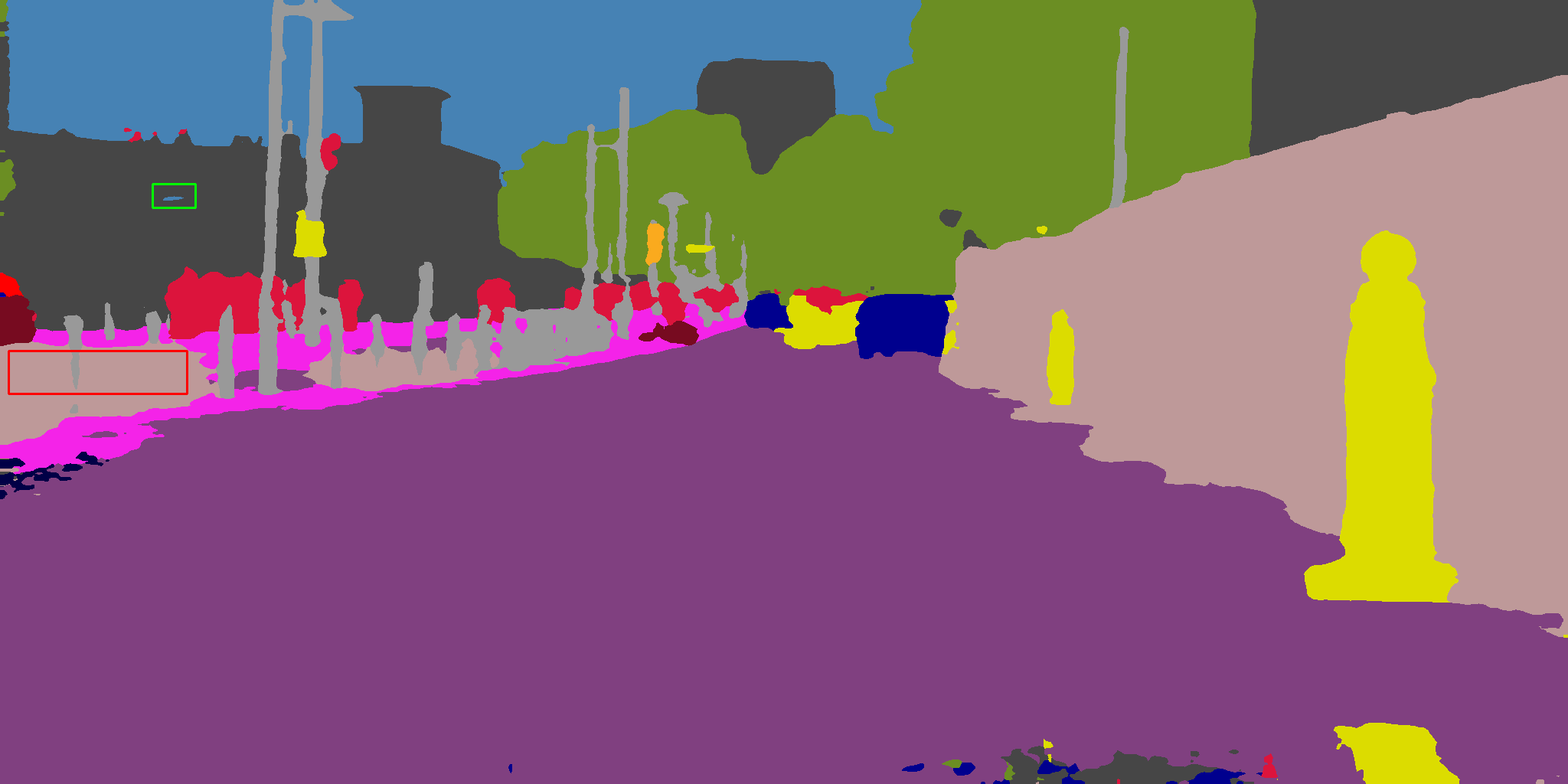}}
  \hfil
  \subfloat{\includegraphics[width=0.25\linewidth]{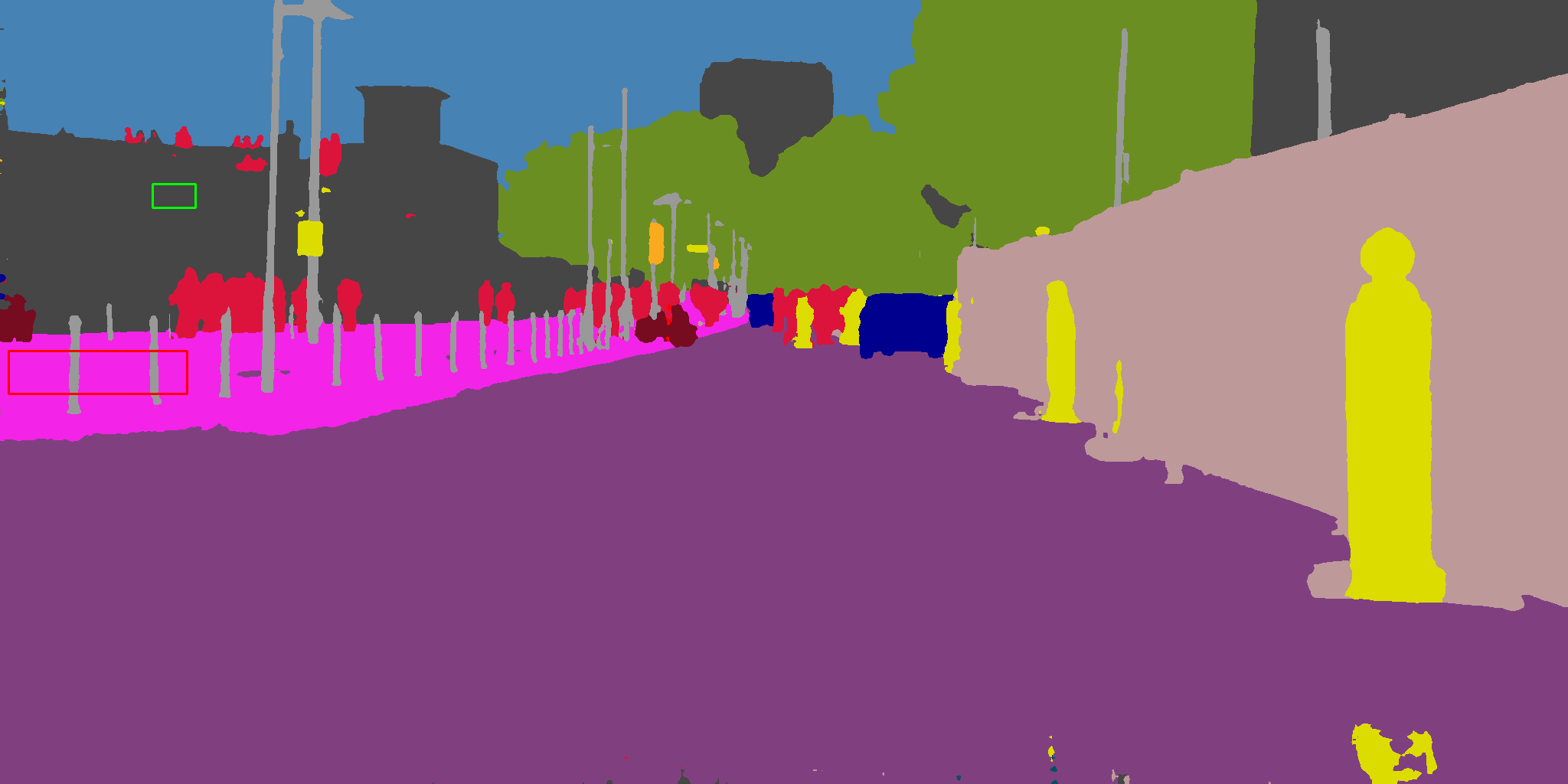}}
  \hfil
  \subfloat{\includegraphics[width=0.25\linewidth]{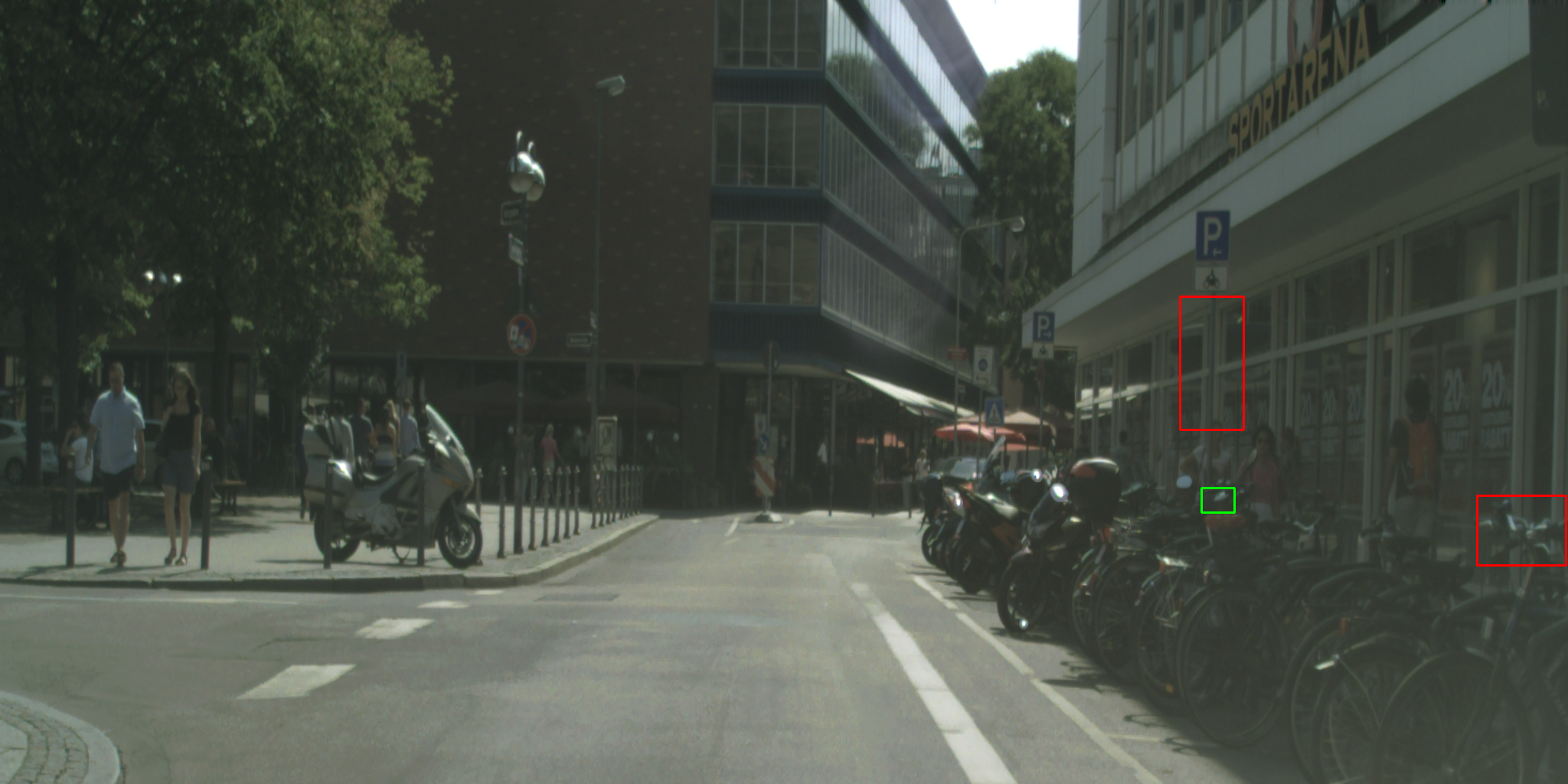}}
  \hfil
  \subfloat{\includegraphics[width=0.25\linewidth]{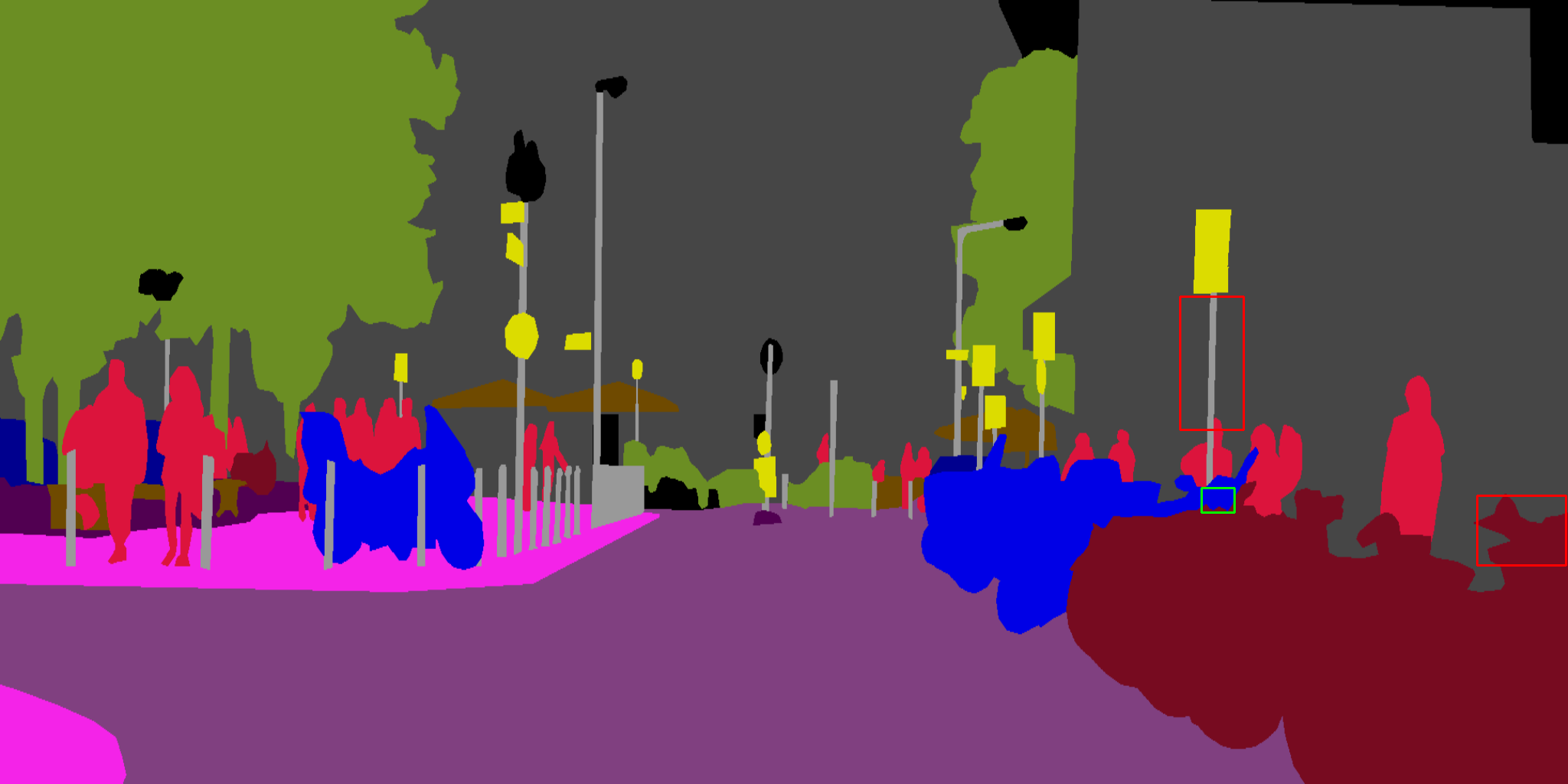}}
  \hfil
  \subfloat{\includegraphics[width=0.25\linewidth]{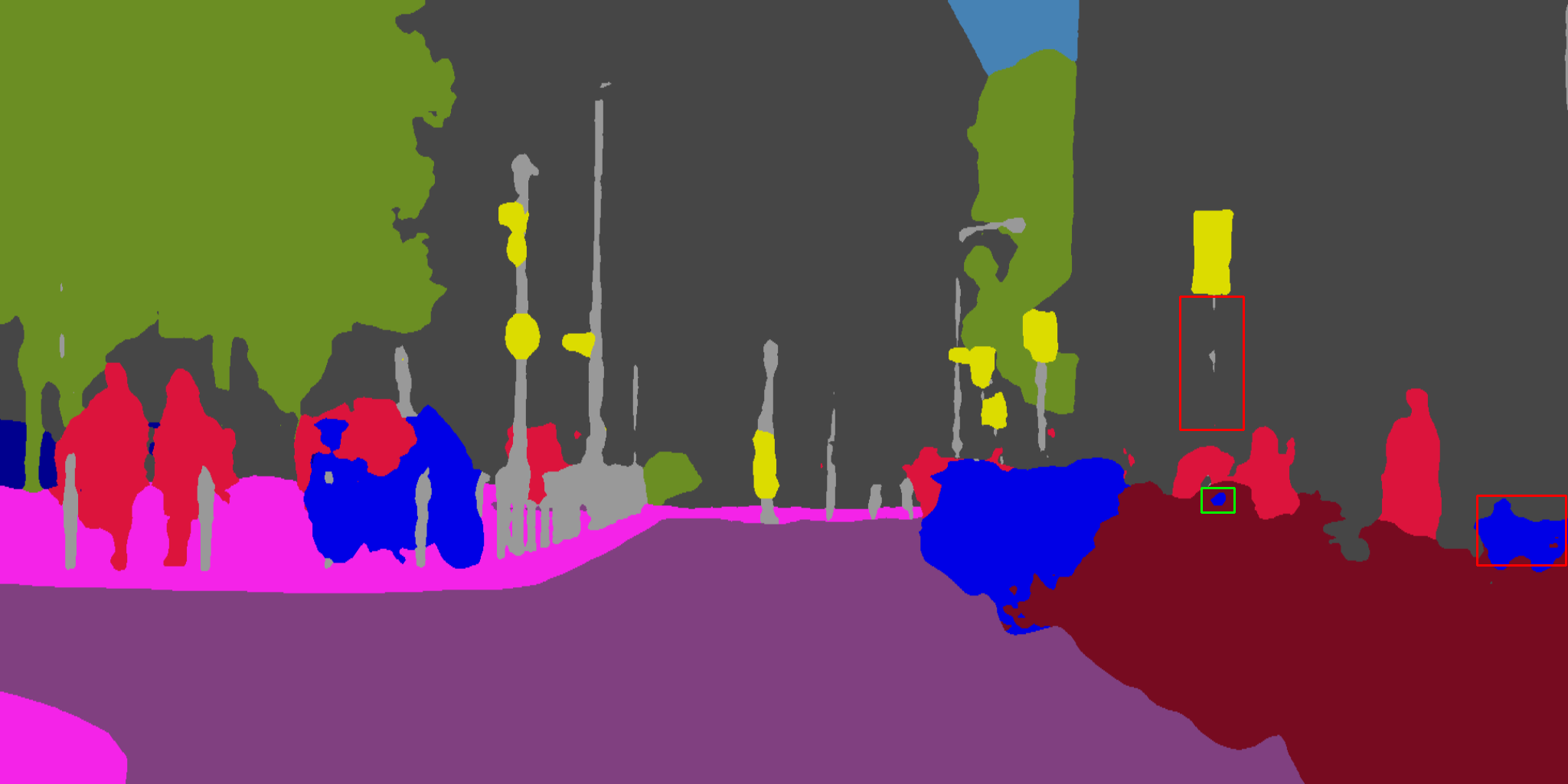}}
  \hfil
  \subfloat{\includegraphics[width=0.25\linewidth]{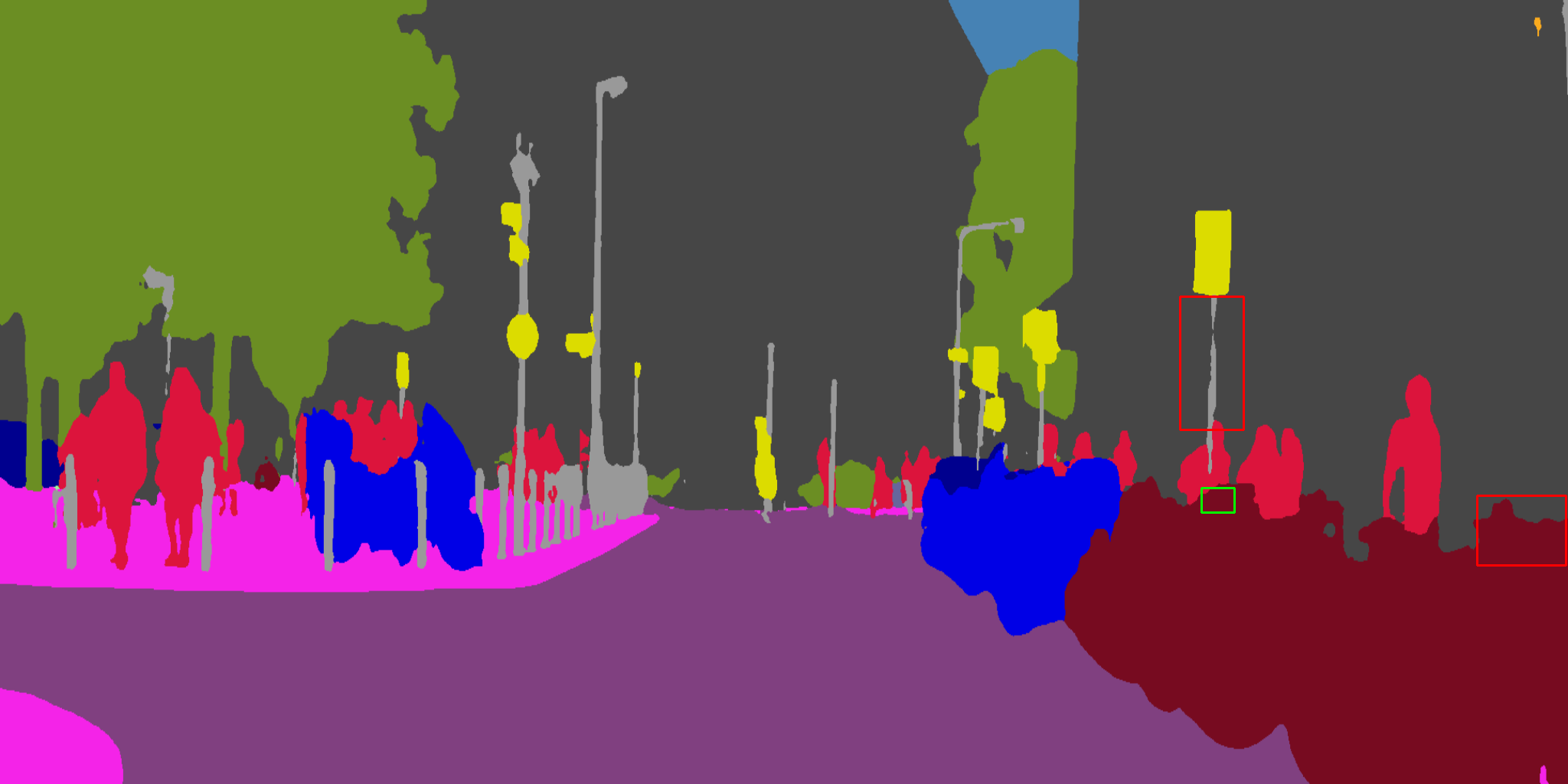}}
  \caption{\textbf{Qualitative comparison on Cityscapes. Best viewed on a screen with full zoom.} From left to right: input image, ground-truth semantic labels, prediction of Mask2Former~\cite{Cheng2022Mask2Former} network and InSeIn.}
  \label{fig:discontinued_class_1}
  \vspace{-5mm}
\end{figure}

\subsection{Comparison to The State of The Art}

Tab.~\ref{tab:results} compares InSeIn with a wide range of state-of-the-art methods for semantic segmentation on Cityscapes~\cite{Cordts_2016_CVPR}, ADE20K~\cite{zhou2019semantic}, and ACDC~\cite{Sakaridis2021ACDC}. Overall, InSeIn achieves significant improvements in mIoU across all five network architectures on which we have implemented it and across all three datasets. On Cityscapes, we observe a consistent $0.5$--$0.8$\% improvement in mIoU with InSeIn on all architectures. We have also included a comparison of our method with another light-weight morphological loss clDice~\cite{clDiceShit2021}, with Mask2Former on Cityscapes. Our method beats the loss. On ADE20K, InSeIn achieves $1.2$\% improvement over the baseline OCRNet~\cite{YuanCW19OCRNet} and $0.7$\% over the baseline SegFormer-B4. On ACDC, InSeIn achieves $1.1$\% improvement over the baseline OCRNet and $2.1$\% over the baseline SegFormer-B4.
Overall, improvements with InSeIn are consistent across all examined network architectures. This evidence shows that InSeIn can improve the segmentation performance of a general baseline network on which it is applied.

\begin{figure}[tb]
  \centering
  \subfloat{\includegraphics[width=0.25\linewidth]{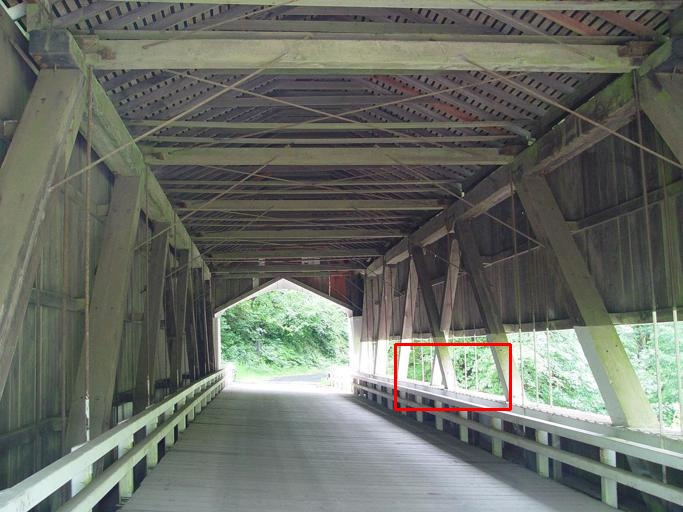}}
  \hfil
  \subfloat{\includegraphics[width=0.25\linewidth]{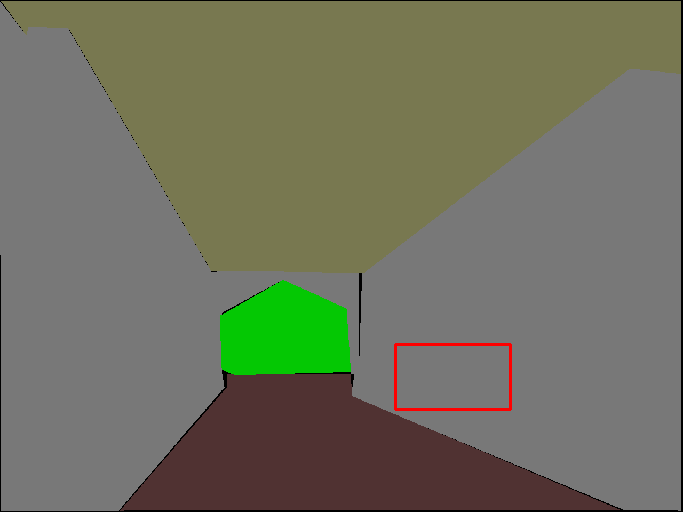}}
  \hfil
  \subfloat{\includegraphics[width=0.25\linewidth]{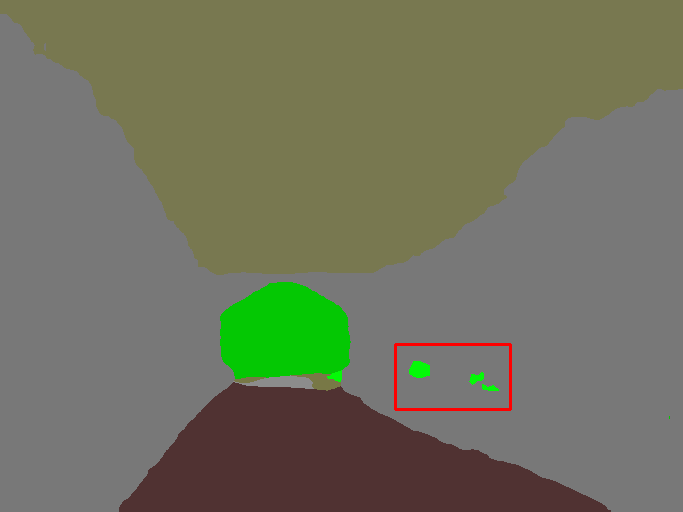}}
  \hfil
  \subfloat{\includegraphics[width=0.25\linewidth]{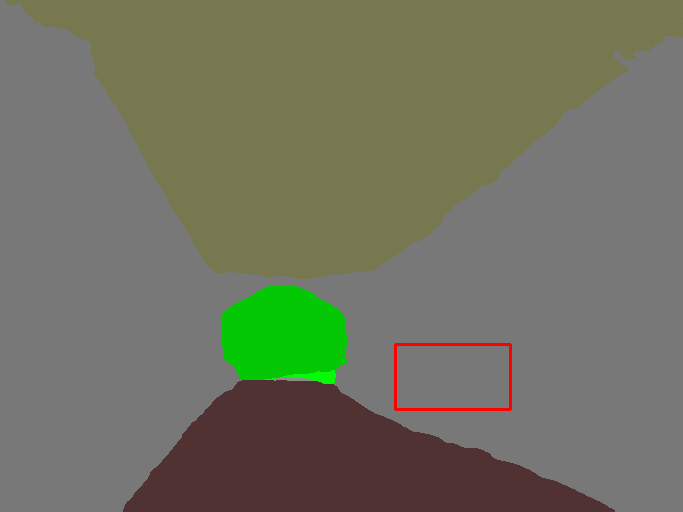}}
  \hfil
    \subfloat{\includegraphics[width=0.25\linewidth]{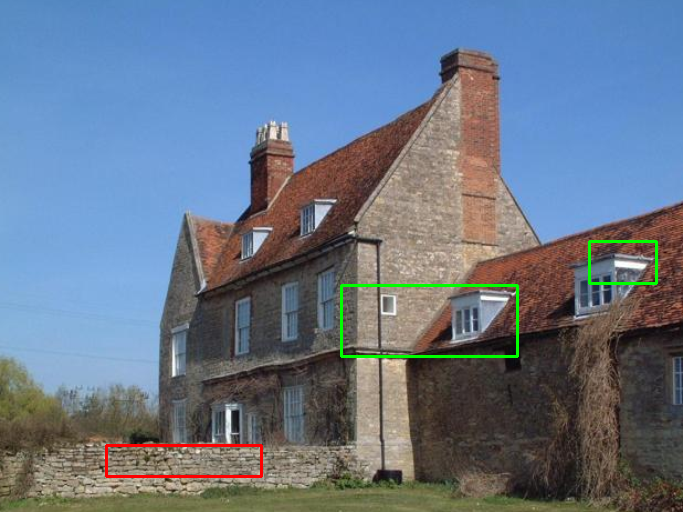}}
  \hfil
  \subfloat{\includegraphics[width=0.25\linewidth]{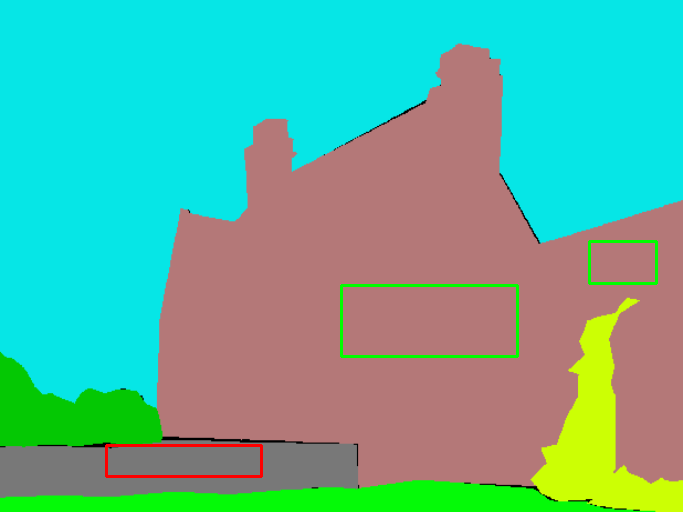}}
  \hfil
  \subfloat{\includegraphics[width=0.25\linewidth]{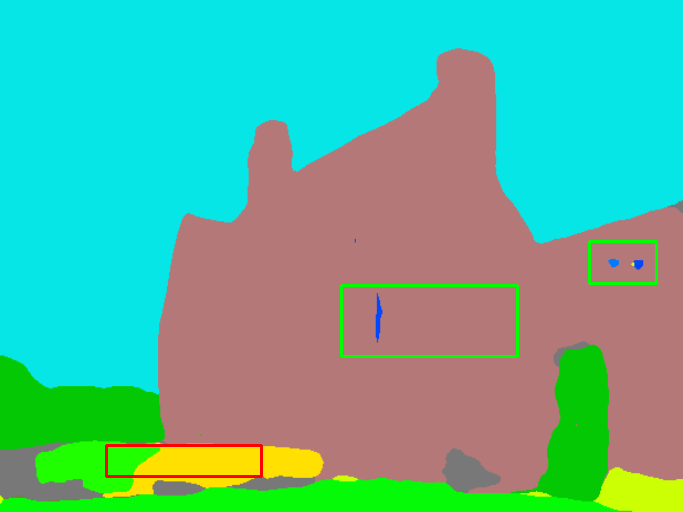}}
  \hfil
  \subfloat{\includegraphics[width=0.25\linewidth]{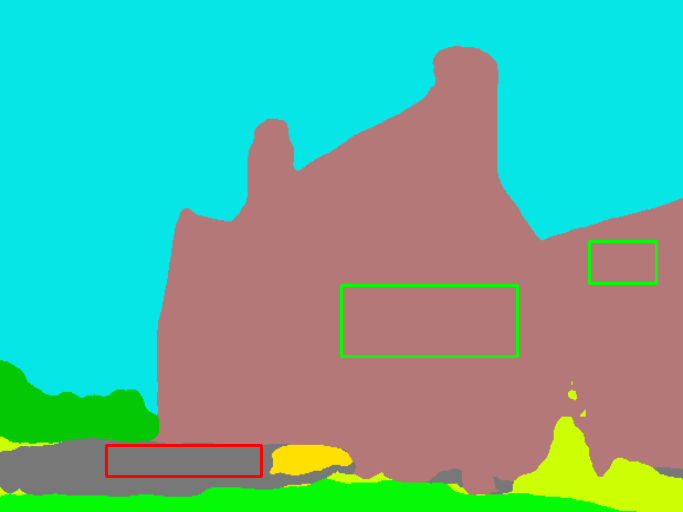}}
  \vspace{-2mm}
  \caption{\textbf{Qualitative comparison on ADE20K. Best viewed on a screen with full zoom.} From left to right: input image, ground-truth semantic labels, and predictions of OCRNet~\cite{YuanCW19OCRNet} and InSeIn.}
  \label{fig:discontinued_class_5}  
\end{figure}

Tab.~\ref{tab:sota_plug_in_check} presents a comparison of InSeIn with a recent competing plug-and-play model, Beyond Pixels~\cite{Howlader2024beyondPixels}, on the Cityscapes dataset, implemented on two of our examined baseline networks, namely SegFormer and Mask2Former, for fully supervised segmentation. InSeIn consistently outperforms Beyond Pixels.
Tab.~\ref{table:supervised:all} shows the class-wise comparison of IoU on the Cityscapes validation set for InSeIn and the corresponding baselines. We observe that InSeIn achieves a consistent improvement in IoU scores across most of the 19 classes. InSeIn with OCRNet~\cite{YuanCW19OCRNet}, SegMAN~\cite{SegMAN}, and SegFormer~\cite{xie2021segformer} show improvement in small challenging classes of the Cityscapes dataset. While we can observe an all-around and consistent improvement with Mask2Former~\cite{Cheng2022Mask2Former} and OneFormer~\cite{jain2023oneformer}. InSeIn also shows improvement over the baseline Mask2Former with another morphological loss, clDice~\cite{clDiceShit2021}.
Tab.~\ref{table:ACDC:comparison} shows the class-wise comparison of IoU on the ACDC test set for InSeIn and the corresponding baselines. We also observe that InSeIn achieves an overall improvement in IoU scores across the classes, especially over the SegFormer baseline. InSeIn achieves significant gains across all baselines for challenging small-instance classes such as motorcycle, bicycle, and rider.

\begin{figure}[tb]
  \centering
  \subfloat{\includegraphics[width=0.25\linewidth]{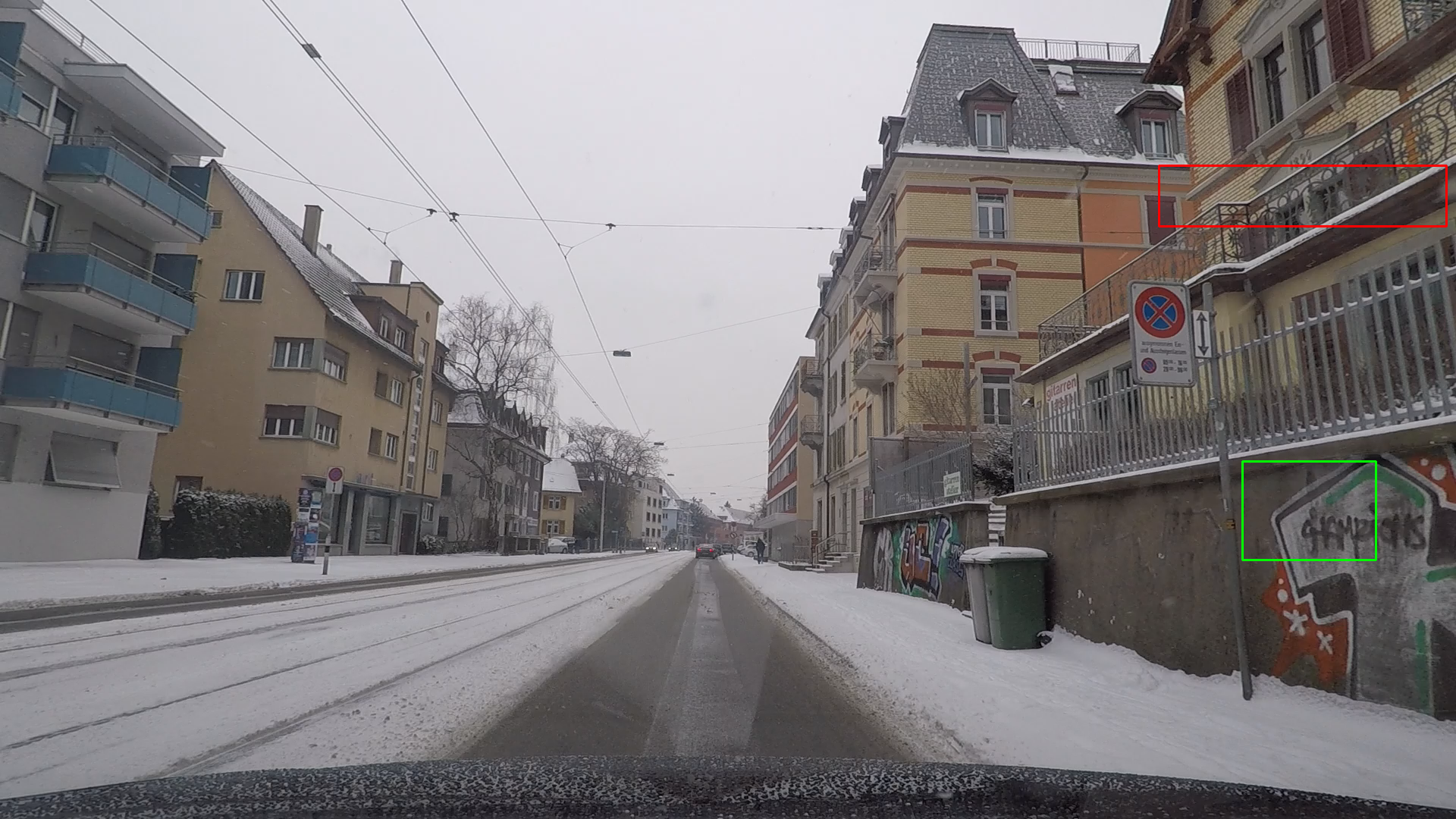}}
  \hfil
  \subfloat{\includegraphics[width=0.25\linewidth]{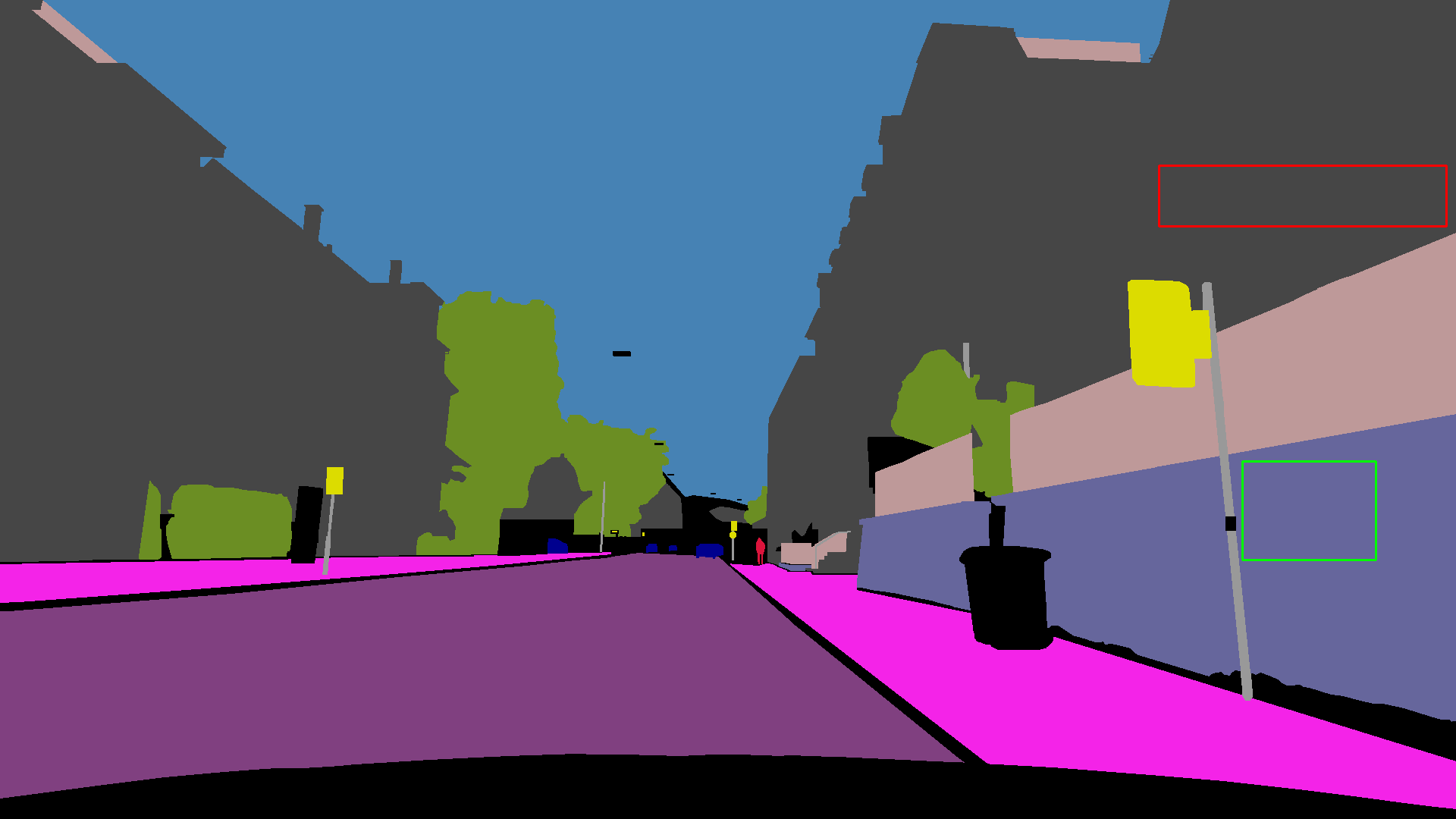}}
  \hfil
  \subfloat{\includegraphics[width=0.25\linewidth]{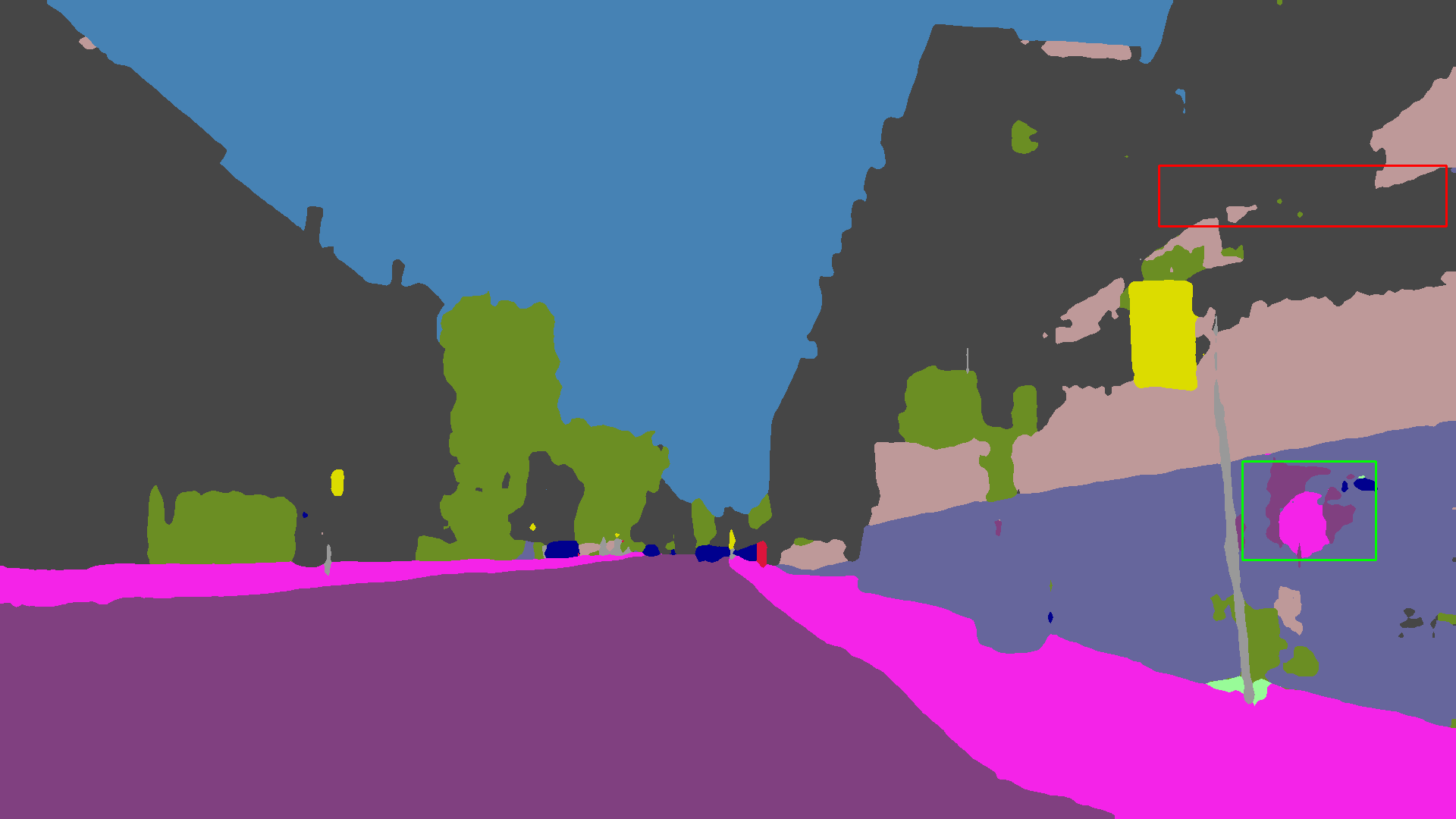}}
  \hfil
  \subfloat{\includegraphics[width=0.25\linewidth]{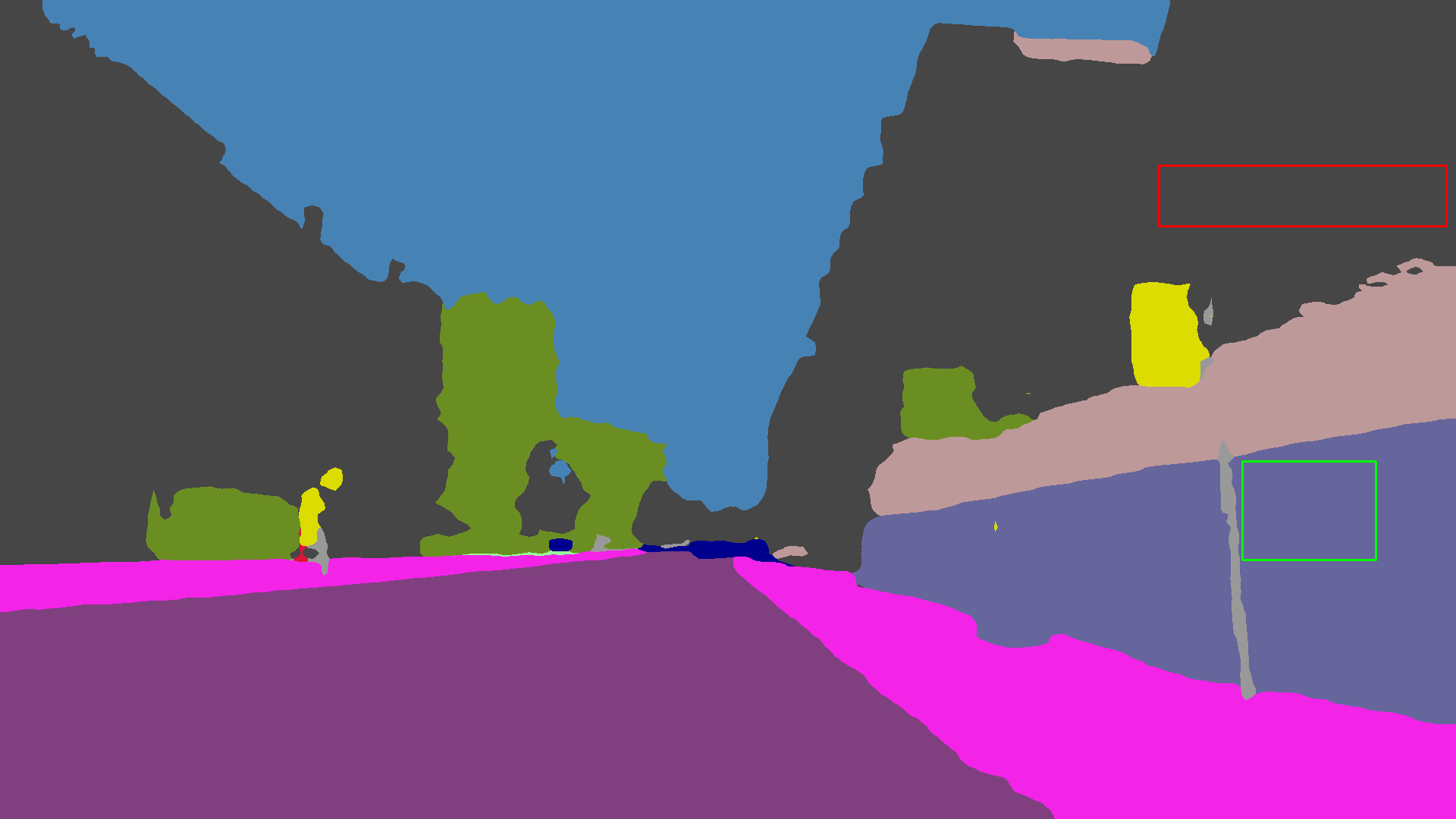}}
  \hfil
  \subfloat{\includegraphics[width=0.25\linewidth]{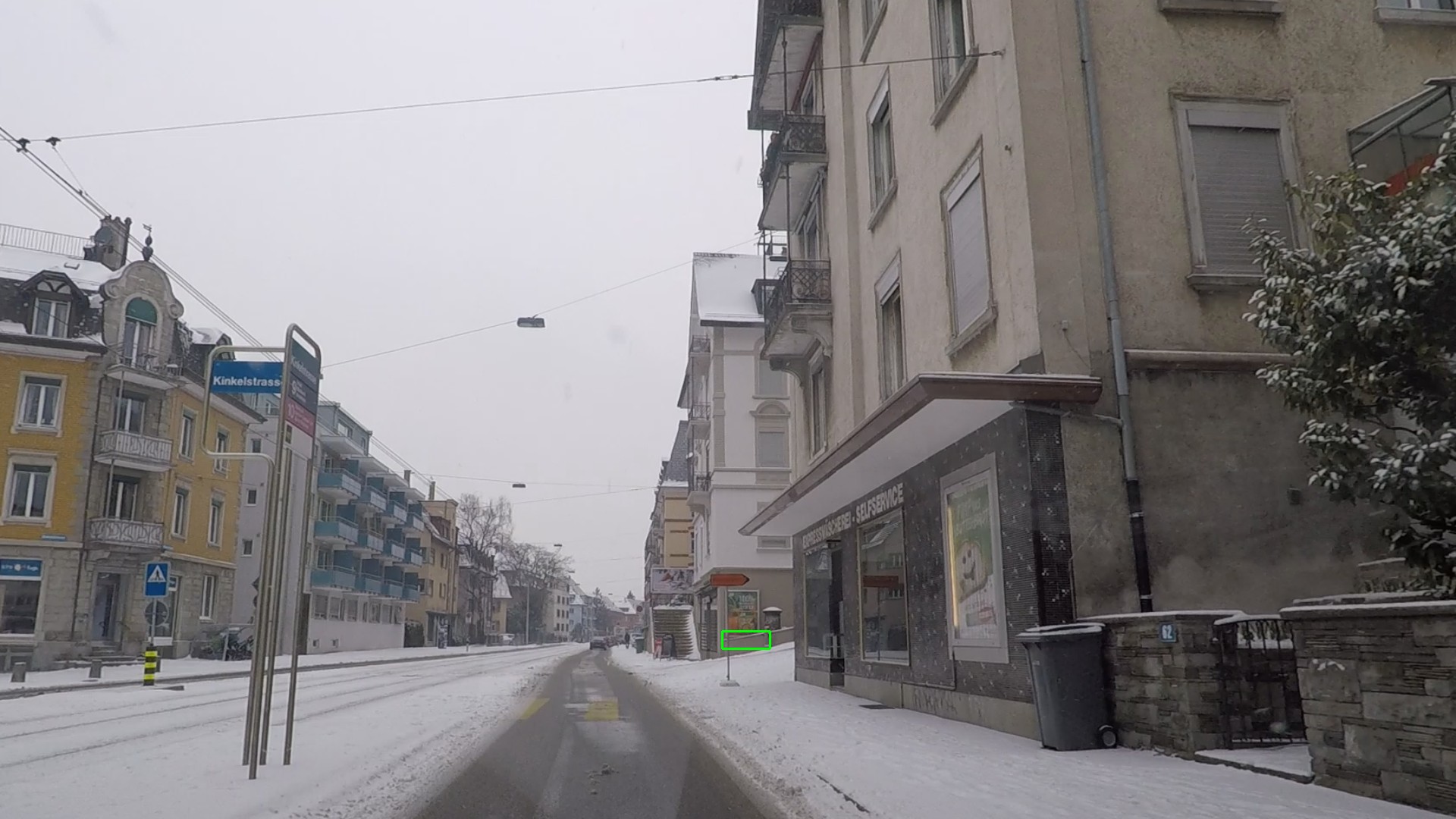}}
  \hfil
  \subfloat{\includegraphics[width=0.25\linewidth]{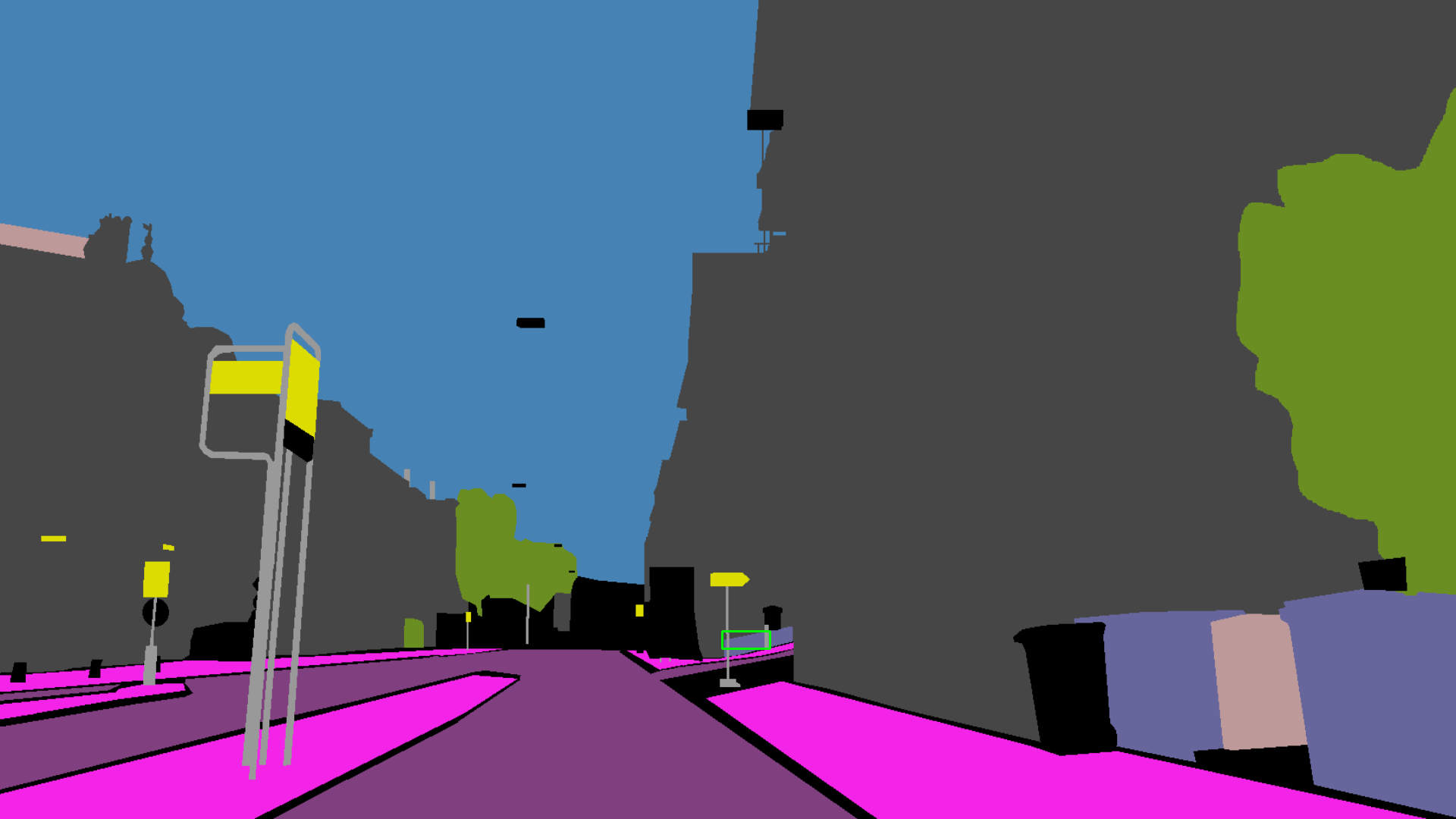}}
  \hfil
  \subfloat{\includegraphics[width=0.25\linewidth]{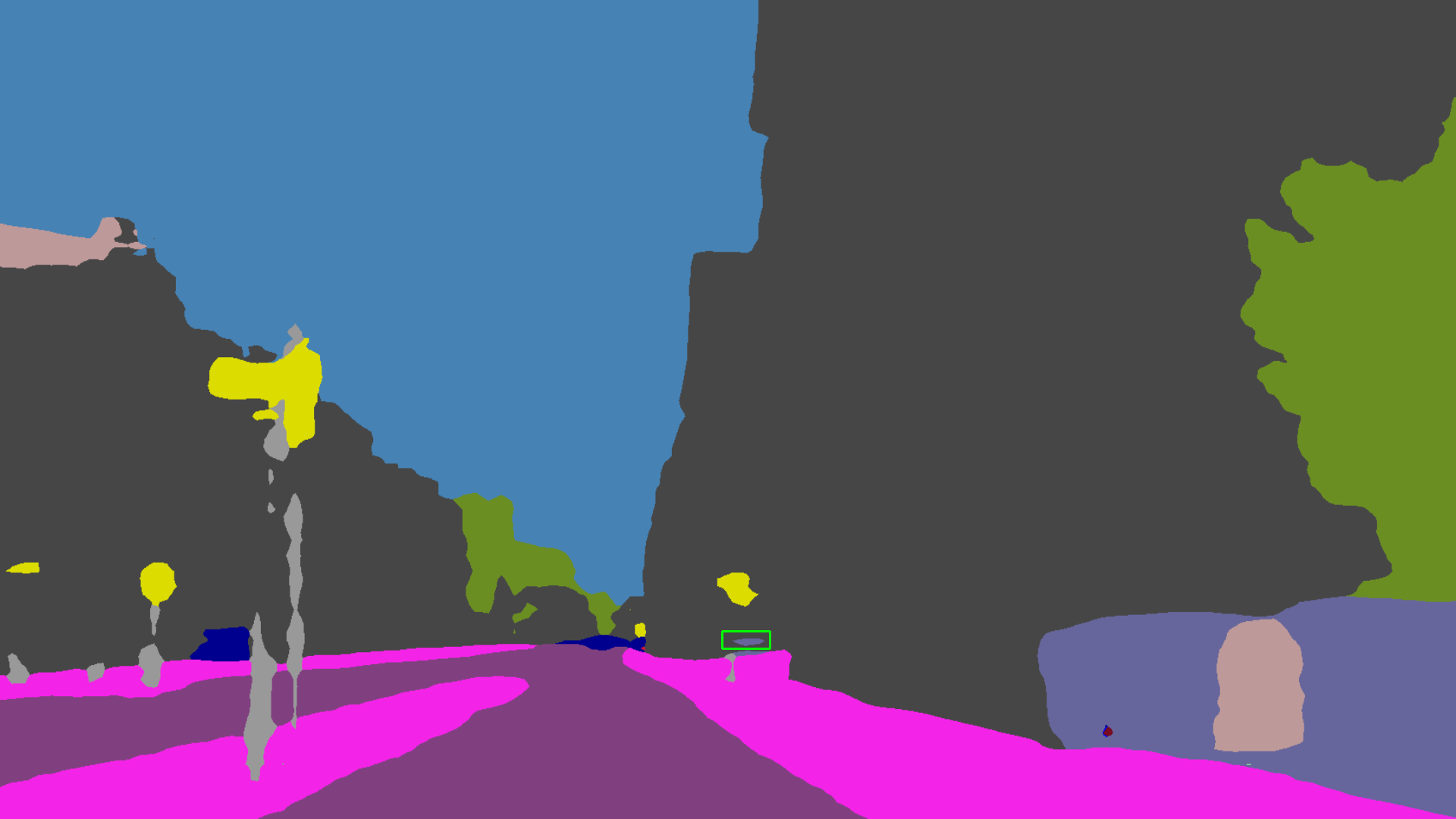}}
  \hfil
  \subfloat{\includegraphics[width=0.25\linewidth]{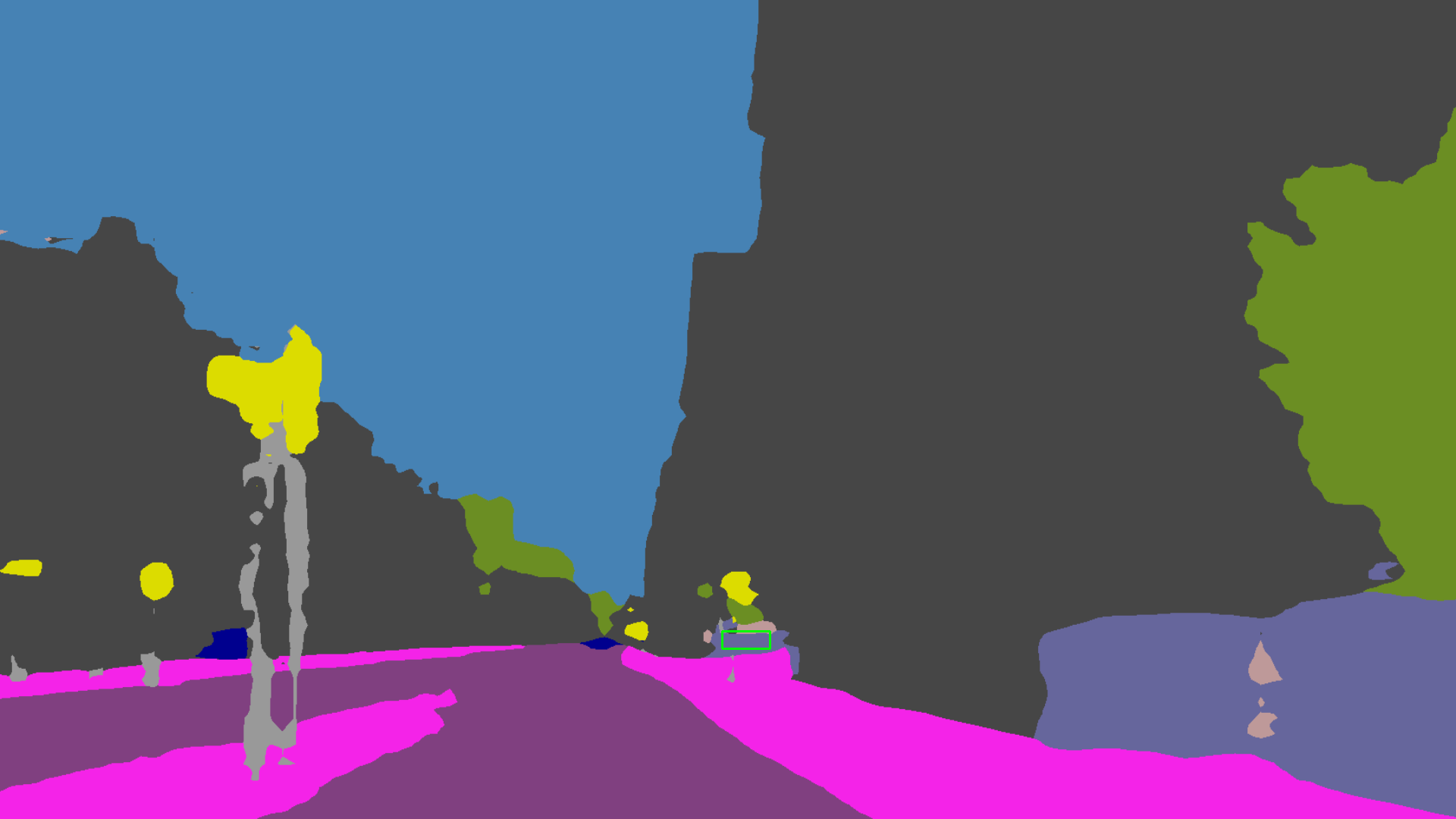}}
  \caption{\textbf{Qualitative comparison on ACDC. Best viewed on a screen at full zoom.} From left to right: input image, ground-truth semantic labels, and predictions of SegFormer~\cite{xie2021segformer} network and InSeIn.}
  \label{fig:discontinued_class_7}
  \vspace{-5mm}
\end{figure}

\begin{table}[tb]
\centering
\caption{\textbf{Comparison for domain shift} Mask2Former and Mask2Former+InSeIn trained on Cityscapes~\cite{Cordts_2016_CVPR} and evaluated on ACDC~\cite{Sakaridis2021ACDC}.}
\vspace{-3mm}
\label{tab:domain_shift_comparison}
    \begin{tabular}{lcccc}
    \toprule
    \textbf{Method} & mIoU (\%) \\
    \midrule
    Mask2Former~\cite{Cheng2022Mask2Former} & 62.2 \\ 
    InSeIn w/ Mask2Former (ours) & \textbf{63.6} \\
    \bottomrule
    \end{tabular} 
\vspace{-5mm}    
\end{table}

\PAR{Qualitative results on Cityscapes.} In Fig.~\ref{fig:discontinued_class_1}, in the first row, baseline prediction, a building segment infeasibly includes a sky segment marked with \emph{green} boxes, InSeIn rectifies it and also better segments the sidewalk marked with \emph{red} boxes. The second row, baseline prediction, a bicycle infeasibly includes a motorcycle segment marked with \emph{green} boxes on the right of the image. InSeIn rectifies this erroneous motorcycle segment. 

\PAR{Qualitative results on ADE20K.} In Fig.~\ref{fig:discontinued_class_5}, in the first row, baseline prediction, the bridge segment includes a vegetation segment marked with \emph{red} boxes, while InSeIn successfully rectifies such erroneous prediction. The second row, baseline prediction, building segment infeasibly includes a segment of the sea marked with \emph{green} boxes. InSeIn rectifies it completely, and the \emph{red} bounding boxes show additional parts of the prediction which are improved by InSeIn.

\PAR{Qualitative results on ACDC.} In Fig.~\ref{fig:discontinued_class_7}, in the first row, baseline prediction, wall segment infeasibly includes a sidewalk segment marked with \emph{green} boxes.InSeIn successfully rectifies the infeasibility. The second row, baseline prediction, building segment infeasibly includes a wall segment marked with \emph{green} boxes.InSeIn successfully predicts a wall label in that region. In both examples, \emph{red} boxes highlight miscellaneous parts of the predictions where InSeIn improves upon baseline.

\subsection{Evaluation of Prediction Feasibility}

Here we demonstrate the efficacy of our model. The complete per-dataset counts of the feasible and infeasible inclusion pairs are reported in Tab.~\ref{tab:inclusion:statistics}. Algorithm~\ref{alg:check_feasible_inclusion_constraint} is applied to the training set of every dataset, even when we train the networks on the union of the training and validation sets.

We define a class-level metric to measure how well predictions conform to our inclusion constraint, which we term mean infeasibility normalized frequency (mINF), as
\vspace{-1mm}
\begin{equation}
\centering
    \text{mINF} = \frac{1}{|\mathcal{C}^\prime|} \sum_{(c_i, c_j) \in \mathcal{C}^\prime} \frac{f_{\text{inclusion}}(c_i, c_j)}{f_{\text{co-occur}}(c_i, c_j)}.
\label{eq:minfeasiblity_check}
\vspace{-2mm}
\end{equation}

\begin{table}[tb]
    \centering
    \footnotesize
    \caption{\textbf{Feasible and infeasible inclusion pair statistics for different datasets.} The statistics are computed on the training set of each dataset. Infeasible pair counts exclude pairs that do not co-occur in any image of the dataset.}
    \vspace{-3mm}
    \label{tab:inclusion:statistics}
    \begin{tabular}{lccccc}
        \toprule
        \textbf{Dataset} \textbackslash{} \textbf{Type of class pair} & \textbf{Feasible} & \textbf{Infeasible} \\
        \midrule
        Cityscapes (19 classes) & 162 & 180 \\ 
        ADE20K (150 classes) & 11070 & 11280 \\ 
        ACDC (19 classes) & 118 & 224 \\ 
        \bottomrule
    \end{tabular}    
    \vspace{-2mm}    
\end{table}

\begin{table}[tb]
\centering
\footnotesize
\caption{\textbf{Comparison of prediction feasibility with state-of-the-art models on Cityscapes and ACDC.} Results are reported with our mINF metric (\%, lower is better) on the Cityscapes and ACDC test sets.}
\label{tab:mean_infeasiblity_check}
\resizebox{\linewidth}{!}{%
    \begin{tabular}{lcccc}
    \toprule
    \textbf{Method} \textbackslash{} \textbf{Dataset} & \textbf{Cityscapes} & \textbf{ACDC} \\
    \midrule
    SegFormer~\cite{xie2021segformer} & 4.0 & 8.0 \\ 
    InSeIn w/ SegFormer (ours) & \textbf{0.1} & \textbf{0.2} \\
    \midrule
    Mask2Former~\cite{Cheng2022Mask2Former} & 3.2 & 7.3 \\ 
    InSeIn w/ Mask2Former (ours) & \textbf{0.1} & \textbf{0.1} \\
    \midrule
    OneFormer~\cite{jain2023oneformer} & 3.0 & 7.1 \\ 
    InSeIn w/ OneFormer (ours) & \textbf{0.1} & \textbf{0.1} \\
    \bottomrule
    \end{tabular}}
\vspace{-2mm}    
\end{table}

\begin{table}[tb]
\centering
\footnotesize
\caption{\textbf{False response error (\%, lower is better) on the Cityscapes and ACDC validation sets.} We have set boundary regions to 15 pixels.}
\vspace{-3mm}
\label{tab:false_response_error_check}
    \begin{tabular}{lcccc}
    \toprule
    \textbf{Method} \textbackslash{} \textbf{Dataset} & \textbf{Cityscapes} & \textbf{ACDC} \\
    \midrule
    SegFormer~\cite{xie2021segformer} & 23.7 & 37.6 \\ 
    InSeIn w/ SegFormer (ours) & \textbf{22.2} & \textbf{35.9} \\
    \midrule
    Mask2Former~\cite{Cheng2022Mask2Former} & 22.0 & 29.3 \\ 
    InSeIn w/ Mask2Former (ours) & \textbf{20.8} & \textbf{28.0} \\
    \midrule
    OneFormer~\cite{jain2023oneformer} & 21.7 & 28.3 \\ 
    InSeIn w/ OneFormer (ours) & \textbf{19.9} & \textbf{26.8} \\
    \bottomrule
    \end{tabular}
    \vspace{-5mm}
\end{table}

We compute $f_{\text{co-occur}}(c_i, c_j)$, i.e., the frequency of co-occurrence of the classes in the infeasible class pair $(c_i,c_j)$ in the set of predicted labelings on the evaluation set. $f_{\text{inclusion}}(c_i, c_j)$ is the frequency of the occurrence of infeasible inclusions for class pair $(c_i, c_j)$ in the set of predicted labelings on the evaluation set. An mINF of 0 signifies the complete absence of infeasible inclusions in the predicted labelings. We report mINF scores in Tab.~\ref{tab:mean_infeasiblity_check} for SegFormer, Mask2Former, and OneFormer models on Cityscapes and ACDC.
These results show that InSeIn can almost fully adhere to the inclusion constraint imposed by our inclusion loss.

We have also computed another pixel-level metric, namely the false response error~\cite{chen2024semantic,zhao2025bfanet}, with results for three representative network architectures shown in Tab.~\ref{tab:false_response_error_check}. Lower values imply that the prediction feature maps contain less erroneous pixels. We observe a significant improvement with InSeIn over the baseline predictions across all three examined architectures.

\PAR{Limitation and discussion.} We conduct an ``oracle'' experiment, by manually filtering with a conceptual process the original set of infeasible pairs extracted from Cityscapes, which contains 180 pairs. After this filtering, the number of infeasible pairs becomes 120. The mIoU on the Cityscapes test set of the InSeIn model trained with this updated set of infeasible pairs based on Mask2Former architecture is 83.9\%, which is on a par with the original score presented in Tab.~\ref{tab:results}. This confirms that our data-driven constraint extraction already provides a meaningful set of constraints for moderate-size datasets like Cityscapes, with minimal to no impact from potential spurious infeasibility constraints that are included at our loss due to the rare occurrence of the respective configurations in practice. But like all data-driven architectures, our model is also dependent on the taxonomy of the dataset. So, our model is also subjected to the noise of the dataset.

\PAR{Evaluation on Domain Shift}
We evaluated our method with respect to domain shift in adverse weather conditions. We took the baseline Mask2Former~\cite{Cheng2022Mask2Former} trained on Cityscapes~\cite{Cordts_2016_CVPR} and evaluated on our ACDC~\cite{Sakaridis2021ACDC} dataset and followed the same procedure with Mask2Former w/ InSeIn, whose results are shown in Tab~\ref{tab:domain_shift_comparison}. It shows our method has achieved significant improvement, thereby showing robustness to domain shift.

\subsection{Ablation Study}

Tab.~\ref{fig:ablation_study_cityscape} ablates InSeIn on Cityscapes using Mask2Former as the baseline. We report validation set performance. We adjust the padding appropriately to preserve spatial dimensions across dilation iterations.  We observe that by increasing the kernel size with a decrease in the number of iterations for max-pooling, performance tends to decrease. The reason behind this decrease is that a bigger structuring element, i.e., the max-pooling kernel, enables the propagation of the border pixels to the interior portion of the feature map, which should otherwise be disconnected, and as a result, the infeasible included segment is not opened correctly. Thus, inclusion loss will assume too high a value, penalizing feasible segments.

\section{Conclusion and Future Work}
\begin{table}[tb]
    \centering
    \small
    \caption{\textbf{Ablation study of InSeIn with Mask2Former-SwinB on the Cityscapes validation set.} Number of iterations used for max-pooling kernels $5\times 5$ and $7\times7$ is 60k.}
    \vspace{-3mm}
    \label{fig:ablation_study_cityscape}
    \begin{tabular}{lcccccl}
        \toprule
        \textbf{Kernel-size} & \textbf{Stride} & \textbf{Padding} & $T$ & \textbf{mIoU} \\
        \midrule
        $3\times 3$ & 1 & 1 & 128 & 85.3 \\ 
        $5\times 5$ & 1 & 2 & 60 & 81.8 \\  
        $7\times 7$ & 1 & 3 & 3 & 79.8 \\  
        \bottomrule
    \end{tabular}  
    \vspace{-5mm}
\end{table}

In this paper, rather than building a new semantic segmentation architecture, we have attempted to enhance any existing SOTA network by enforcing a data-driven high-level constraint into the network. InSeIn constitutes a light-weight, plug-and-play module for semantic segmentation, which can be coupled with any network architecture to improve accuracy and conformity to our exemplary inclusion prior. InSeIn effectively achieves structurally feasible semantic segmentation, both qualitatively and quantitatively.

{
    \small
    \bibliographystyle{ieeenat_fullname}
    \bibliography{main}
}

\appendix
\setcounter{figure}{6}
\setcounter{section}{0}
\setcounter{table}{9}

\maketitlesupplementary

\section{Additional Qualitative Comparisons}

\begin{figure*}[tb]
  \centering
  \subfloat{\includegraphics[width=0.25\linewidth]{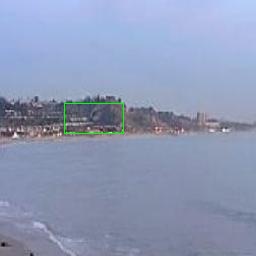}}
  \hfil
  \subfloat{\includegraphics[width=0.25\linewidth]{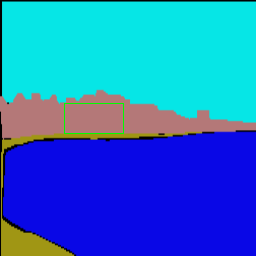}}
  \hfil
  \subfloat{\includegraphics[width=0.25\linewidth]{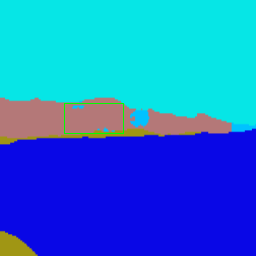}}
  \hfil
  \subfloat{\includegraphics[width=0.25\linewidth]{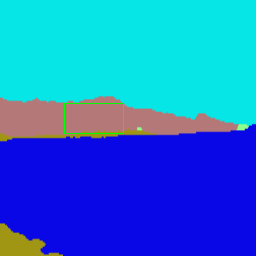}}
  \hfil
  \subfloat{\includegraphics[width=0.25\linewidth]{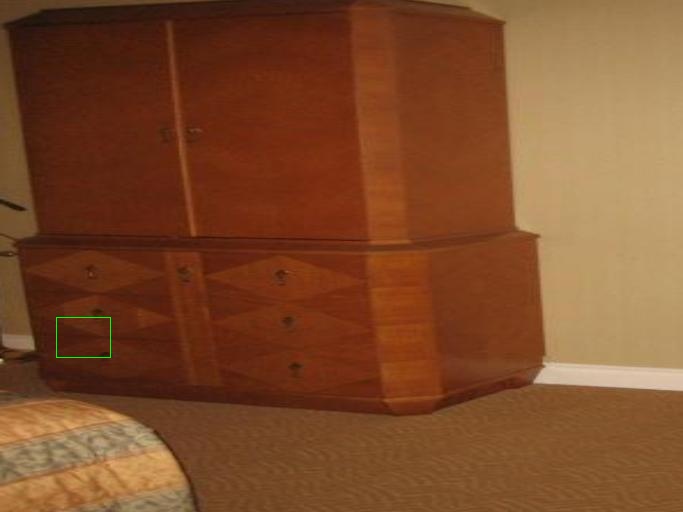}}
  \hfil
  \subfloat{\includegraphics[width=0.25\linewidth]{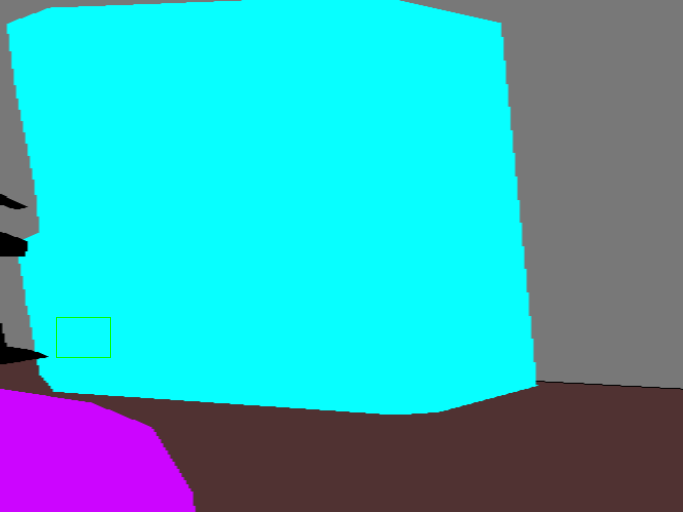}}
  \hfil
  \subfloat{\includegraphics[width=0.25\linewidth]{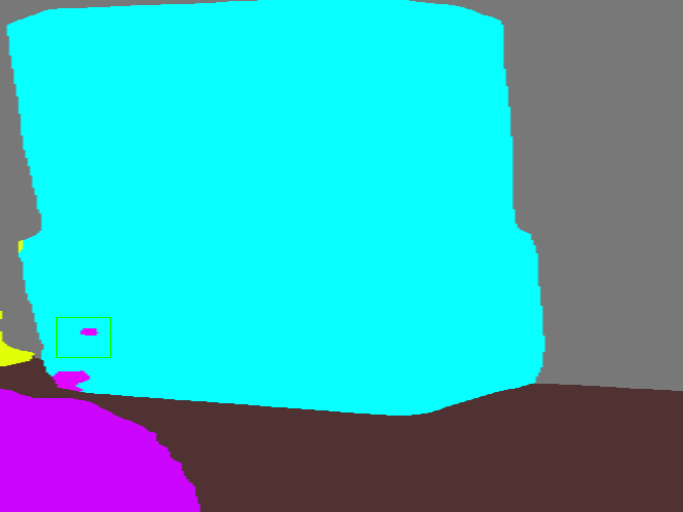}}
  \hfil
  \subfloat{\includegraphics[width=0.25\linewidth]{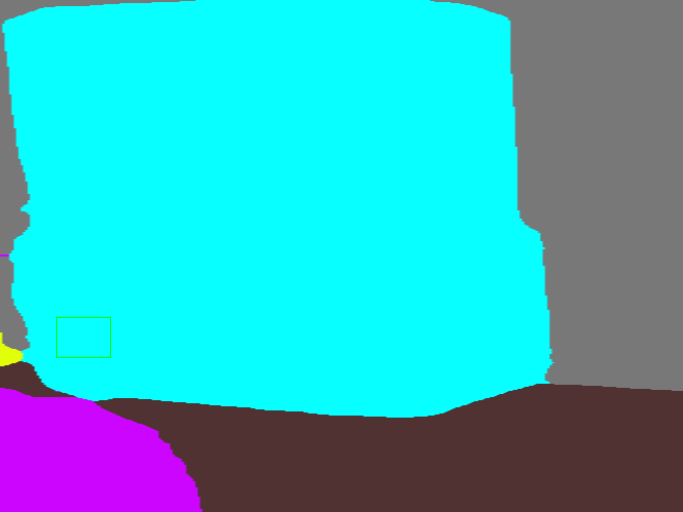}}
  \caption{\textbf{Additional qualitative comparison on ADE20K.} From left to right: input image, ground-truth semantic labels, and predictions of OCRNet~\cite{YuanCW19OCRNet} and InSeIn. Best viewed on a screen and zoomed in.}
  \label{fig:discontinued class_50}
\end{figure*}

\PAR{ADE20K.} In Fig.~\ref{fig:discontinued class_50} showing additional qualitative results on the ADE20K val set, we observe that in the prediction of the basic OCRNet~\cite{YuanCW19OCRNet} network on the first example, the building includes sky, which is rectified in the InSeIn-upgraded version of the network. In the second example, we observe that a segment of bed is infeasibly included in a cupboard segment, which is corrected by our InSeIn model. In both of the cases, the infeasible region is marked with \emph{green} bounding boxes.

\begin{figure*}[tb]
  \centering
  \subfloat{\includegraphics[width=0.25\textwidth]{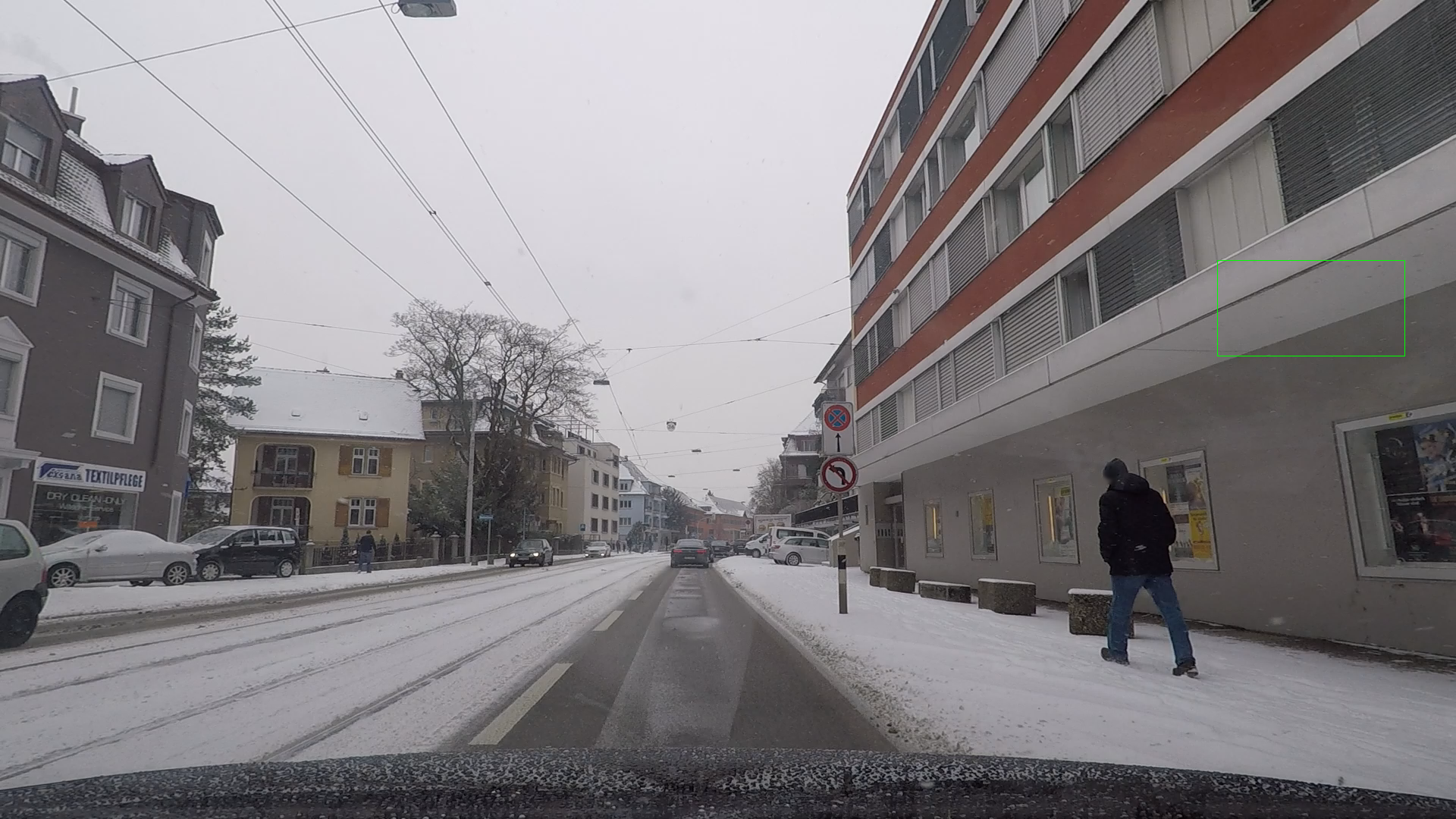}}
  \hfil
  \subfloat{\includegraphics[width=0.25\textwidth]{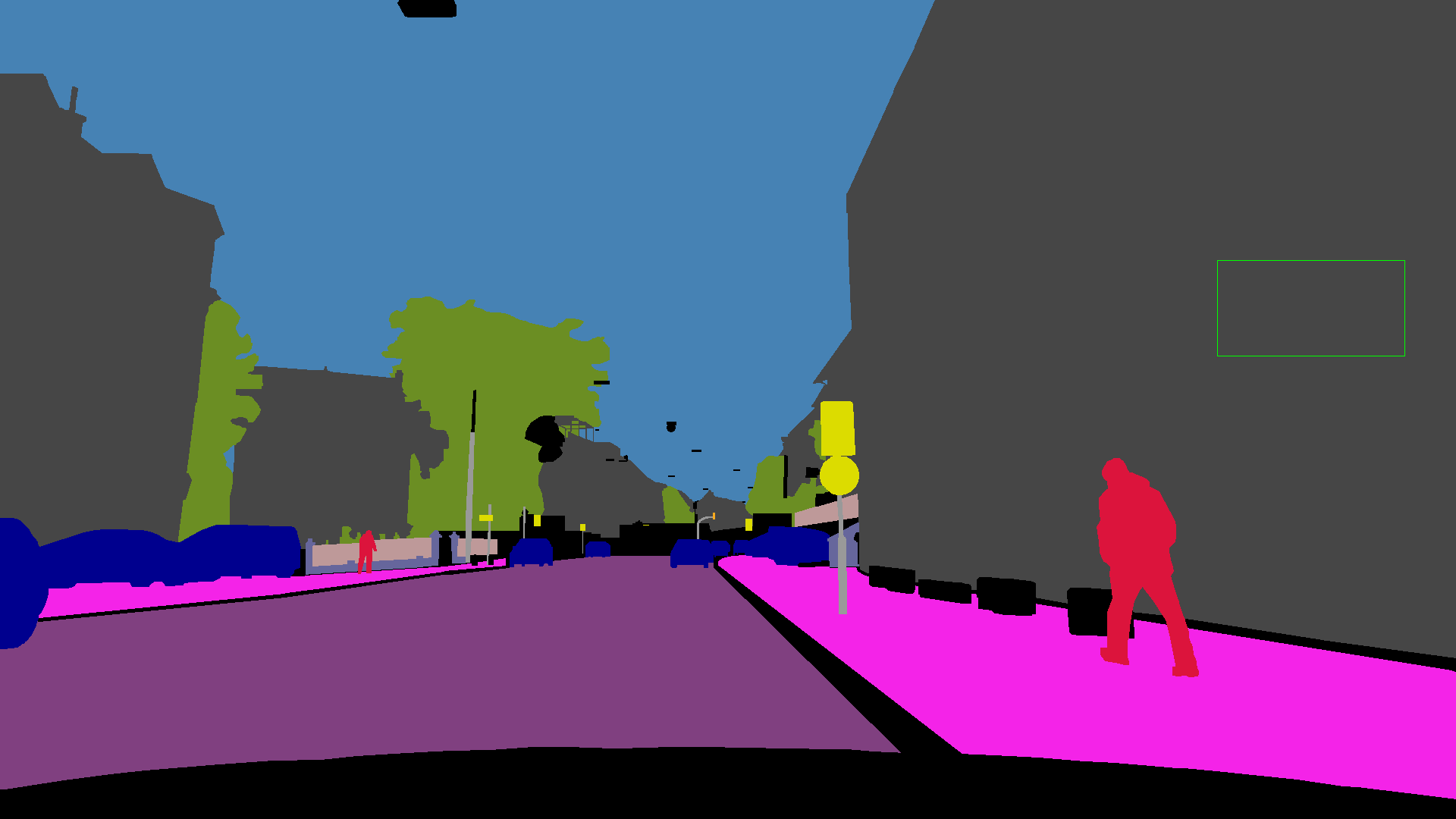}}
  \hfil
  \subfloat{\includegraphics[width=0.25\textwidth]{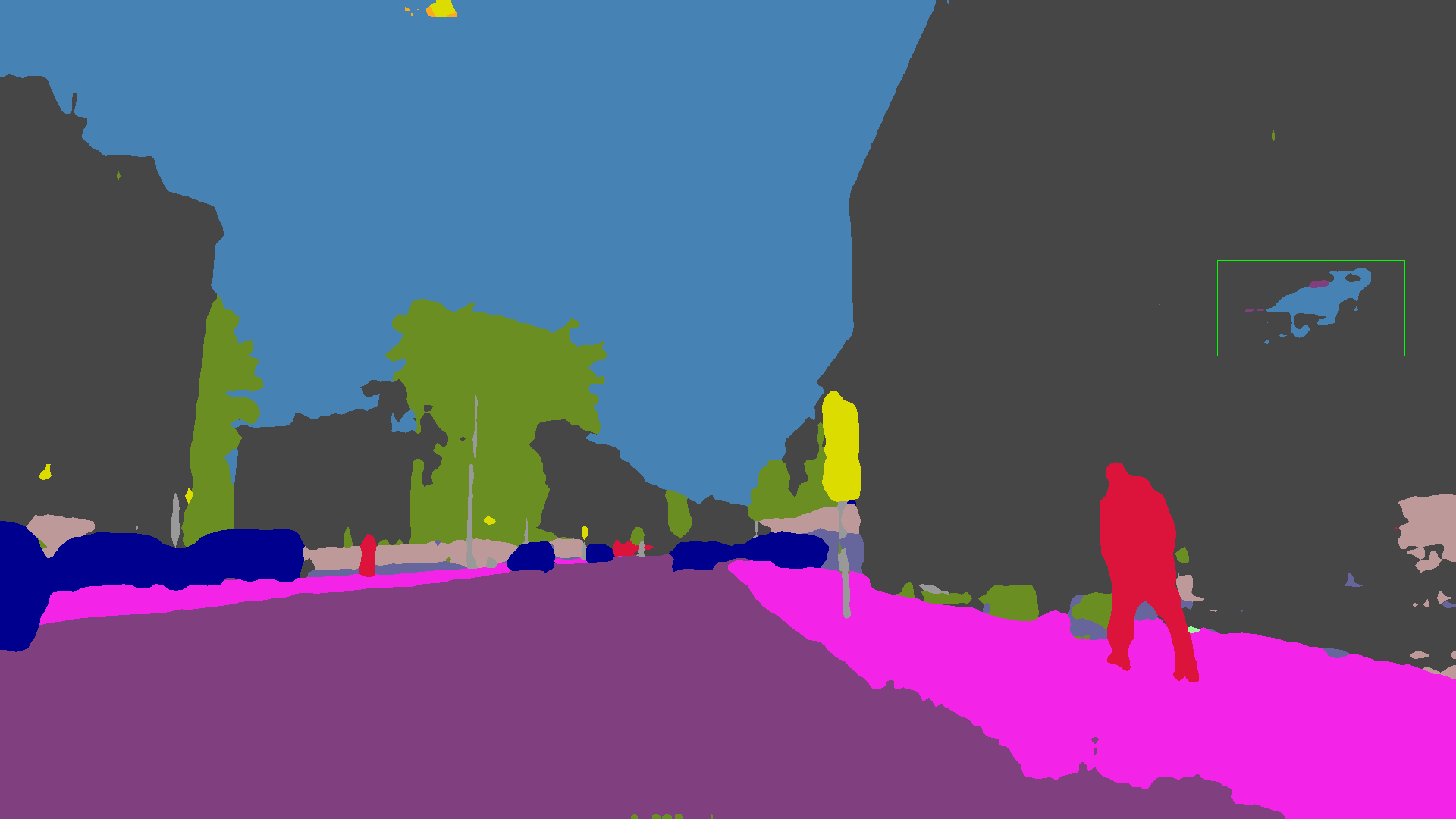}}
  \hfil
  \subfloat{\includegraphics[width=0.25\textwidth]{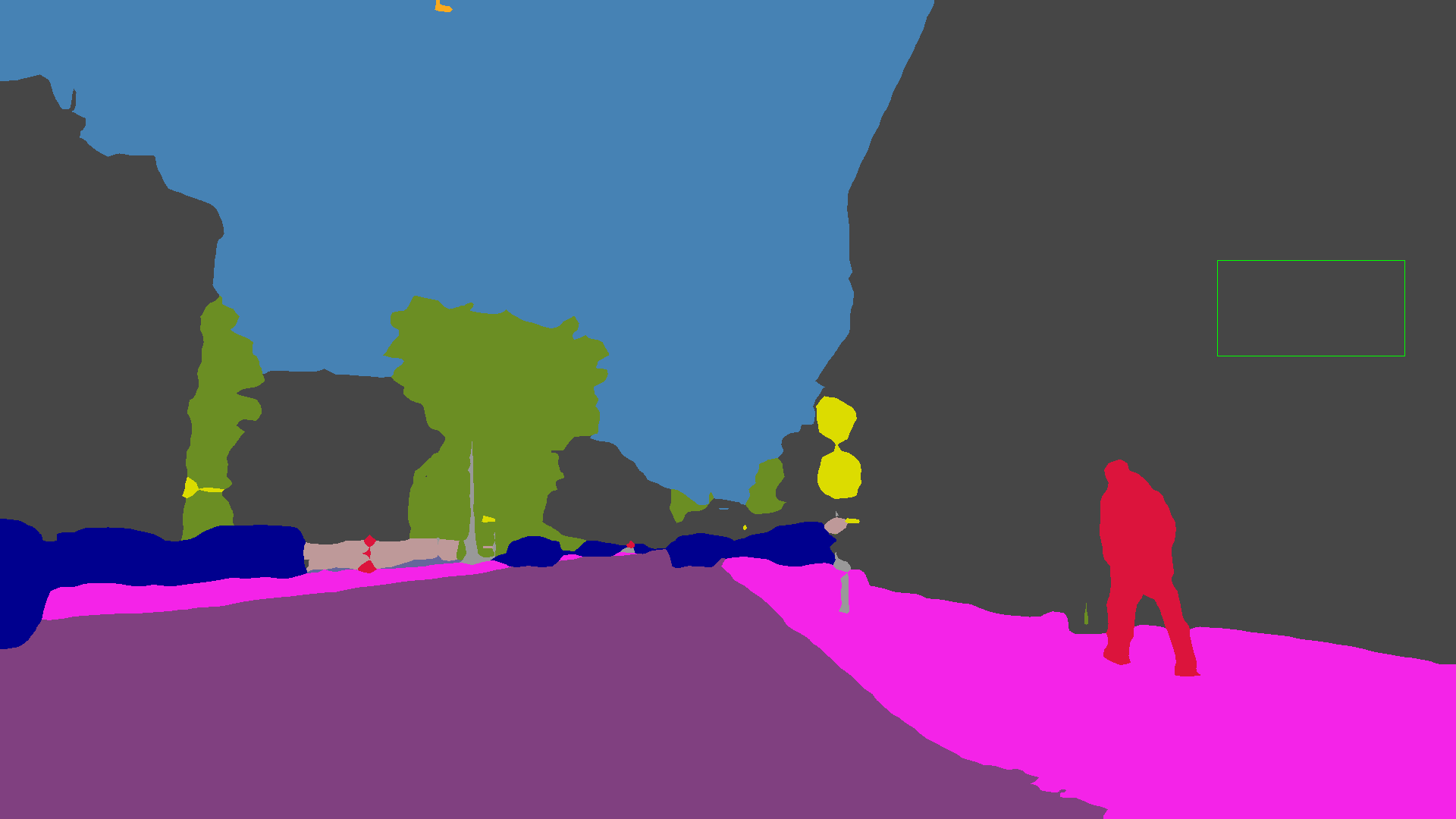}}
  \caption{\textbf{Additional qualitative comparison on ACDC.} From left to right: input image, ground-truth semantic labels, and predictions of SegFormer-B4~\cite{xie2021segformer} and InSeIn. Best viewed on a screen and zoomed in.}
  \label{fig:discontinued class_52}
\end{figure*}

\PAR{ACDC.} In the first example of Fig.~\ref{fig:discontinued class_52}, we observe that a segment of sky is completely included in a building marked with \emph{green} bounding boxes, which is a physical anomaly and is solved by InSeIn.

\begin{figure*}[tb]
  \centering
  \subfloat{\includegraphics[width=0.25\textwidth]{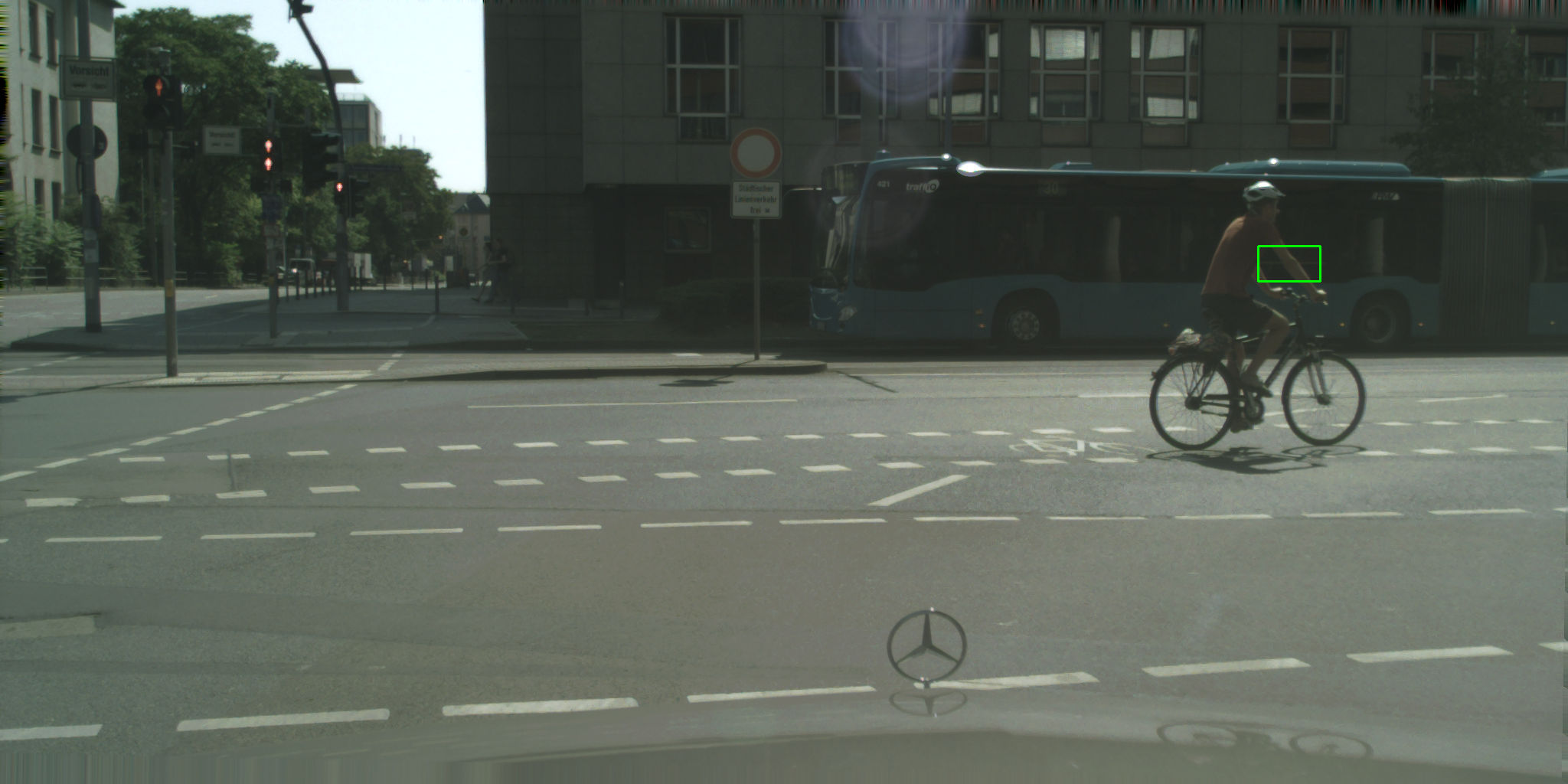}}
  \hfill
  \subfloat{\includegraphics[width=0.25\textwidth]{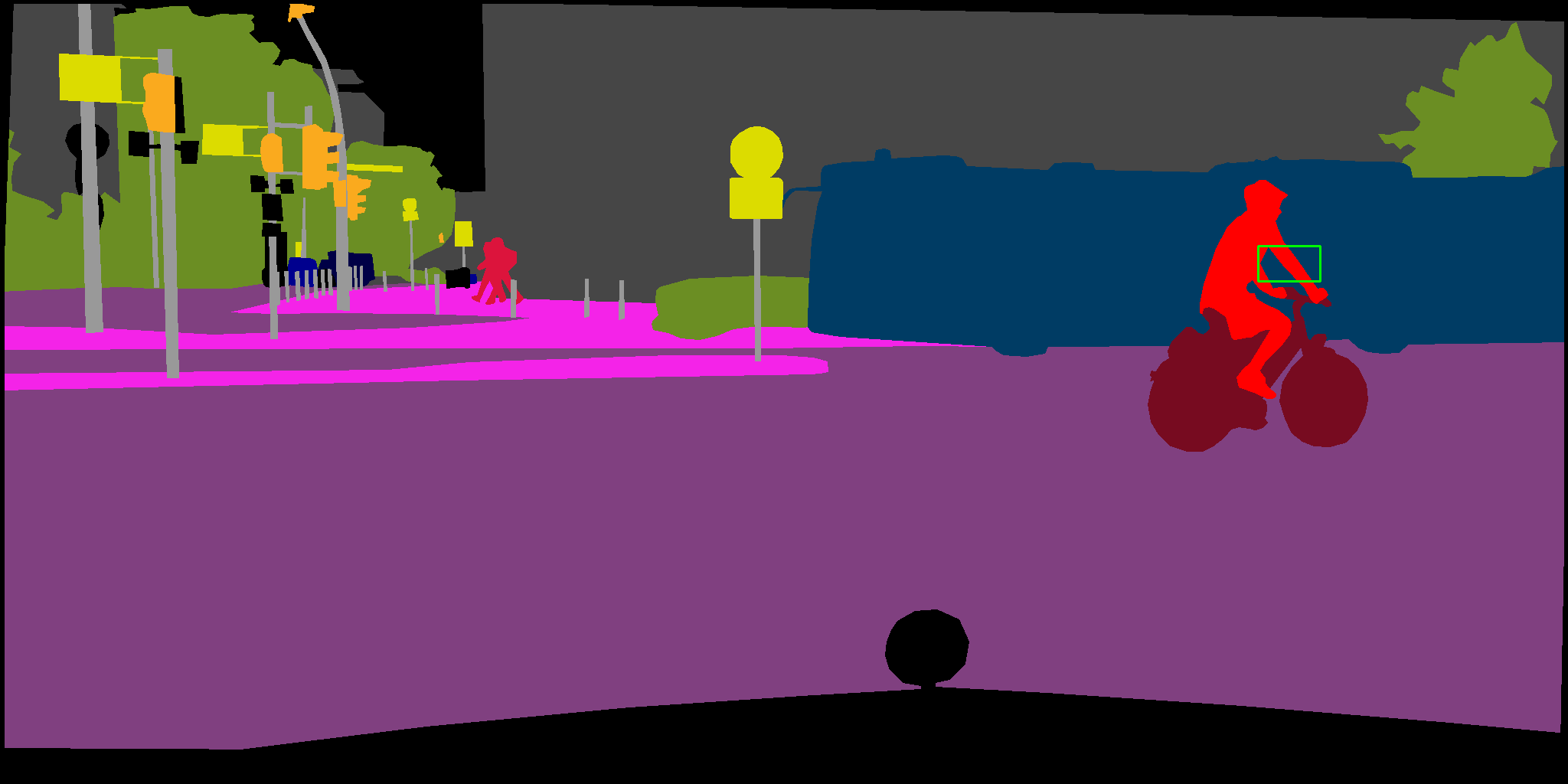}}
  \hfil
  \subfloat{\includegraphics[width=0.25\textwidth]{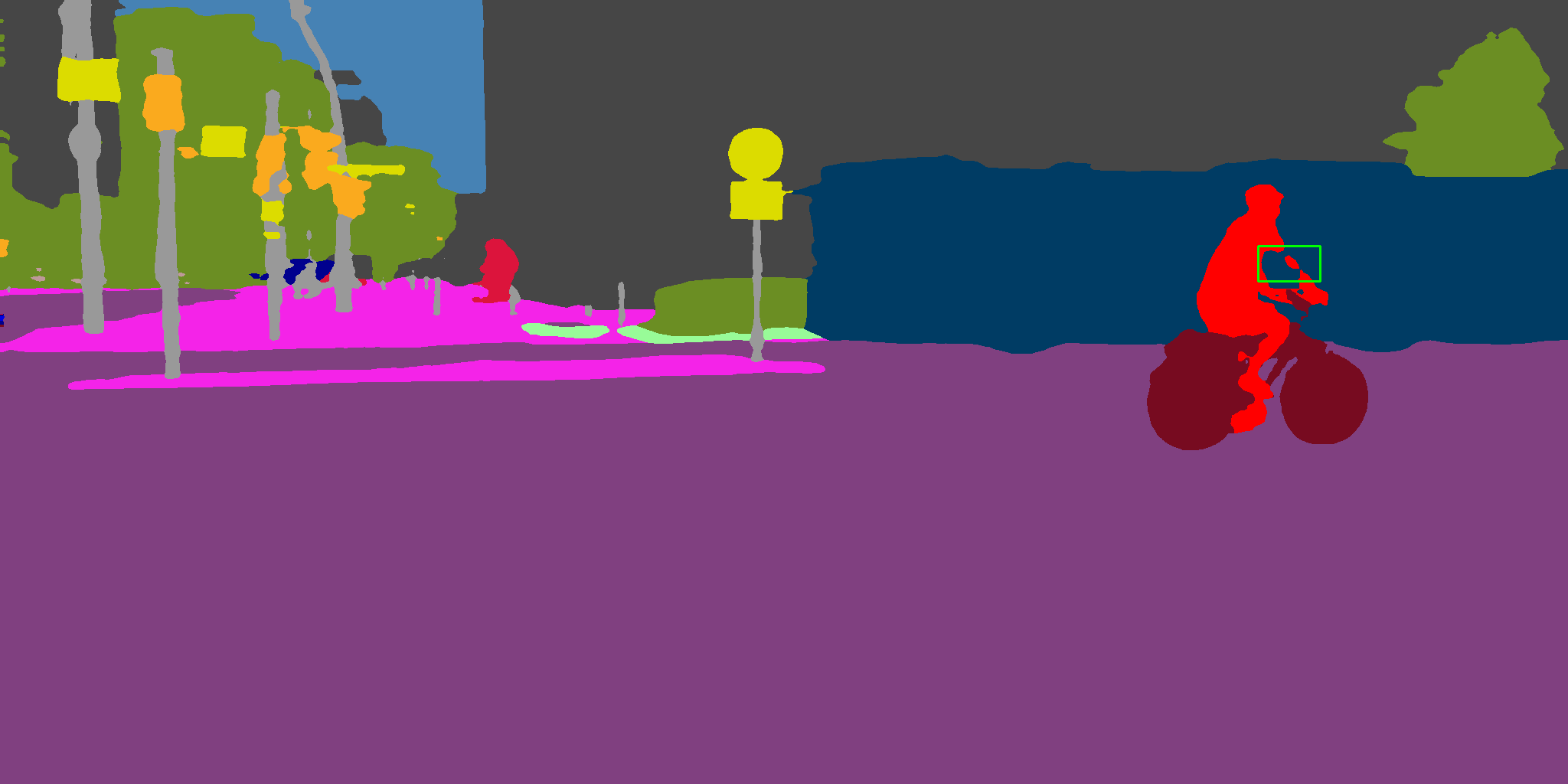}}
  \hfil
  \subfloat{\includegraphics[width=0.25\textwidth]{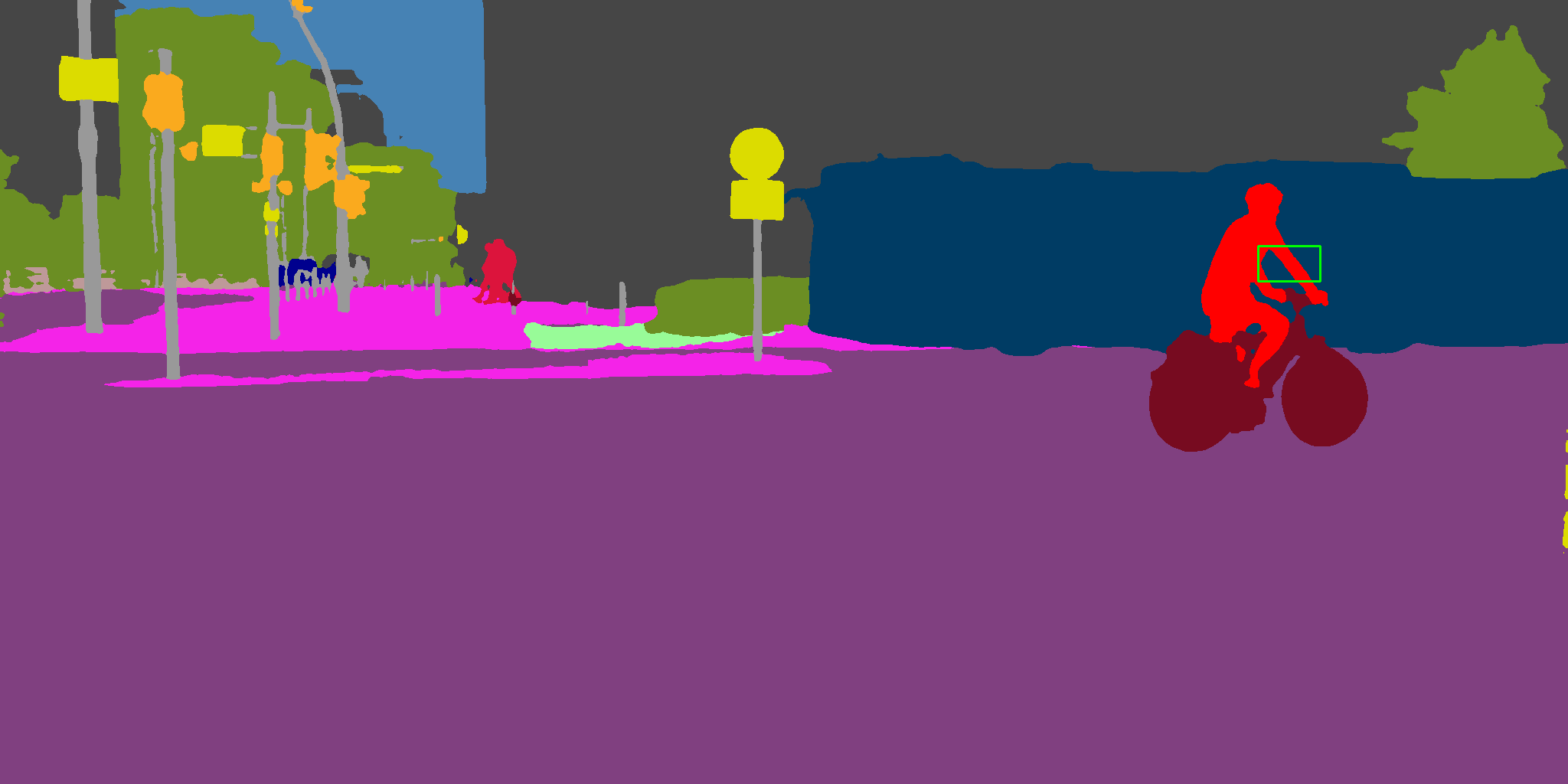}}
  \hfill
 \subfloat{\includegraphics[width=0.25\linewidth]{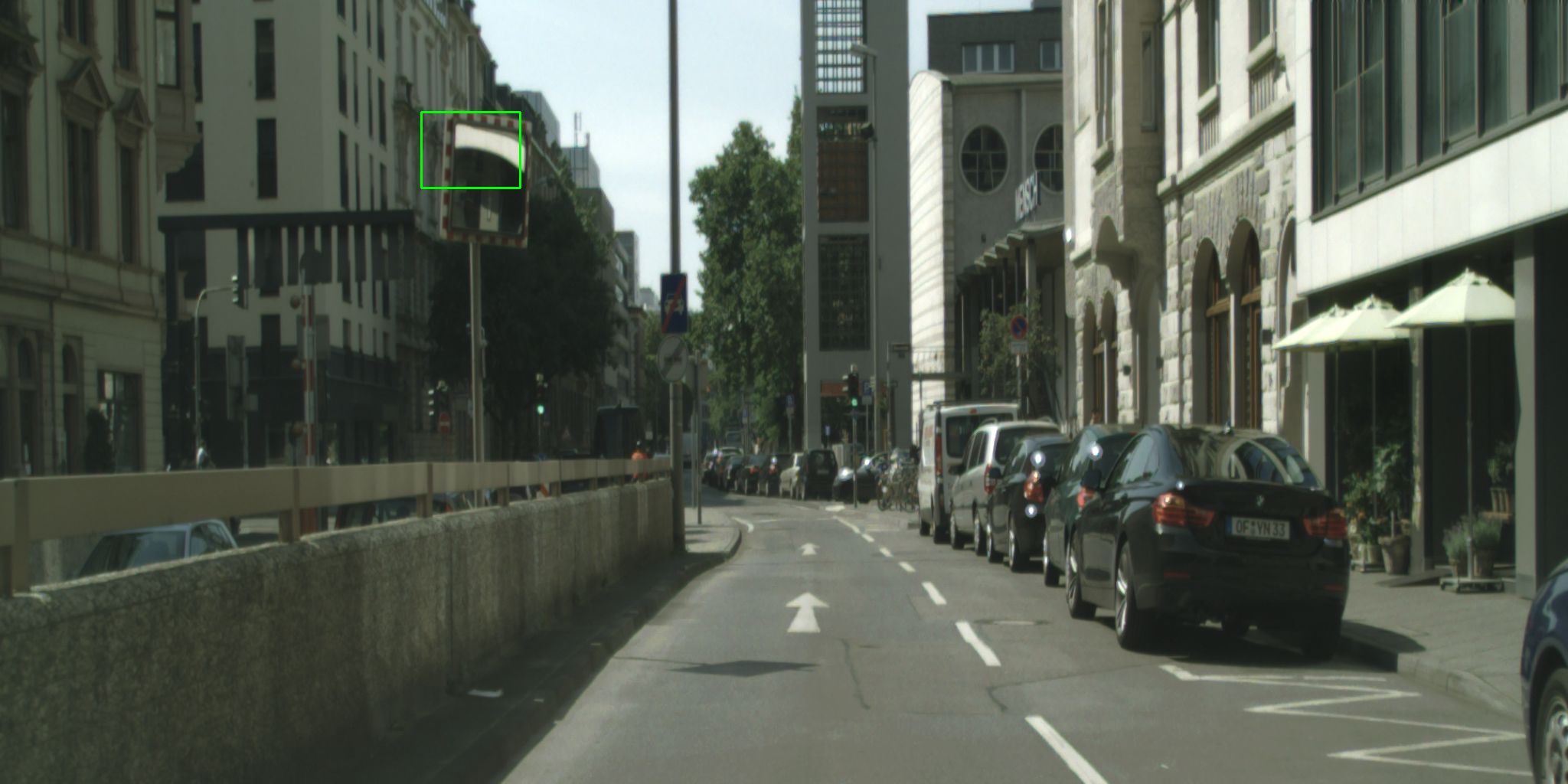}}
  \hfil
  \subfloat{\includegraphics[width=0.25\linewidth]{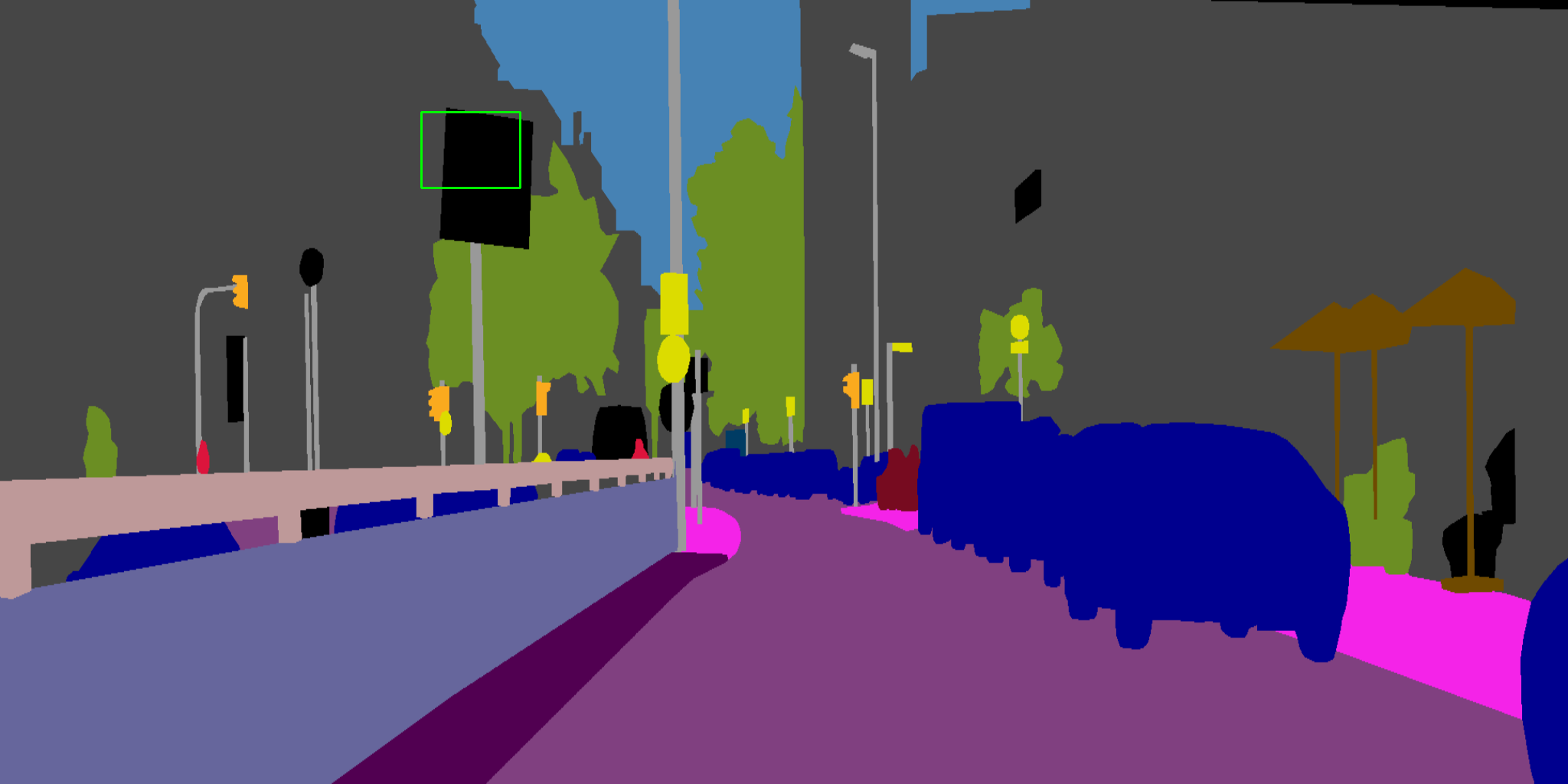}}
  \hfil
  \subfloat{\includegraphics[width=0.25\linewidth]{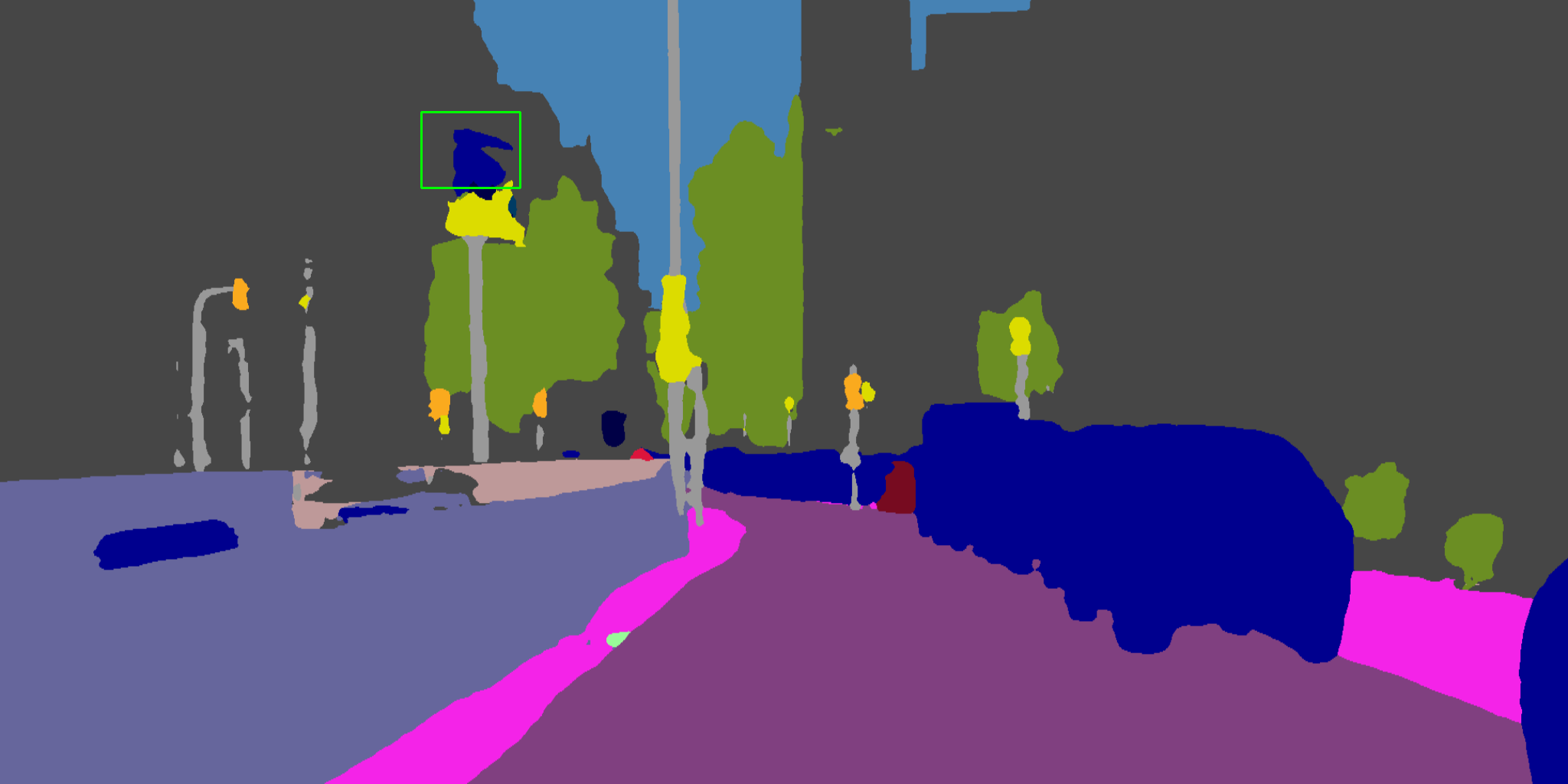}}
  \hfil
  \subfloat{\includegraphics[width=0.25\linewidth]{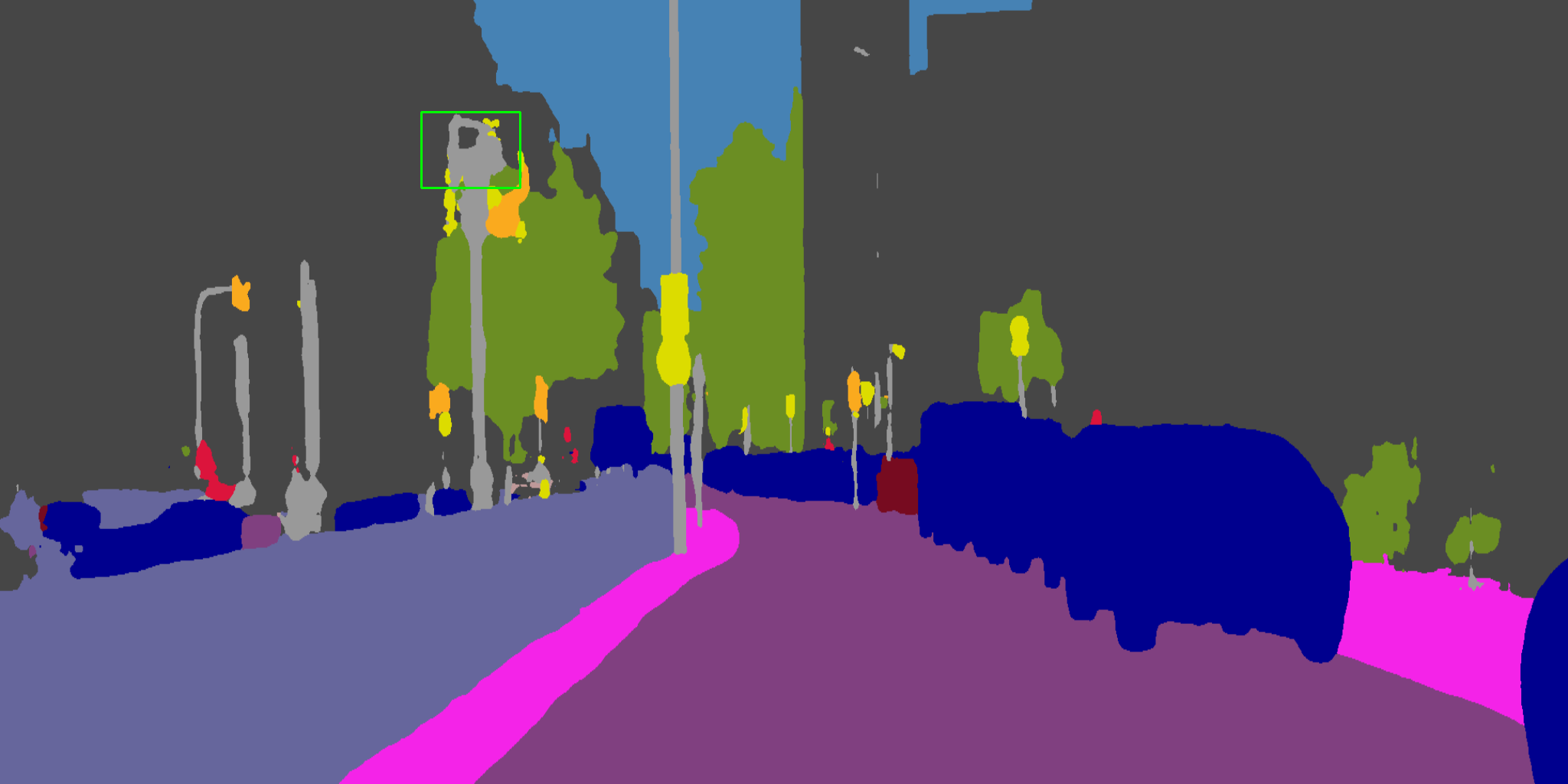}}
  \caption{\textbf{Additional qualitative comparison on Cityscapes.} From left to right: input image, ground-truth semantic labels, and predictions of Mask2Former~\cite{Cheng2022Mask2Former} and InSeIn. Best viewed on a screen and zoomed in.}
  \label{fig:discontinued class_3}
\end{figure*}

\begin{figure*}[tb]
  \centering
  \subfloat{\includegraphics[width=0.25\linewidth]{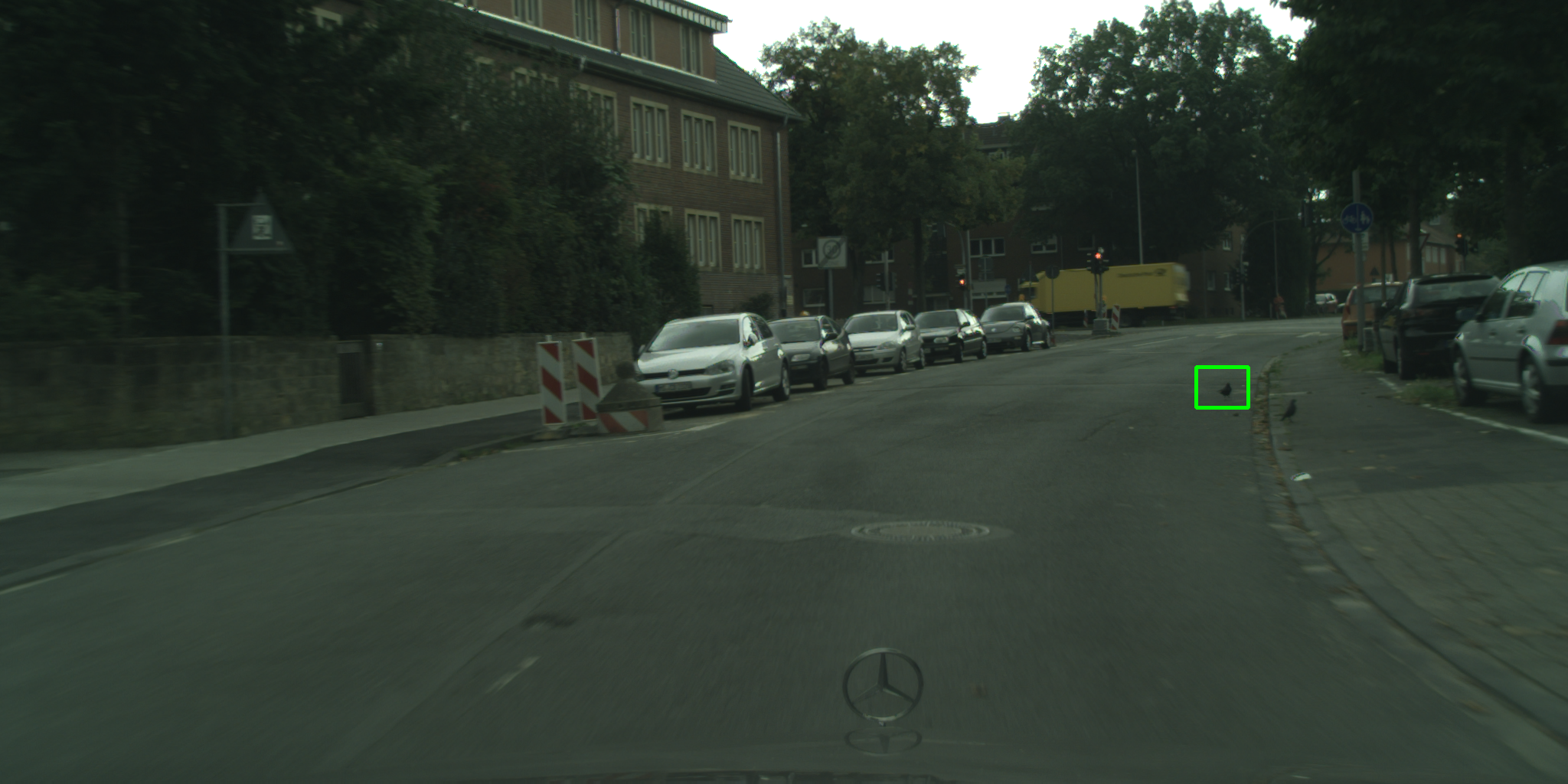}}
  \hfill
  \subfloat{\includegraphics[width=0.25\linewidth]{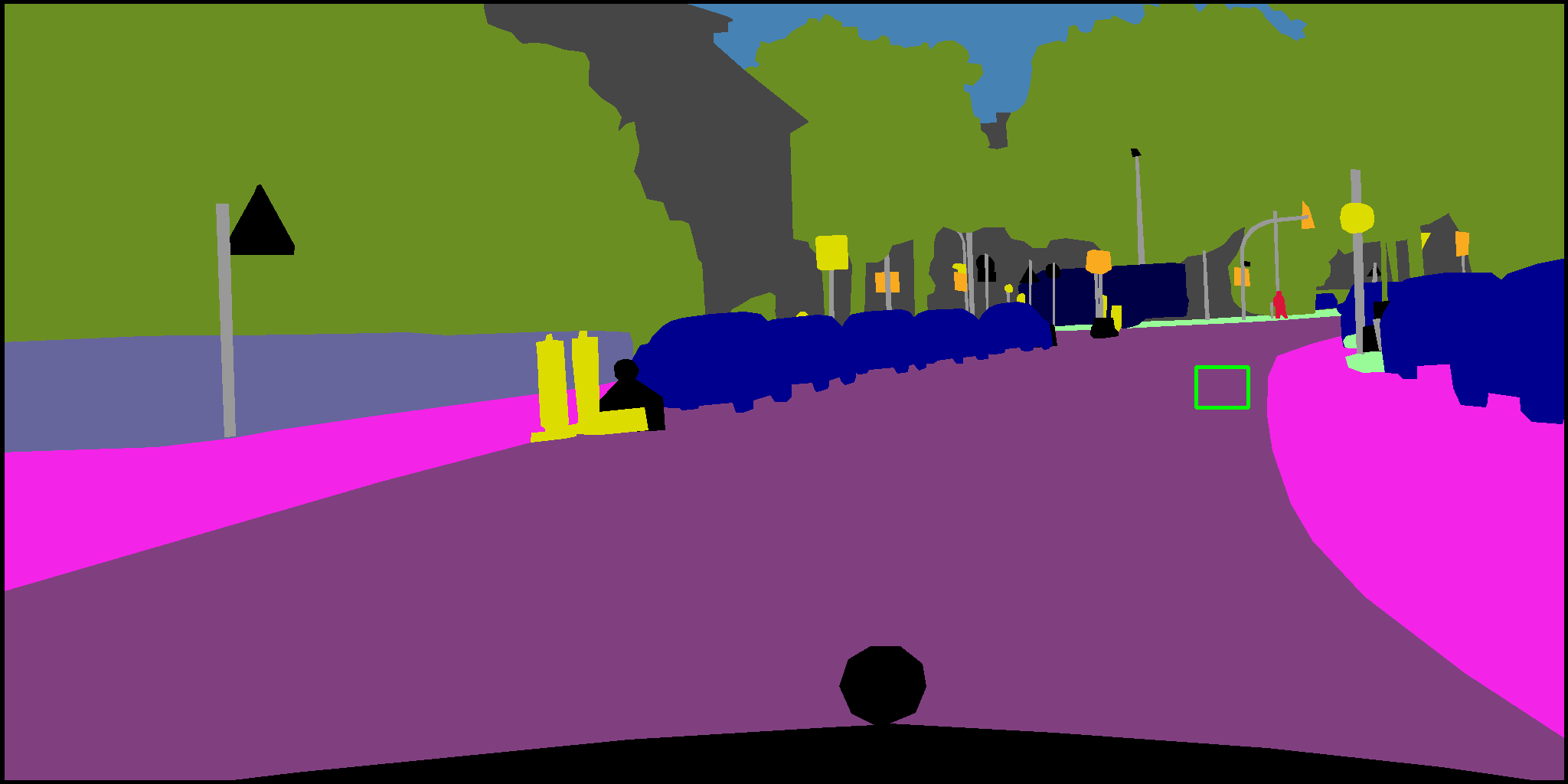}}
  \hfill
    \subfloat{\includegraphics[width=0.25\linewidth]{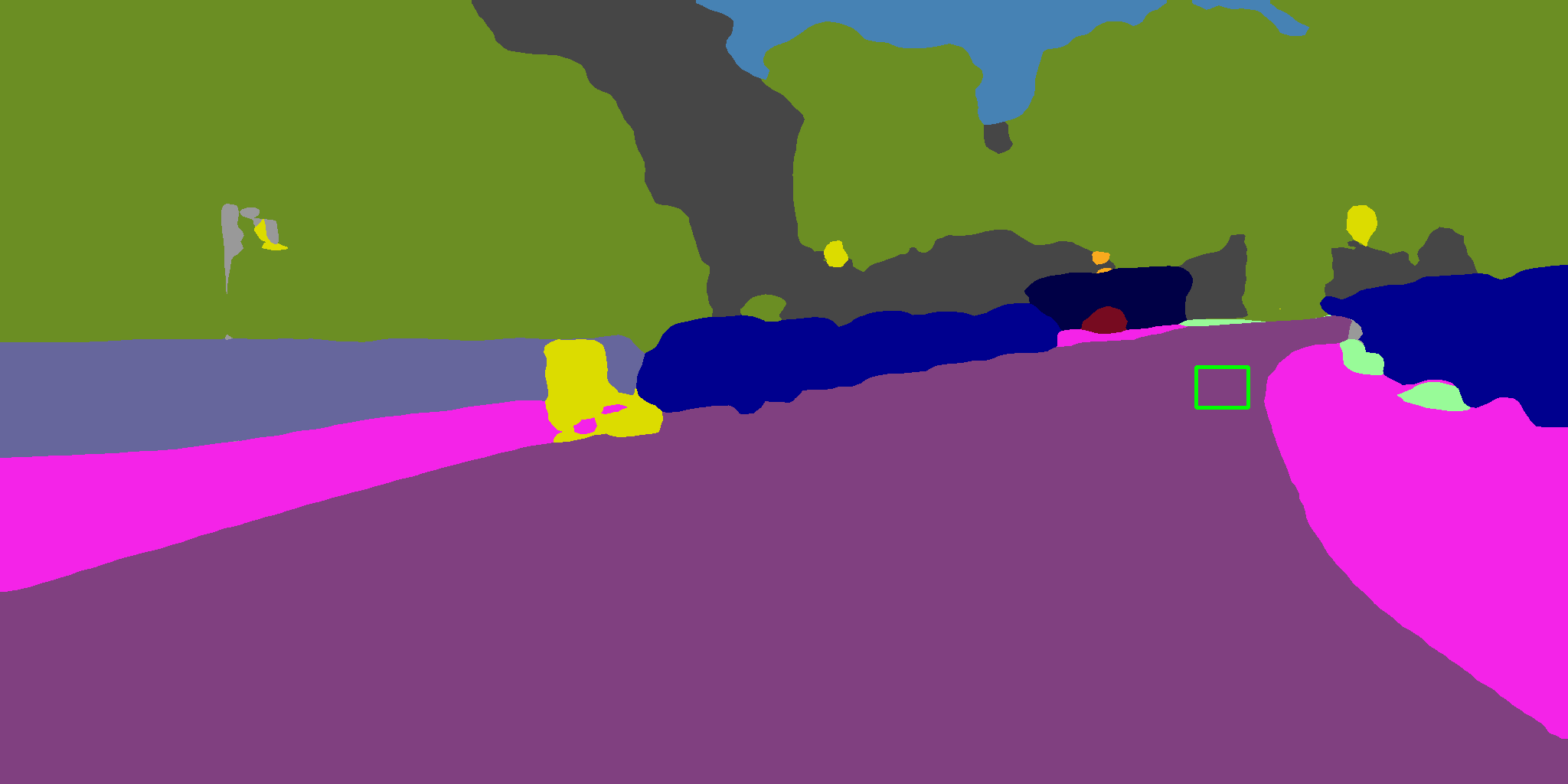}} 
  \hfill  
  \subfloat{\includegraphics[width=0.25\linewidth]{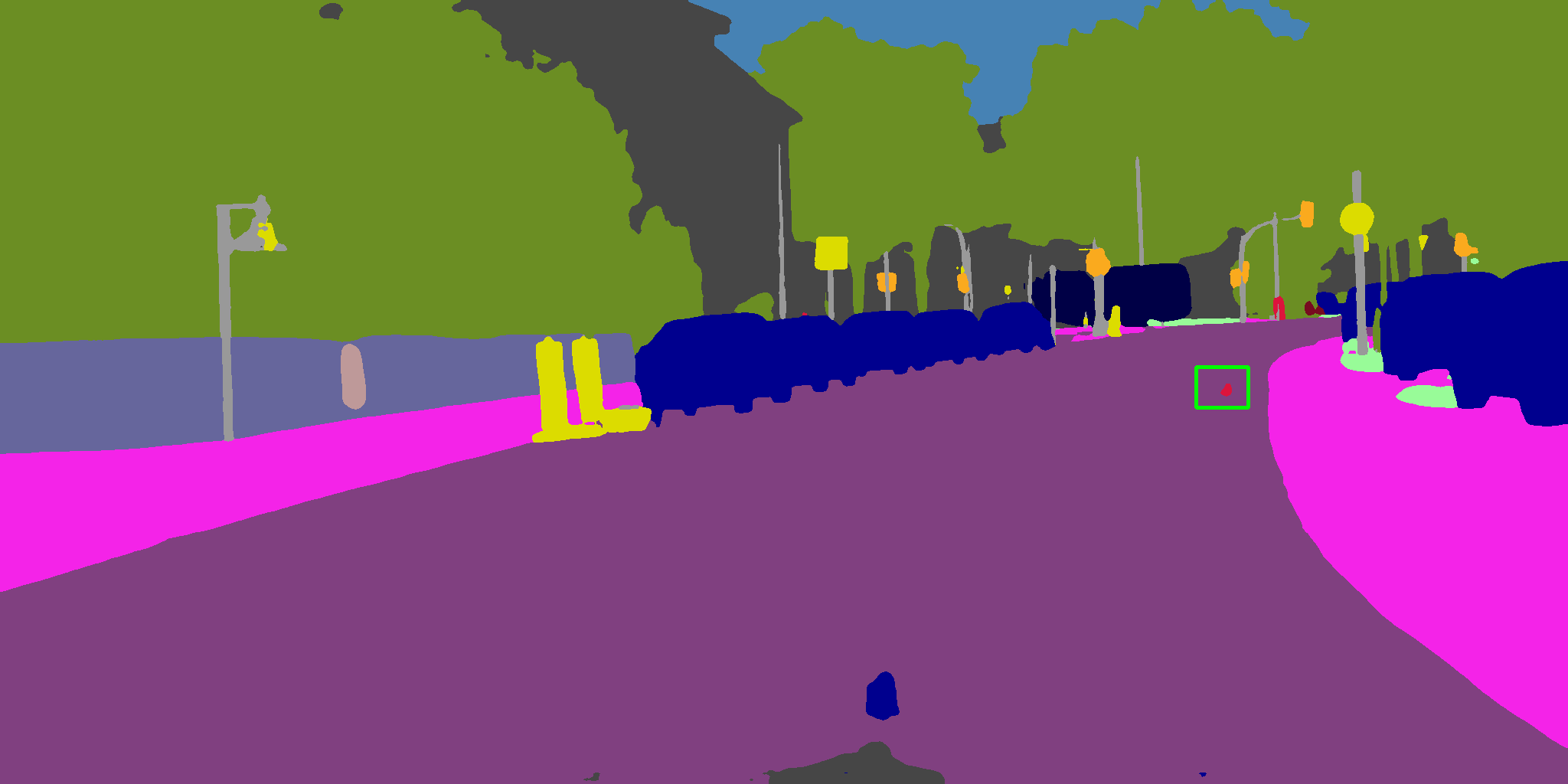}}   
  \caption{\textbf{Failure case in class-wise IOU comparison for Cityscapes.} From left to right:input image,ground truth, baseline-prediction OCRNet~\cite{YuanCW19OCRNet} and InSeIn. Best viewed on a screen and zoomed in.}
  \label{fig:cityscapes_failure_case}
\end{figure*}

\PAR{Cityscapes.} In Fig.~\ref{fig:discontinued class_3}, we note that a rider segment corresponding to the rider's arm is infeasibly included in a bus segment in the prediction of Mask2Former~\cite{Cheng2022Mask2Former}, whereas InSeIn manages to correctly connect this part of the arm with the rest of the rider's body. In the second example, a segment of the truck is infeasibly included by the building segment. InSeIn rectifies the anomaly. The infeasible regions are marked with \emph{green} bounding boxes.



\section{Discussion over choosing SOTA models} 

We have chosen DeepLab series~\cite{ChenDeepLab2018,deeplabv3plus2018}, HRNet~\cite{WangSCJDZLMTWLX19}, OCRNet~\cite{YuanCW19OCRNet} because they are top-performing CNN-based methods for semantic segmentation. SegFormer~\cite{xie2021segformer} is chosen since it is a top-performing CNN and transformer-based method. Mask2Former~\cite{Cheng2022Mask2Former} and OneFormer~\cite{jain2023oneformer} are chosen as they are standard mask-based methods. SegMAN~\cite{SegMAN} is the latest transformer-based method outperforming Mask2Former.
On the other hand, InSeIn is a light-weight (as it does not contain any learned parameter like these baselines) plug-and-play module, which, when plugged to these baselines and retrained on the 3 datasets, we have observed a consistent increase in the performance across the 3 datasets.

\section{Additional discussion on class-wise comparison}

 From Tab.3, we can observe that out of 95 comparisons, there are 16 comparisons where our method has a lower IOU score than its corresponding baseline. 
 
 This is due to the few new False Positive cases arising in the retrained version with InSeIn. This happens since, while re-training with InSeIn, we are initialising the baseline model's weights randomly, which leads the optimisation to converge to different local minima than the original baseline trained without InSeIn in the loss landscape. 
 
 And as InSeIn only penalises infeasible pair inclusion, some of these False Positives do not belong to the category of infeasible inclusion, and they remain in the prediction feature map. For example, in Fig.~\ref{fig:cityscapes_failure_case}, we can observe that in the re-trained version, a small segment of the person class is included by the road class (marked with green bounding box), which is a False Positive according to the ground truth, but it is not an infeasible inclusion with respect to the taxonomy of the Cityscapes dataset.

 \begin{table}[tb]
    \centering
    \caption{Wall-clock time for one SGD iteration on full input images and GPU memory usage during training for a batch size of 4.}
    \footnotesize
    \label{tab:complexity_comparison}
    \vspace{-3mm}
    \begin{tabular}{lccc}
        \toprule
        \textbf{Model}& \textbf{Space (GB)}& \textbf{\#Params}& \textbf{Time (sec.)}\\
        \midrule
        Mask2Former & 21 & 216M & 7.2\\ 
        Mask2Former w/ InSeIn & 41 & 216M & 8.4\\ 
        \midrule
        SegFormer-B4  &8 & 64.1M & 3.5\\
        SegFormer-B4 w/ InSeIn &29 & 64.1M &4.0\\
        \bottomrule
    \end{tabular}
    \vspace{-5mm}
\end{table}

\section{Discussion over memory usage and time complexity} 

Tab.~\ref{tab:complexity_comparison} demonstrates the memory usage and time complexity of InSeIn. The memory overhead induced by InSeIn only applies to \emph{training}, while at inference, we do not require any additional memory compared to the baseline networks. The parameters remain the same as the baselines since InseIn is free of any learnable parameters.

\end{document}